\documentclass[10pt,twocolumn,letterpaper]{article}

\usepackage[pagenumbers]{iccv}      

\definecolor{iccvblue}{rgb}{0.21,0.49,0.74}
\usepackage[pagebackref,breaklinks,colorlinks,allcolors=iccvblue]{hyperref}

\usepackage{multirow}
\usepackage{colortbl}  
\usepackage{array}   

\PassOptionsToPackage{x11names,table,xcdraw}{xcolor} 
\usepackage[most]{tcolorbox}  
\usepackage{xcolor}  
\definecolor{lightbeige}{HTML}{F5F5DC}  

\usepackage{amssymb}
\usepackage{fontawesome}  

\usepackage[T1]{fontenc}
\usepackage{CJKutf8}

\definecolor{mygreen}{rgb}{0.5, 1.0, 0.5}  
\definecolor{myred}{rgb}{1.0, 0.5, 0.5}    

\def\paperID{3443} 
\def\confName{ICCV}
\def\confYear{2025}

\title{OODFace: Benchmarking Robustness of Face Recognition under Common Corruptions and Appearance Variations}

\author{Caixin Kang$^{1}$, Yubo Chen$^{1}$, Shouwei Ruan$^{1}$, Shiji Zhao$^{1}$, Ruochen Zhang$^{1}$, \\ Jiayi Wang$^{2}$, Shan Fu$^{2}$, Xingxing Wei$^{1}$\thanks{Corresponding authors.} \\
  $^{1}$ Institute of Artificial Intelligence, Beihang University \hspace{1ex} \\
  $^{2}$ China Academy of Information and Communications Technology \\
  \footnotesize{\texttt{\{caixinkang,xxwei\}@buaa.edu.cn}}
}

\begin{document}
\begin{CJK}{UTF8}{gbsn}

\maketitle
\begin{abstract}
With the rise of deep learning, facial recognition technology has seen extensive research and rapid development. Although facial recognition is considered a mature technology, we find that existing open-source models and commercial algorithms lack robustness in certain complex Out-of-Distribution (OOD) scenarios, raising concerns about the reliability of these systems. In this paper, we introduce \textbf{OODFace}, which explores the OOD challenges faced by facial recognition models from two perspectives: common corruptions and appearance variations. We systematically design 30 OOD scenarios across 9 major categories tailored for facial recognition. By simulating these challenges on public datasets, we establish three robustness benchmarks: LFW-C/V, CFP-FP-C/V, and YTF-C/V. We then conduct extensive experiments on 19 facial recognition models and 3 commercial APIs, along with extended physical experiments on face masks to assess their robustness. Next, we explore potential solutions from two perspectives: defense strategies and Vision-Language Models (VLMs). Based on the results, we draw several key insights, highlighting the vulnerability of facial recognition systems to OOD data and suggesting possible solutions. Additionally, we offer a unified toolkit that includes all corruption and variation types, easily extendable to other datasets. We hope that our benchmarks and findings can provide guidance for future improvements in facial recognition model robustness.

\end{abstract}

\vspace{-3ex}    
\section{Introduction}
\label{sec:intro}

In recent years, the rise of deep learning has significantly advanced facial recognition (FR) technology, leading to extensive research and rapid development worldwide. Innovations in loss functions such as SphereFace~\cite{liu2017sphereface}, CosFace~\cite{wang2018cosface} and ArcFace~\cite{deng2019arcface}, have greatly enhanced the efficiency of FR models, achieving unprecedented accuracy under standard conditions. Additionally, the release of large-scale facial datasets like MS-Celeb-1M~\cite{guo2016ms} and VGGFace2~\cite{cao2018vggface2} has further boosted algorithm performance. State-of-the-art open-source algorithms~\cite{kim2022adaface,boutros2022elasticface,dan2023transface,dan2024topofr} have achieved near-perfect accuracy in FR, while mature commercial APIs from companies like Microsoft, Tencent and Alibaba have enabled the widespread deployment of FR systems in various fields, including security surveillance, identity verification, and financial services, indicating that FR has become a well-established technology.

\begin{figure}[t]
  \centering
   \includegraphics[width=0.99\linewidth]{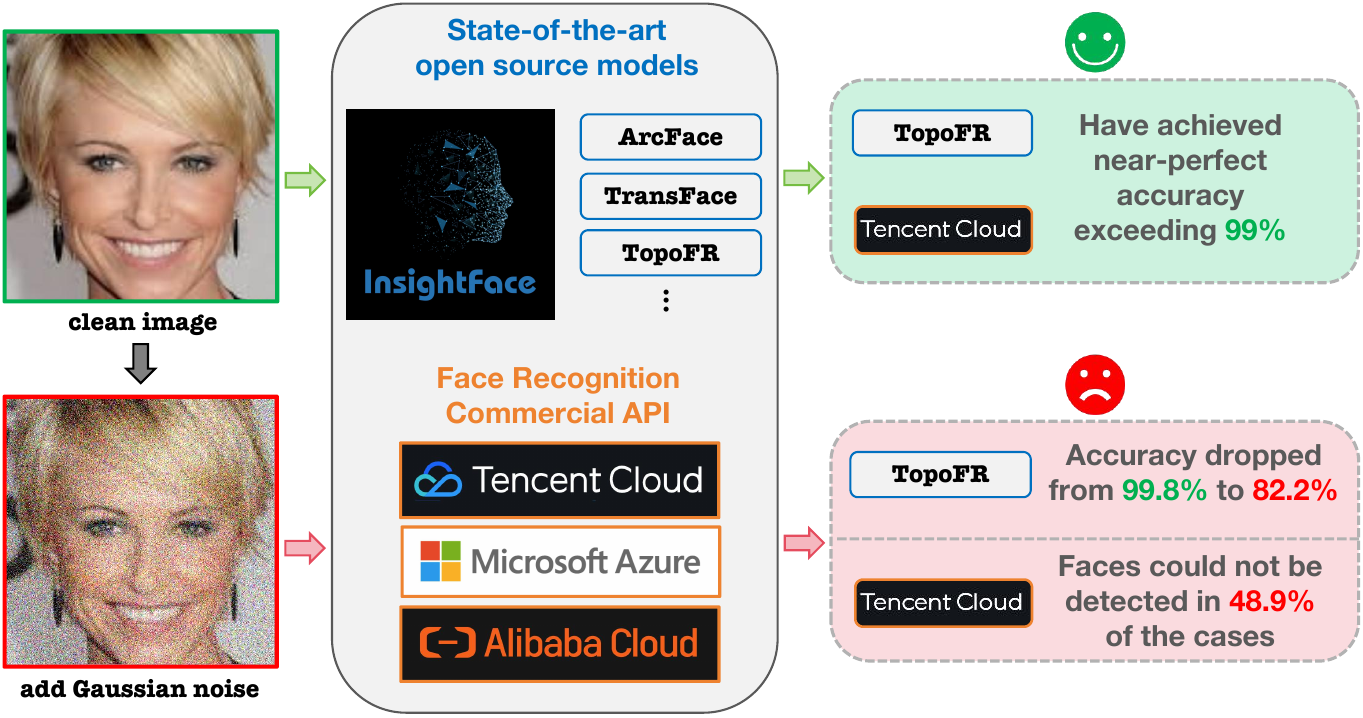}

   \caption{Challenges in FR systems. Simple Gaussian noise poses a threat to the performance of state-of-the-art open-source FR models and commercial FR APIs. (Accuracy tested on LFW.)}
   \label{fig:intro}
      \vspace{-3ex}
\end{figure}

\begin{tcolorbox}[colframe=black, colback=gray!10, coltitle=black, sharp corners=all, boxrule=0.5mm, boxsep=0.1mm]
   \textbf{(Problem statement)} Since FR has become a well-established technology, are FR models capable of handling all complex scenarios? 
\end{tcolorbox}

\begin{figure*}[!t]
  \centering
   \includegraphics[width=0.85\linewidth]{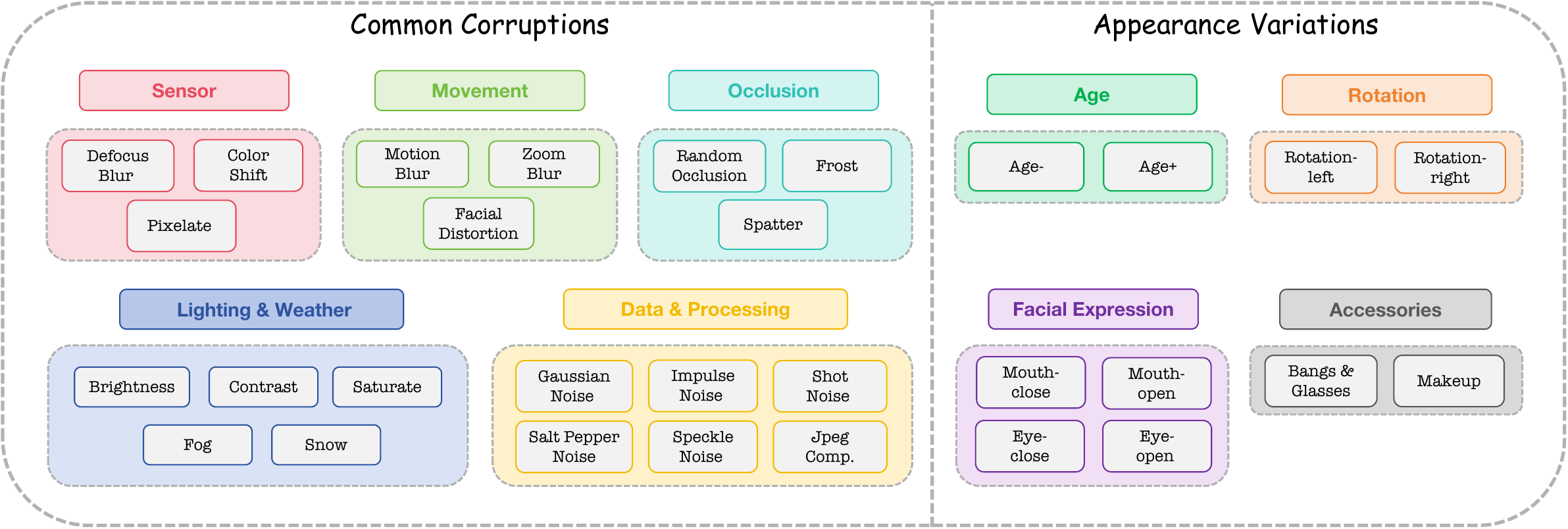}
   \vspace{-1ex}
   \caption{Overview of \textbf{OODFace}’s 30 OOD scenarios. OODs are divided into two major categories, common corruptions and appearance variations, further subdivided into 20 and 10 subcategories, each with 5 severity levels.}
   \label{fig:category2}
      \vspace{-3ex}
\end{figure*}

In real-life situations, users may encounter issues, such as failing to unlock apartment access during snowy weather or unsuccessful facial recognition for apps that have not been used for a while~\cite{ahsan2021evaluating}.
To explore the causes, we conduct a simple experiment by adding light Gaussian noise to facial images (potentially introduced during data processing or transmission), as shown in Fig.~\ref{fig:intro}. Although humans can easily recognize the person in both images, we find that state-of-the-art open-source models~\cite{dan2024topofr} and the commercial API of Tencent experience a significant drop in performance, with the detection success rate dropping to 82.2\%. The API fails to detect faces in 48.9\% of the images, disrupting the normal functioning of the FR system.

This indicates that while current FR algorithms perform well under undisturbed conditions, they are still vulnerable to various natural disturbances in real-world applications. Despite the existence of many datasets, such as typical benchmarks~\cite{cao2018vggface2,guo2016ms,huang2008labeled,sengupta2016frontal,moschoglou2017agedb,wolf2011face}, most of them focus on ideal conditions. Previous studies have examined the robustness of FR, but they mainly address adversarial perturbations~\cite{yang2020robfr}, fairness issues~\cite{lin2024ai} or certain categories such as weather~\cite{ahsan2021evaluating}, occlusion~\cite{zhang2018facial,neto2022beyond}. Thus, comprehensively characterizing challenging samples from diverse real-world Out-of-Distribution (OOD) scenarios and fairly evaluating the natural robustness of existing models within a unified framework remains an open problem.
\vspace{-1ex}

In this paper, we explore the OOD challenges faced by FR models from the perspectives of common corruptions and appearance variations. We systematically design \textbf{30} common OOD scenarios tailored for FR to rigorously assess the OOD robustness of current models. Common corruptions are classified into five categories: \textit{Lighting \& Weather, Sensor Failures, Motion Errors, Data Processing Corruptions, and Object Occlusion,} covering 20 subclasses. Appearance variations are grouped into four categories: \textit{Age, Facial Expression, Head Pose, and Accessories,} comprising 10 subclasses, covering most real-world disturbances (see Fig.~\ref{fig:category2}). 
Following~\cite{hendrycks2019benchmarking}, each scenario includes five severity levels, resulting in \textbf{150} unique corruptions and variations.
Many of these scenarios, such as Facial Distortion, Random Occlusion, Age, and Accessories, are designed specifically for FR and have not been previously explored by~\cite{yang2020robfr,lin2024ai,ahsan2021evaluating,zhang2018facial,neto2022beyond}, which also investigate OOD challenges but focus on specific OOD types 
with a limited evaluation scope and fewer FR models (Detailed comparison in Appendix.~\ref{sup:A3}, Tab.~\ref{tab:compare}).
By applying these scenarios to standard FR datasets—LFW~\cite{huang2008labeled}, CFP-FP~\cite{sengupta2016frontal}, and YTF~\cite{wolf2011face}—we establish three comprehensive benchmarks for robustness evaluation: \textbf{LFW-C/V}, \textbf{CFP-C/V}, and \textbf{YTF-C/V}, representing common corruptions and appearance variations, respectively. We hope these large-scale OOD datasets can serve as standard benchmarks for fair and comprehensive evaluation of OOD robustness in FR models, facilitating future research in the field.

We conduct extensive experiments to compare the OOD robustness of existing FR models. Specifically, we evaluate \textbf{19 open-source models} and \textbf{3 commercial APIs}, covering a variety of loss functions, backbones, and model sizes, across the LFW-C/V, CFP-FP-C/V, and YTF-C/V benchmarks.
Based on the evaluation results, we find that:
1) Common corruptions robustness is uncorrelated with clean data performance, whereas appearance variations robustness shows the opposite trend;
2) FR models show significant performance drops under common corruptions, with the largest impact from \textit{Data \& Processing};
3) different FR models show varying sensitivity to different types of OOD scenarios.
More discussions are provided in Sec.~\ref{sec:5}. Additionally, we explore the effectiveness of \textbf{physical face masks} under OOD conditions. 
Finally, we examine \textbf{10 robust models} using \textit{input transformation}~\cite{xie2017mitigating,xu2017feature},  \textit{adversarial training}~\cite{mkadry2017towards,zhang2019theoretically} and \textbf{3 restoration methods}~\cite{wang2021towards, zhou2022towards, lin2024diffbir} based on \textit{GANs}, \textit{Transformers}, and \textit{Diffusion models} as potential defense, and investigate leveraging the generalization capabilities of \textbf{Vision-Language Models} to address OOD challenges. However, the observed robustness gains are limited. Therefore, despite their widespread use in critical applications, FR models still exhibit vulnerabilities, and enhancing their robustness remains an open challenge.

\begin{figure*}[t]
  \centering
   \includegraphics[width=0.95\linewidth]{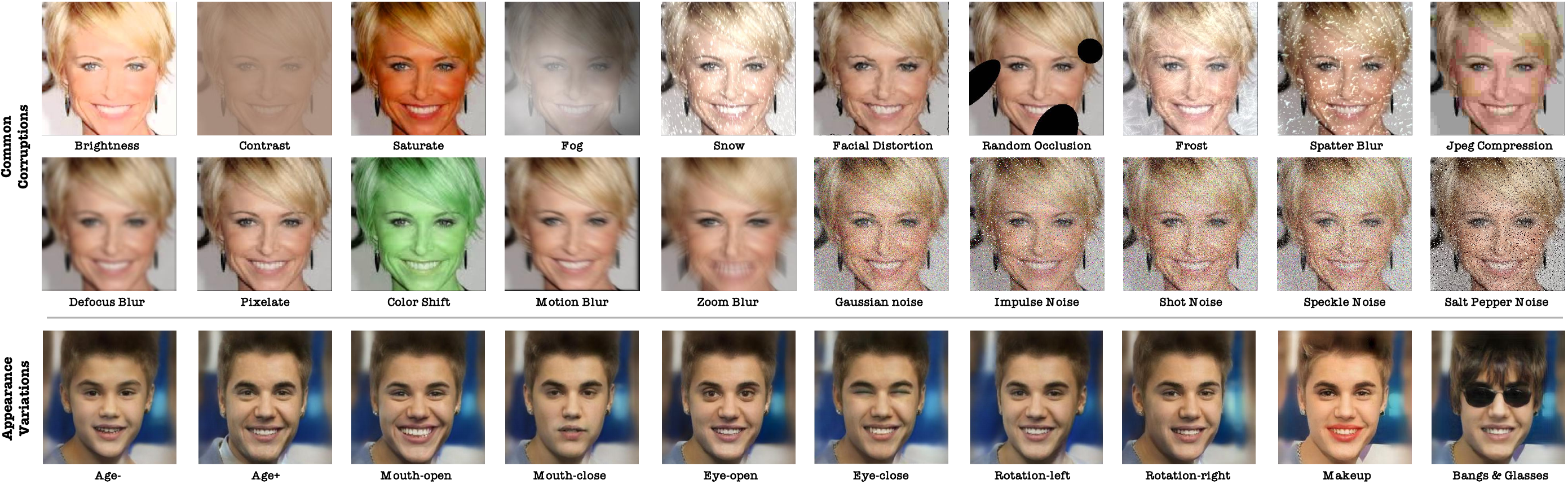}
    \vspace{-2ex}
   \caption{Visualization of 30 subcategories of common corruptions and appearance variations. More results are available in Appendix~\ref{sup:H}.}
    \label{fig:pixel_demo}
       \vspace{-3ex}
\end{figure*}

\vspace{-1ex}

\section{Related Work}
\subsection{Face Recognition}
Face recognition, a key task in computer vision, has made significant progress in recent years. Early models like DeepFace~\cite{taigman2014deepface} and FaceNet~\cite{schroff2015facenet} are the first to demonstrate the power of deep learning for FR tasks. Later, models such as SphereFace~\cite{liu2017sphereface}, CosFace~\cite{wang2018cosface}, and ArcFace~\cite{deng2019arcface}, which leverage angular margin-based loss functions, further enhanced feature discriminability. AdaFace~\cite{kim2022adaface} introduced an adaptive margin function to adjust sample weighting based on image quality, improving performance on low-quality datasets. TransFace~\cite{dan2023transface} focuses on FR by employing a patch-level augmentation strategy (DPAP) and an entropy-based hard sample mining (EHSM), which preserves facial structure while enhancing sample diversity. The latest TopoFR~\cite{dan2024topofr} model incorporates topology alignment and hard sample mining, effectively preserving data structure and improving generalization.

Most state-of-the-art FR models are trained on large datasets like MS-Celeb-1M~\cite{guo2016ms} and VGGFace2~\cite{cao2018vggface2}, achieving outstanding performance on benchmarks such as LFW~\cite{huang2008labeled}, CFP~\cite{sengupta2016frontal}, YTF~\cite{wolf2011face}, and IJB~\cite{klare2015pushing,whitelam2017iarpa,maze2018iarpa}, with accuracy often exceeding 99\%. Additionally, commercial FR APIs from providers like Tencent, Alibaba, and iFlytek also demonstrate high accuracy.


\subsection{Robustness Benchmarks}
It is well known that deep learning models lack robustness against adversarial samples~\cite{goodfellow2015explaining,szegedy2014intriguing}, common corruptions~\cite{hendrycks2019benchmarking,dong2023benchmarking}, and other types of distribution shifts~\cite{geirhos2019imagenet,geirhos2018generalisation,taori2020measuring}. In the field of FR, despite the introduction of many datasets such as LFW~\cite{huang2008labeled} and YTF~\cite{wolf2011face}, typical benchmarks mainly focus on FR under natural conditions. Some studies have proposed datasets to evaluate model robustness under various conditions, such as facial rotation (frontal-profile) in CFP~\cite{sengupta2016frontal} and age-related attributes in AgeDB~\cite{moschoglou2017agedb}. However, due to the high cost of collecting rare data, these datasets cover only a limited range of scenarios. Additionally, certain types of OOD face images are difficult to obtain due to their rarity and specific characteristics, such as sensor failure or age progression, posing challenges for real-world data collection.

A promising approach is to synthesize realistic OODs on clean datasets for benchmarking model robustness. For example, ImageNet-C~\cite{hendrycks2019benchmarking} introduced 15 types of corruptions for image classification, including noise, blur, weather, and digital artifacts. Similar methods have been applied to object detection~\cite{michaelis2019benchmarking}, point cloud recognition ~\cite{ren2022benchmarking,zheng2022benchmarking}, and autonomous driving~\cite{dong2023benchmarking}. Some works have explored robustness evaluation for FR, but they mainly focus on adversarial perturbations~\cite{yang2020robfr} or fairness issues~\cite{lin2024ai}. Given the diverse real-world applications of FR, building a comprehensive robustness evaluation benchmark remains a challenging task.
\vspace{-4ex}
\section{OOD in Face Recognition}
We introduce the categories of common corruptions and appearance variations in Sec.~\ref{sec:3.1} and Sec.~\ref{sec:3.2}, respectively, with more details provided in Appendix.~\ref{sup:A}. These include implementation details for corruptions and variations (\ref{sup:A1} \& \ref{sup:A2}), a comparison with related work (\ref{sup:A3}), and a discussion on the naturalness of OOD synthesis (\ref{sup:A4}).
 
\subsection{Common Corruptions}
\label{sec:3.1}

Real-world corruptions in FR arise from diverse application scenarios. Based on this, we classify corruptions into five categories: \textit{Lighting \& Weather, Sensor, Movement, Data \& Processing, and Occlusion}. Considering the varied environments of FR applications, we identify 20 distinct types of corruptions across these categories, as illustrated in Fig.~\ref{fig:category2}. We visualize examples of these corruptions in Fig.~\ref{fig:pixel_demo}.

\noindent\textbf{Lighting \& Weather:}
Changes in lighting conditions and complex weather are common in FR scenarios, such as variations in daylight, indoor/outdoor lighting, or adverse weather like fog and snow~\cite{ahsan2021evaluating}. These conditions can reduce image clarity, blur facial features, or introduce partial occlusions. We categorize these as \textit{brightness, contrast, saturation, fog, and snow}. Image enhancement techniques~\cite{hendrycks2019benchmarking} are used to simulate realistic weather and lighting effects.

\noindent\textbf{Sensor:}
Sensors can suffer from internal or external disturbances (e.g., sensor vibrations, lighting conditions~\cite{hendrycks2019benchmarking,li2022bevformer}, reflective surfaces), causing data degradation. Based on prior studies on sensor noise~\cite{hendrycks2019benchmarking,dong2023benchmarking}, we design three realistic sensor-level corruptions: \textit{defocus blur, color shift, and pixelation}. Defocus blur simulates the effect of an out-of-focus lens. Color shift alters the overall hue of the image, mimicking color bias due to sensor issues or ambient lighting. Pixelation reduces image detail by compressing it into larger pixel blocks, obscuring facial features.

\noindent\textbf{Movement:}
Motion is a common cause of image blur, stemming from camera movement or subject motion~\cite{dong2023benchmarking}. We introduce three motion-level corruptions: \textit{motion blur, zoom blur, and facial distortion}. Motion blur simulates blur from fast motion or camera shake. Zoom blur mimics rapid zooming. Facial distortion simulates unnatural deformation from quick facial movements during image capture.

\noindent\textbf{Data \& Processing:}
During processing, interference can degrade image quality~\cite{prasad2023systematic}. We simulate this with \textit{Gaussian noise, impulse noise, shot noise, speckle noise, salt-and-pepper noise, and JPEG compression}. 
These noises add random perturbations, simulating sensor noise during image acquisition and transmission. JPEG compression simulates image quality loss due to excessive compression.

\noindent\textbf{Occlusion:}
Occlusion includes \textit{random occlusion, frost, and spatter}. Random occlusion adds blocks of varying shapes and sizes, mimicking occlusions from objects like hands or hair. Frost, inspired by~\cite{hendrycks2019benchmarking}, simulates a white frost-like noise on the camera lens. Spatter simulates quality degradation from raindrops or mud splashes on the lens.

\begin{figure}[h]
  \centering
   \includegraphics[width=0.99\linewidth]{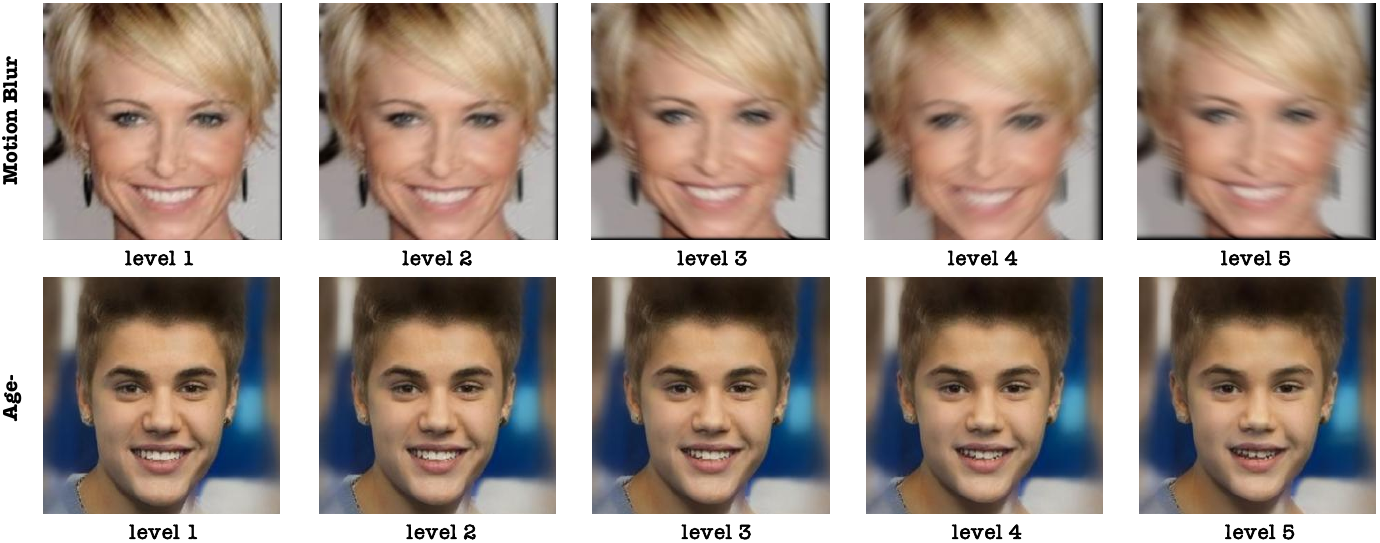}
    \vspace{-1ex}
   \caption{Visualization of severity levels. \textbf{Top:} Motion Blur from level 1 to 5; \textbf{Bottom:} Age- from level 1 to 5. Full visual results are available in Appendix~\ref{sup:H}.}
    \label{fig:pixel_level}
       \vspace{-4ex}
\end{figure}

\noindent\textbf{Corruption levels:}
We follow~\cite{hendrycks2019benchmarking} to define five severity levels for each type of corruption, as shown in Fig.~\ref{fig:pixel_level}. Using Motion Blur as an example, we visualize results from level 1 (mild) to level 5 (extreme).

\vspace{-2ex}

\subsection{Appearance Variations}
\label{sec:3.2}

Beyond common corruptions, we also consider semantic-level facial variations that may occur in daily life. These are categorized into four groups: \textit{Age, Facial Expression, Head Pose, and Accessories,} covering 10 unique types of changes as Fig.~\ref{fig:category2}. 
We provide visual examples in Fig.~\ref{fig:pixel_demo}.

\noindent\textbf{Age:}
Age is a key semantic variable, as facial features change noticeably over time~\cite{ling2007study}. Using generative facial aging techniques~\cite{roich2022pivotal}, we simulate facial changes across different age stages. These changes involve skin texture, looseness, wrinkles, and overall facial structure. 

\noindent\textbf{Facial Expression:}
Facial expression is another common appearance variation, especially under emotional fluctuations that affect the mouth and eye regions~\cite{dhavalikar2014face}. Using generative models~\cite{roich2022pivotal}, we edit facial expressions to simulate changes like smiling or frowning.

\noindent\textbf{Head Pose:}
Head pose (or viewpoint) frequently changes in real-life scenarios, particularly when faces are captured from various angles by surveillance cameras. Using editing algorithms~\cite{roich2022pivotal}, we modify head orientation to simulate different facial angles, from frontal to side profiles. 

\noindent\textbf{Accessory \& Makeup:}
In everyday life, people often wear various accessories that can obscure or alter facial features~\cite{neto2022beyond}. Based on~\cite{li2021image}, we add accessories to the images to simulate real-world accessories occlusions.
Makeup, especially common in social settings, can visually alter facial features by changing eyebrow shapes, eyeliner, blush, and lip color. We follow \cite{li2018beautygan} to add diverse makeup styles.
\noindent\textbf{Variation Levels.}
Similar to common corruptions, we define five severity levels for appearance variations (see Fig.~\ref{fig:pixel_level})
, with detailed level settings provided in Appendix~\ref{sup:A3}.

\noindent\textbf{Naturalness of OOD Synthesis.} Our synthesis method can be regarded as
closely approximating
real-world effects, with evaluation and analysis detailed in Appendix~\ref{sup:A4}.

\begin{table}[t]
\centering
   \scalebox{0.68}{
\setlength{\tabcolsep}{2pt}  
\begin{tabular}{c|c|cccc}
\hline
\multicolumn{1}{c}{\textbf{Models}} & \multicolumn{1}{c}{\textbf{Venue}} & \multicolumn{1}{c}{\textbf{Backbone}} & 
\multicolumn{1}{c}{\textbf{Loss}} & \multicolumn{1}{c}{\textbf{Para.(M)}} & \multicolumn{1}{c}{\textbf{Acc.(\%)}} \\
\hline
FaceNet~\cite{schroff2015facenet}                            & CVPR15                             & InceptionResNet                       & Triplet                           & 27.9                                 & 99.2                                  \\
SphereFace~\cite{liu2017sphereface}                         & CVPR17                             & Sphere20                              & A-Softmax                         & 28.1                                 & 98.2                                  \\
CosFace~\cite{wang2018cosface}                            & CVPR18                             & Sphere20                              & LMCL                              & 22.7                                 & 98.7                                  \\
ArcFace~\cite{deng2019arcface}                            & CVPR19                             & IR-SE50                               & Arcface                           & 43.8                                  & 99.5                                  \\
AdaFace~\cite{kim2022adaface}                             & CVPR22                             & IResNet50                             & AdaFace                          & 43.6                                 & 99.8                                  \\
ElasticFace~\cite{boutros2022elasticface}                         & CVPR22                             & IResNet100                             & ElasticFace                     & 65.2                                 & 99.8                                  \\
TransFace~\cite{dan2023transface}                           & ICCV23                             & ViT-S                                 & ArcFace+                         & 86.7                                  & 99.8                                  \\
TopoFR~\cite{dan2024topofr}                              & NeurIPS24                          & ResNet50                              & ArcFace+                         & 87.5                                 & 99.8                                  \\ \hline
MobileFace~\cite{chen2018mobilefacenets}                          & ECCV18                           & MobileFaceNet                           & LMCL                              & 1.2                                     & 99.5                                  \\
MobileNet~\cite{howard2017mobilenets}                           & CVPR17                             & MobileNet                             & LMCL                              & 3.8                                  & 99.4                                  \\
MobileNet-v2~\cite{sandler2018mobilenetv2}                         & CVPR18                            & MobileNetv2                            & LMCL                              & 2.9                                    & 99.3                                  \\
ShuffleNet~\cite{zhang2018shufflenet}                        & CVPR18                             & ShuffleNetV1                          & LMCL                              & 1.5                                  & 99.5                                  \\
ShuffleNet-v2~\cite{ma2018shufflenet}                        & ECCV18                             & ShuffleNetV2                          & LMCL                              & 1.8                                  & 99.2                                  \\
ResNet50~\cite{deng2019arcface}                            & CVPR16                             & ResNet50                              & LMCL                              & 40.3                                 & 99.7                                  \\ \hline
Softmax-IR~\cite{bishop2006pattern}                          & -                                  & IResNet50                             & Softmax                           & 43.6                                 & 99.6                                  \\
SphereFace-IR~\cite{liu2017sphereface}                       & -                                  & IResNet50                             & A-Softmax                         & 43.6                                 & 99.6                                  \\
AM-IR~\cite{wang2018additive}                               & -                                  & IResNet50                             & AM-Softmax                        & 43.6                                 & 99.2                                  \\
CosFace-IR~\cite{wang2018cosface}                          & -                                 & IResNet50                             & LMCL                              & 43.6                                  & 99.7                                  \\
ArcFace-IR~\cite{deng2019arcface}                          & -                                 & IResNet50                             & ArcFace                           & 43.6                                  & 99.7                                  \\ \hline
iFLYTEK API                         & -                                  & -                                     & -                                 & -                                     & 98.0                                  \\
Aliyun API                          & -                                  & -                                     & -                                 & -                                     & 99.7                                 \\
Tencent API                         & -                                  & -                                     & -                                 & -                                     & 99.8                          \\ \hline       
\end{tabular}
}
   \vspace{-1ex}
\caption{Tested FR models and commercial APIs.}
\label{tab:testing_models}
   \vspace{-4ex}
\end{table}

\vspace{-1ex}

\section{OOD Benchmarks}

\vspace{-1ex}

To thoroughly evaluate the robustness of FR models, we establish robustness benchmarks using widely adopted datasets: LFW~\cite{huang2008labeled}, CFP-FP~\cite{sengupta2016frontal}, and YTF~\cite{wolf2011face}. For each dataset, we create two benchmark versions: one for common corruptions (denoted as -C) and one for appearance variations (denoted as -V). 
Below, we detail the datasets, evaluation metrics, and FR models used in our benchmarks.

\vspace{-1ex}

\subsection{LFW-C/V}

We first conduct experiments using the LFW~\cite{huang2008labeled} dataset, a widely used benchmark in FR. LFW consists of 5,749 identities and 13,233 images, forming 6,000 pairs of face images. We preprocess the entire LFW dataset using MTCNN~\cite{zhang2016joint}, obtaining aligned images. We then apply 20 types of common corruptions and 10 types of appearance variations. Each includes five levels of severity.

\noindent\textbf{Corruption Dataset LFW-C.}
To comprehensively evaluate the robustness of FR models under various types of corruption. For each model, we first obtain its performance on the original LFW dataset, denoted as $\mathrm{Acc}_{clean}$. We then re-evaluate the model's performance under each corruption type \( c \) and severity level \( s \) in LFW-C, denoted as $\mathrm{Acc}_{c,s}$.

We calculate the average corruption robustness among 5 severity levels of the model using the following formula:

   \vspace{-2ex}

\begin{equation}
\small\label{eq:1}
\mathrm {Acc_{cor}} = \frac {1}{|\mathcal {C}|}\sum _{c\in \mathcal {C}} \frac {1}{5}\sum _{s=1}^{5}\mathrm {Acc}_{c,s}
\end{equation}

where \( c \) represents all corruption types. To further analyze the degradation of the model under each corruption, we introduce the \textit{Relative Corruption Error (RCE)}, which measures the percentage decrease in performance as follows:

   \vspace{-2ex}

\begin{equation}
\small\label{eq:2}
\mathrm {RCE}_{c,s}=\frac {\mathrm {Acc_{clean}}-\mathrm {Acc}_{c,s}}{\mathrm {Acc_{clean}}};\;\mathrm {RCE}=\frac {\mathrm {Acc_{clean}}-\mathrm {Acc_{cor}}}{\mathrm {Acc_{clean}}}
\end{equation}

\noindent\textbf{Variations Dataset LFW-V.}
Similarly, we obtain the $\mathrm{Acc}_{clean}$ and evaluate the model's performance under each variation category \( v \) and severity level \( s \), denoted as $\mathrm{Acc}_{v,s}$.

We calculate the average robustness of appearance variations for the model using the following equation:

   \vspace{-2ex}

\begin{equation}
\small\label{eq:3}
\mathrm {Acc_{var}} = \frac {1}{|\mathcal {V}|}\sum _{v\in \mathcal {V}} \frac {1}{5}\sum _{s=1}^{5}\mathrm {Acc}_{v,s}
\end{equation}

where \( v \) represents all appearance variations. To further analyze the performance variation of the model under each appearance variations condition, we introduce the \textit{Relative Variations Error (RVE)}, which measures the percentage decrease in performance as follows:

   \vspace{-2ex}

\begin{equation}
\small\label{eq:4}
\mathrm {RVE}_{v,s}=\frac {\mathrm {Acc_{clean}}-\mathrm {Acc}_{v,s}}{\mathrm {Acc_{clean}}};\;\mathrm {RVE}=\frac {\mathrm {Acc_{clean}}-\mathrm {Acc_{var}}}{\mathrm {Acc_{clean}}}
\end{equation}

   \vspace{-2ex}

\subsection{CFP-C/V}
The CFP~\cite{sengupta2016frontal} dataset consists of frontal and profile images of celebrities. It includes clear frontal and side-view face images, with a total of 500 identities and 7,000 pairs. Following a similar procedure as LFW, we first align all images using MTCNN~\cite{zhang2016joint}, then apply designed 20 types of common corruptions and 10 types of appearance variations.

\noindent\textbf{Corruption Dataset CFP-C.}
Similarly, CFP-C introduces 20 types of corruptions into the CFP validation set. These corruptions are applied with five levels of severity. For each model, we also first compute $\mathrm{Acc}_{clean}$, and evaluate it under each corruption type \( c \) and severity level \( s \), denoted as $\mathrm{Acc}_{c,s}$. We follow the Equation~\ref{eq:1} to calculate $\mathrm {Acc_{cor}}$ and Equation~\ref{eq:2} to compute the \textit{RCE} in a similar manner.

\noindent\textbf{Variations Dataset CFP-V.}
We also design CFP-V, a variations version based on the CFP, and calculate $\mathrm{Acc}_{clean}$, $\mathrm{Acc}_{v,s}$, $\mathrm {Acc_{var}}$, \textit{RVE} in a comparable format with LFW-V.


\subsection{YTF-C/V}

The YouTube Faces (YTF)~\cite{wolf2011face} dataset comprises 3,425 YouTube videos from 1,595 subjects (a subset of LFW celebrities). The videos range from 48 to 6,070 frames in length and include 5,000 video pairs and 10 splits.
Following~\cite{yang2020robfr}, we extract the central frame from each video, selecting 5,000 pairs. These images are then preprocessed using MTCNN~\cite{zhang2016joint}. We apply our 30 designed complex scenarios, each with five severity levels~\cite{hendrycks2019benchmarking}.

Similarly, to evaluate the robustness of FR models against common corruptions and appearance variations, we design two benchmark versions: YTF-C and YTF-V, with the calculation for evaluation metrics being similar to those used in LFW-C/V and CFP-C/V.

\subsection{Testing FR Models}
\label{sec:4.4}

Following the setup in~\cite{yang2020robfr}, to evaluate the robustness of FR systems, we select 19 state-of-the-art FR models, 
as summarized in Tab.~\ref{tab:testing_models}. First, we include the open-source models, such as~\cite{schroff2015facenet,kim2022adaface,boutros2022elasticface,dan2023transface,dan2024topofr}. Second, to investigate the impact of model architecture, we include several models across different backbones, varying in weight size. These include mainstream architectures like ResNet50~\cite{he2016deep} and lightweight networks such as MobileFace~\cite{chen2018mobilefacenets}, MobileNet~\cite{howard2017mobilenets}, MobileNet-v2~\cite{sandler2018mobilenetv2}, ShuffleNet~\cite{zhang2018shufflenet}, and ShuffleNet-v2~\cite{ma2018shufflenet}, all trained with LMCL~\cite{wang2018cosface} loss. Additionally, we examine the effect of different loss functions by including models based on the same architecture (IResNet50~\cite{deng2019arcface}), optimized using distance-based loss (e.g., Softmax~\cite{bishop2006pattern}) and angular margin-based losses (e.g., SphereFace~\cite{liu2017sphereface}, AM-Softmax~\cite{wang2018additive}, CosFace~\cite{wang2018cosface}, and ArcFace~\cite{deng2019arcface}). To further enrich the evaluation, we also include three commercial API services, though their underlying mechanisms and training data remain unknown.

\begin{table*}[t]
\centering
   \scalebox{0.47}{
\setlength{\tabcolsep}{2pt}  
\begin{tabular}{c|c|cccccccc|cccccc|ccccc}
\hline
\multicolumn{2}{c}{Models}                               & \multicolumn{8}{c}{Open-source Model Eval}                                                                                & \multicolumn{6}{c}{Architecture Eval}                                                                         & \multicolumn{5}{c}{Loss Function Eval}                                     \\
\multicolumn{2}{c}{Corruptions}                          & FaceNet        & SphereFace & CosFace & ArcFace        & ElasticFace& AdaFace        & TransFace      & TopoFR & MobileFace     & Mobilenet & Mobilenet-v2 & ShuffleNet& ShuffleNet-v2& ResNet50       & Softmax-IR & SphereFace-IR  & Am-IR          & CosFace-IR     & ArcFace-IR     \\ \hline\hline
\rowcolor{lightgray}\multicolumn{2}{c}{None (clean)}                         & 99.23          & 98.20      & 98.63   & 99.50          & 99.80            & \textbf{99.83} & 99.75          & 99.78       & 99.43          & 99.40     & 99.10       & 99.43                  & 99.15                          & \textbf{99.72} & 99.53      & 99.57          & 99.18          & \textbf{99.70} & 99.67          \\ \hline
\multirow{5}{*}{Lighting \& Weather} & Brightness        & 98.20          & 95.68      & 96.96   & 99.05          & 99.70            & \textbf{99.73} & 99.58          & 99.50       & 98.86          & 98.29     & 97.78       & 98.76                  & 98.02                          & \textbf{99.19} & 99.08      & 99.17          & 98.08          & \textbf{99.49} & 99.42          \\
                                     & Contrast          & 84.64          & 84.01      & 84.27   & 92.92          & 95.51            & \textbf{99.10} & 96.93          & 94.75       & 87.26          & 84.07     & 83.93       & 86.22                  & 83.81                          & \textbf{89.61} & 89.92      & \textbf{90.38} & 88.74          & 89.60          & 89.24          \\
                                     & Saturate          & 98.59          & 96.81      & 97.86   & 98.20          & 98.87            & \textbf{99.70} & 99.20          & 98.95       & 99.07          & 98.80     & 98.51       & 99.04                  & 98.60                          & \textbf{99.38} & 99.33      & 99.32          & 98.59          & \textbf{99.56} & 99.49          \\
                                     & Fog               & 93.25          & 86.10      & 88.30   & 92.56          & \textbf{93.38}   & 91.02          & 93.10          & 93.02       & \textbf{89.69} & 83.59     & 85.45       & 88.42                  & 87.44                          & 88.61          & 90.43      & 89.41          & 89.21          & \textbf{92.30} & 92.18          \\
                                     & Snow              & 90.96          & 87.23      & 91.84   & 96.53          & 97.48            & \textbf{98.83} & 97.87          & 96.80       & \textbf{93.86} & 89.99     & 91.10       & 92.61                  & 92.58                          & 93.06          & 94.92      & 93.97          & 94.66          & 96.09          & \textbf{96.25} \\
                                     \hline
\multirow{3}{*}{Sensor}              & Defocus Blur      & \textbf{94.30} & 79.71      & 82.12   & 88.31          & 87.98            & 87.34          & 88.70          & 85.53       & 86.16          & 84.78     & 85.29       & \textbf{87.30}         & 86.29                          & 86.73          & 87.65      & 86.18          & 88.67          & 88.98          & \textbf{89.10} \\
                                     & Color Shift       & 98.81          & 97.11      & 98.22   & 99.45          & 99.80            & \textbf{99.81} & 99.78          & 99.73       & 99.24          & 99.14     & 98.81       & 99.23                  & 98.84                          & \textbf{99.54} & 99.45      & 99.49          & 98.86          & 99.62          & \textbf{99.62} \\
                                     & Pixelate          & 98.83          & 94.42      & 96.21   & 98.46          & 99.51            & \textbf{99.80} & 99.23          & 98.61       & 98.29          & 97.52     & 96.99       & 98.51                  & 98.16                          & \textbf{99.09} & 98.95      & 98.76          & 97.98          & 99.21          & \textbf{99.21} \\
                                     \hline
\multirow{3}{*}{Movement}            & Motion Blur       & \textbf{96.41} & 87.67      & 89.28   & 95.04          & 95.59            & 94.09          & 96.36          & 93.56       & 92.41          & 91.06     & 91.72       & \textbf{93.43}         & 92.75                          & 93.25          & 94.21      & 93.89          & 93.66          & \textbf{95.26} & 95.25          \\
                                     & Zoom Blur         & 97.69          & 96.62      & 97.40   & 98.77          & 99.49            & \textbf{99.58} & 99.26          & 99.12       & 98.95          & 98.25     & 97.81       & 98.73                  & 98.36                          & \textbf{99.30} & 99.07      & 99.09          & 98.32          & \textbf{99.41} & 99.38          \\
                                     & Facial Distortion & 94.89          & 79.58      & 84.62   & 93.60          & 91.84            & \textbf{95.28} & 93.33          & 92.24       & 92.21          & 87.67     & 88.68       & \textbf{92.25}         & 91.29                          & 90.98          & 92.46      & 91.66          & 92.78          & 93.97          & \textbf{94.34} \\
                                     \hline
\multirow{6}{*}{Data \& Processing}  & Gaussian Noise    & 87.84          & 73.13      & 80.82   & \textbf{93.21} & 87.19            & 87.68          & 88.40          & 82.18       & \textbf{84.66} & 75.69     & 77.30       & 76.73                  & 82.43                          & 84.10          & 81.77      & 80.40          & \textbf{89.23} & 88.70          & 87.05          \\
                                     & Impulse Noise ️   & 89.02          & 72.90      & 81.90   & \textbf{94.65} & 90.10            & 90.27          & 90.03          & 84.25       & \textbf{85.94} & 77.41     & 78.37       & 77.02                  & 82.09                          & 84.42          & 83.41      & 82.81          & 90.97          & \textbf{91.06} & 88.90          \\
                                     & Shot Noise        & 84.80          & 70.20      & 77.65   & \textbf{93.02} & 86.67            & 86.70          & 87.46          & 81.55       & 81.53          & 72.96     & 73.98       & 73.77                  & 78.08                          & \textbf{81.66} & 78.61      & 77.63          & \textbf{87.21} & 86.42          & 84.48          \\
                                     & Speckle Noise     & 90.67          & 74.61      & 83.65   & \textbf{95.90} & 93.53            & 94.23          & 92.99          & 89.79       & 87.61          & 79.00     & 79.45       & 79.97                  & 83.66                          & \textbf{88.01} & 86.27      & 84.54          & 91.36          & \textbf{92.25} & 90.41          \\
                                     & Salt Pepper Noise & 76.74          & 62.82      & 66.06   & \textbf{88.90} & 82.51            & 87.13          & 79.20          & 70.70       & \textbf{76.15} & 64.35     & 66.89       & 59.54                  & 63.54                          & 69.04          & 64.73      & 65.23          & \textbf{82.50} & 78.36          & 74.51          \\
                                     & Jpeg Compression  & 98.67          & 95.44      & 97.05   & 98.92          & \textbf{99.49}   & 99.48          & 99.47          & 99.24       & 98.67          & 98.36     & 98.11       & 98.57                  & 97.91                          & \textbf{99.14} & 98.96      & 98.88          & 98.12          & 99.32          & \textbf{99.33} \\
                                     \hline
\multirow{3}{*}{Occlusion}           & Random Occlusion  & 93.74          & 86.13      & 89.07   & 94.58          & 97.36            & \textbf{98.43} & 97.38          & 98.10       & 91.73          & 88.89     & 87.63       & 90.94                  & 91.21                          & \textbf{93.24} & 94.05      & 93.56          & 92.73          & \textbf{96.10} & 95.22          \\
                                     & Frost             & 90.93          & 83.24      & 89.98   & 94.50          & 94.29            & \textbf{96.88} & 94.36          & 93.94       & \textbf{93.04} & 88.73     & 89.52       & 92.17                  & 91.54                          & 92.44          & 93.36      & 92.09          & 93.33          & 94.83          & \textbf{94.99} \\
                                     & Spatter           & 91.25          & 89.20      & 92.26   & 96.60          & 98.27            & 98.82          & \textbf{98.93} & 98.30       & \textbf{94.70} & 90.88     & 92.36       & 93.62                  & 92.71                          & 93.87          & 96.87      & 96.80          & 95.69          & \textbf{98.02} & 97.64          \\
                                     \hline
\rowcolor{lightgray}\multicolumn{2}{c}{Average}                              & 92.51          & 84.63      & 88.28   & 95.16          & 94.43            & \textbf{95.20} & 94.58          & 92.49       & \textbf{91.50} & 87.47     & 87.98       & 88.84                  & 89.47                          & 91.23          & 91.17      & 90.66          & 93.03          & \textbf{93.93} & 93.30   \\
\hline
\end{tabular}
}
\vspace{-1ex}
\caption{Accuracy of 19 FR models on LFW-C, categorized into \textit{Open-source Model Eval}, \textit{Architecture Eval} and \textit{Loss Function Eval}.}
\label{tab:accuracy_LFW-C}
\vspace{-3ex}
\end{table*}

\vspace{-1ex}
\section{Benchmarking Results and Insights}
\label{sec:5}
We present the evaluation results on LFW-C in Sec.~\ref{sec:5.1}, and LFW-V in Sec.~\ref{sec:5.2}, and leave the results on CFP-C/V and YTF-C/V in Appendix~\ref{sup:C} and Appendix~\ref{sup:D}. In Sec.~\ref{sec:5.3}, we discuss commercial API evaluation, followed by extended physical experiments on face masks in Sec.~\ref{sec:5.4}.

\vspace{-1ex}

\subsection{Common Corruptions Evaluation Results}
\label{sec:5.1}
Tab.~\ref{tab:accuracy_LFW-C} shows the robustness evaluation of 19 FR models on LFW-C, categorized into \textit{Open-source Model Eval}, \textit{Architecture Eval}, and \textit{Loss Function Eval}. We report the average performance across five levels for each category. It is evident that model robustness does not strongly correlate with $\mathrm{Acc}_{clean}$. For example, models with high $\mathrm{Acc}_{clean}$ (e.g., TopoFR~\cite{dan2024topofr}) does not achieve the high $\mathrm {Acc_{cor}}$. In Fig.~\ref{fig:RCE_lfw}, we further illustrate the \textit{Relative Corruption Error (RCE)} for each model across corruption categories. Based on these evaluations, we provide the following analysis:

\begin{figure}[th]
  \centering
   \includegraphics[width=1.0\linewidth]{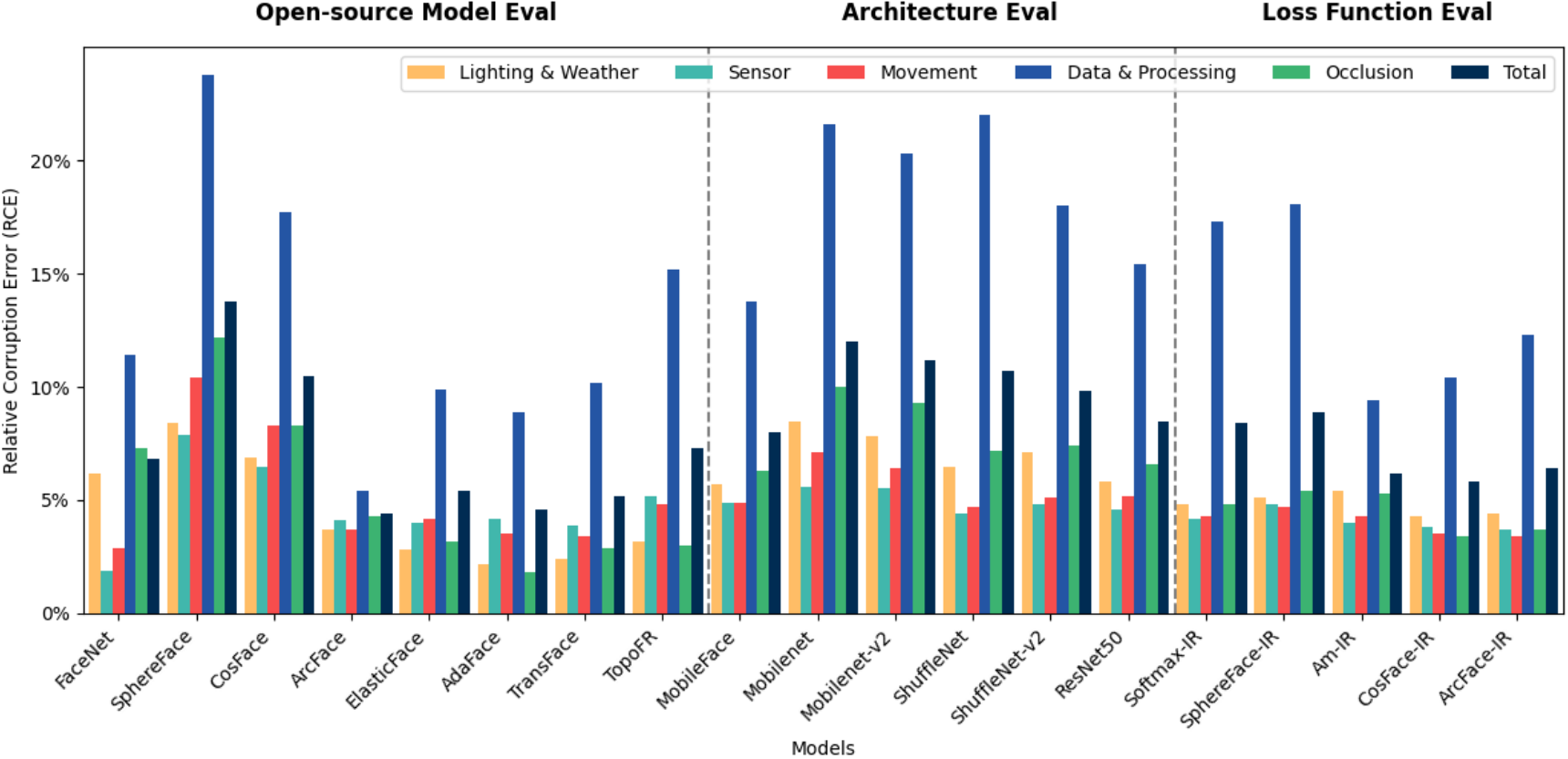}
    \vspace{-5ex}
   \caption{RCE results on LFW-C.}
   \label{fig:RCE_lfw}
   \vspace{-4ex}
\end{figure}

\noindent\textbf{Comparison of Corruption Types.}
As shown in Tab.~\ref{tab:accuracy_LFW-C}  and Fig.~\ref{fig:RCE_lfw}, all corruption types lead to performance degradation in FR models. Among these, \textit{Data \& Processing} has the highest RCE exceeding 20\%. Additionally, \textit{Occlusion} causes substantial performance drops, likely due to the loss of key facial features.
 On the other hand, most models exhibit negligible degradation under \textit{Sensor Corruptions} (e.g., \textit{color shift, defocus blur}), possibly because these corruptions are partially present in natural face datasets.

\vspace{-1ex}
 \begin{tcolorbox}[colframe=black, colback=gray!8, coltitle=black, sharp corners=all, boxrule=0.5mm, boxsep=0.1mm]
   \textbf{Insight 1:} 
   \textbf{FR models suffer significant performance degradation under corruptions, with \textit{Data \& Processing} causing the most severe impact.} This occurs as noise disrupts the feature space distribution, shifting the decision boundary of linear classifiers, and the destructiveness of noise corruption also stems from the model’s over-reliance on high-frequency phase, which is linked to the frequency response characteristics of FR architectures, leading to degraded cross-layer feature fusion. (Virtualized in Appendix Fig.~\ref{fig:tsne})
\end{tcolorbox}

\noindent\textbf{Comparison of FR Models.}
While all models experience performance decline, different models show varying sensitivities to specific corruptions. 
For instance, FaceNet~\cite{chen2018mobilefacenets} demonstrates robustness against \textit{Sensor} and \textit{Motion} corruptions.
ArcFace~\cite{kim2022adaface} achieves the highest accuracy in \textit{Data \& Processing}, benefiting from its Angular Margin Loss, which provides robust features even under low-resolution or noisy sensor data.
Meanwhile, AdaFace~\cite{kim2022adaface} achieves the best average FR accuracy across all corruptions, attributed to its adaptive margin loss function.

\noindent\textbf{Comparison of Architecture and Loss Function.}
MobileFace~\cite{chen2018mobilefacenets} achieves the best average robustness with the fewest parameters (1.2M), particularly excelling in the \textit{Data \& Processing} and \textit{Occlusion}. ResNet50~\cite{he2016deep} ranks second in average robustness, showing strong performance in the \textit{Lighting \& Weather}, and \textit{Sensor} categories.
Among the loss functions, LMCL~\cite{wang2018cosface} achieves the best average results.
This can be attributed to its optimization of inter-class boundaries in the feature space, enhancing discrimination for similar samples. Additionally, AM-Softmax~\cite{wang2018additive} shows superior robustness in the \textit{Data \& Processing} category.

\vspace{-1ex}
  \begin{tcolorbox}[colframe=black, colback=gray!8, coltitle=black, sharp corners=all, boxrule=0.5mm, boxsep=0.1mm]
   \textbf{Insight 2:} 
   \textbf{Different models exhibit varying sensitivity to OOD scenarios, with some being robust to specific OOD types while vulnerable to others.} This highlights the critical role of architecture, loss functions, and training procedures in shaping model performance under different conditions.

   \begin{itemize}
    \item Loss functions induce distinct decision boundary geometries. For instance, AdaFace leverages spherical compression with adaptive feature norm adjustment, mitigating the local perturbation sensitivity observed in losses like Triplet Loss.
    \item  Mainstream datasets are curated with limited diversity, lacking spatio-temporal noise modeling in real-world conditions. The feature representation capacity of models is constrained by data quality.

   \end{itemize}

\end{tcolorbox}

\vspace{-2ex}

\begin{table*}[t]
\centering
 \scalebox{0.48}{
 \setlength{\tabcolsep}{2pt}  
\begin{tabular}{c|c|cccccccc|cccccc|ccccc}
\hline
\multicolumn{2}{c}{Models}                           & \multicolumn{8}{c}{Open-source Model Eval}                                                                                                                                        & \multicolumn{6}{c}{Architecture Eval}                                                                     & \multicolumn{5}{c}{Loss Function Eval}                           \\
\multicolumn{2}{c}{Variations}                          & FaceNet        & SphereFace & CosFace & ArcFace        & ElasticFace& AdaFace        & TransFace      & TopoFR & MobileFace     & Mobilenet & Mobilenet-v2 & ShuffleNet& ShuffleNet-v2& ResNet50       & Softmax-IR & SphereFace-IR  & Am-IR          & CosFace-IR     & ArcFace-IR     \\ \hline\hline
\rowcolor{lightgray} \multicolumn{2}{c}{None (clean)}                    & 99.23   & 98.20      & 98.63   & 99.50   & 99.80            & \textbf{99.83} & 99.75     & 99.78       & 99.43      & 99.40     & 99.10       & 99.43                  & 99.15                          & \textbf{99.72} & 99.53      & 99.57         & 99.18 & \textbf{99.70} & 99.67          \\
\hline
\multirow{2}{*}{Age}               & Age-           & 95.16   & 94.13      & 94.24   & 95.91   & 96.32            & \textbf{96.48} & 96.12     & 96.08       & 95.85      & 95.24     & 95.02       & 95.60                  & 95.11                          & \textbf{95.94} & 95.74      & 95.78         & 95.27 & 96.10          & \textbf{96.16} \\
                                   & Age+           & 95.22   & 93.85      & 94.67   & 96.00   & 96.39            & \textbf{96.60} & 96.39     & 96.19       & 95.86      & 95.52     & 95.34       & 95.88                  & 95.39                          & \textbf{96.16} & 96.14      & 96.13         & 95.43 & 96.30          & \textbf{96.42} \\
                                   \hline
\multirow{4}{*}{Facial Expression} & Mouth-close    & 95.68   & 94.42      & 94.89   & 96.06   & 96.37            & \textbf{96.69} & 96.39     & 96.26       & 96.02      & 95.68     & 95.53       & 95.97                  & 95.54                          & \textbf{96.34} & 96.27      & 96.24         & 95.63 & \textbf{96.41} & 96.36          \\
                                   & Mouth-open     & 95.57   & 94.25      & 94.86   & 96.06   & 96.48            & \textbf{96.65} & 96.53     & 96.26       & 95.95      & 95.83     & 95.35       & 96.01                  & 95.60                          & \textbf{96.34} & 96.21      & 96.11         & 95.51 & 96.31          & \textbf{96.32} \\
                                   & Eye-close      & 94.91   & 94.09      & 94.32   & 95.87   & 96.48            & \textbf{96.61} & 96.48     & 96.31       & 95.91      & 95.47     & 95.17       & 95.85                  & 95.45                          & \textbf{96.22} & 96.03      & 95.97         & 95.37 & 96.30          & \textbf{96.32} \\
                                   & Eye-open       & 95.64   & 94.51      & 95.03   & 96.13   & 96.57            & \textbf{96.71} & 96.54     & 96.45       & 96.01      & 95.67     & 95.36       & 96.03                  & 95.49                          & \textbf{96.29} & 96.20      & 96.26         & 95.58 & \textbf{96.44} & 96.36          \\
                                   \hline
\multirow{2}{*}{Rotation}          & Rotation-left  & 95.95   & 94.99      & 95.35   & 96.20   & 96.65            & \textbf{96.79} & 96.61     & 96.54       & 96.21      & 96.09     & 95.77       & 96.17                  & 95.77                          & \textbf{96.50} & 96.37      & 96.36         & 95.80 & \textbf{96.52} & 96.51          \\
                                   & Rotation-right & 95.86   & 94.76      & 95.32   & 96.27   & 96.63            & \textbf{96.77} & 96.60     & 96.39       & 96.28      & 96.10     & 95.69       & 96.25                  & 95.87                          & \textbf{96.53} & 96.39      & 96.47         & 95.65 & 96.60          & \textbf{96.60} \\
\multirow{2}{*}{Accessories}       & Bangs\&Glasses & 94.80   & 93.15      & 94.88   & 98.04   & 99.13            & \textbf{99.31} & 98.85     & 98.91       & 97.64      & 95.46     & 95.03       & 97.17                  & 96.43                          & \textbf{97.89} & 97.89      & 98.01         & 96.71 & \textbf{98.78} & 98.73          \\
                                   & Makeup         & 98.32   & 96.60      & 97.42   & 98.97   & 99.70            & \textbf{99.70} & 99.54     & 99.58       & 99.02      & 98.55     & 98.32       & 98.86                  & 98.38                          & \textbf{99.29} & 99.16      & 99.29         & 98.50 & \textbf{99.46} & 99.42          \\
                                   \hline
\rowcolor{lightgray} \multicolumn{2}{c}{Average}                         & 96.03   & 94.81      & 95.42   & 96.82   & 97.32            & \textbf{97.47} & 97.26     & 97.16       & 96.74      & 96.27     & 95.97       & 96.66                  & 96.20                          & \textbf{97.02} & 96.90      & 96.92         & 96.24 & \textbf{97.17} & \textbf{97.17} \\
\hline
\end{tabular}
}
\caption{Accuracy of 19 FR models on LFW-V, categorized into \textit{Open-source Model Eval}, \textit{Architecture Eval} and \textit{Loss Function Eval}.}
\label{tab:accuracy_LFW-v}
\end{table*}

\subsection{Appearance Variations Evaluation Results}
\label{sec:5.2}

\begin{figure}[bh]
  \centering
   \includegraphics[width=1.0\linewidth]{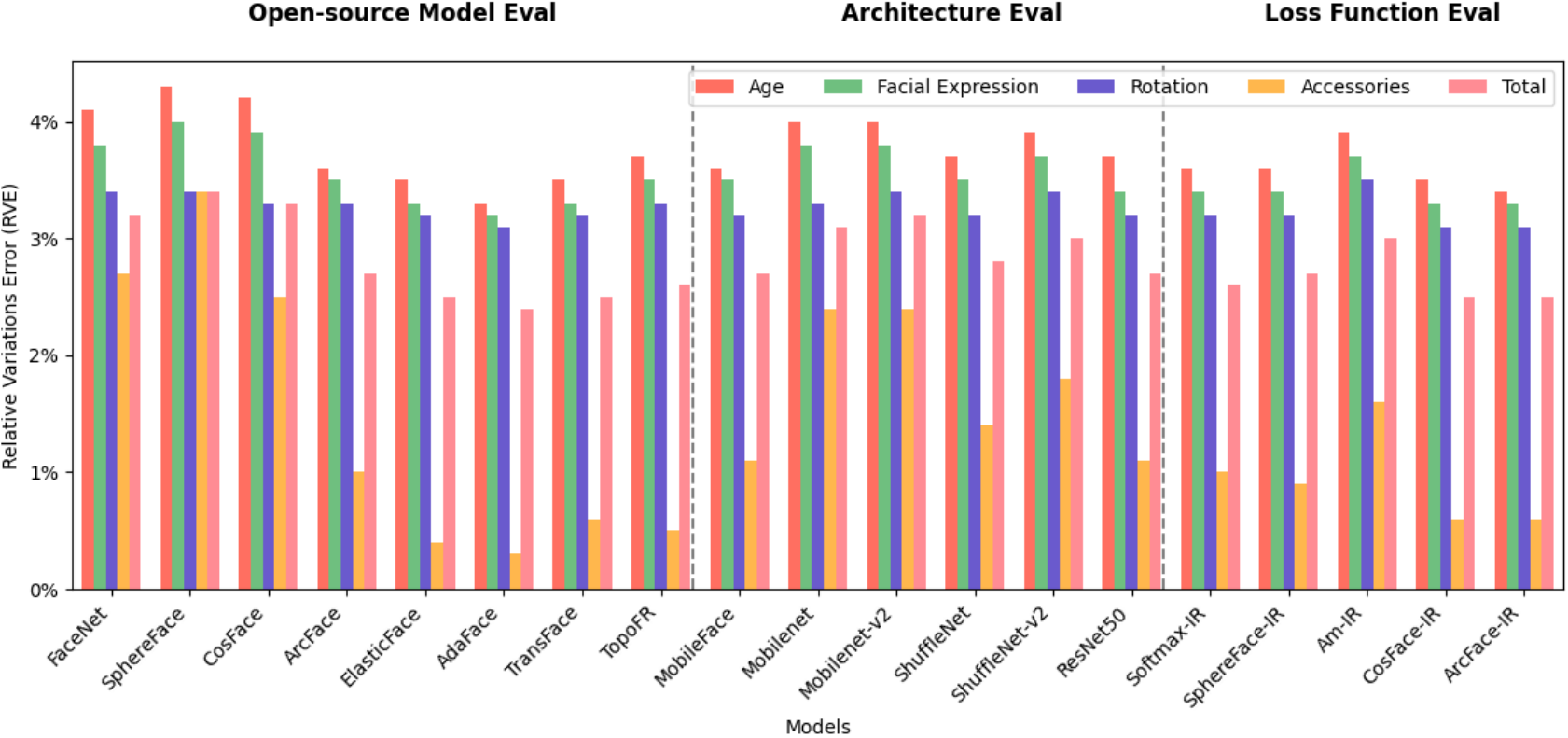}
   \caption{RVE results on LFW-V.}
   \label{fig:RVE_lfw}
\end{figure}

\noindent\textbf{Comparison of Variations Types.}
As shown in Tab.~\ref{tab:accuracy_LFW-v}  and Fig.~\ref{fig:RVE_lfw}, appearance variations also degrade model performance, but their impact is generally less severe than corruptions, with the highest RVE around 4\%. Unlike corrupted data, FR models are trained with data that includes redundancy for changes such as age, accessories, and others.
Additionally, \textit{Accessories} exhibit the least overall degradation, likely because variations such as makeup are superficial and do not significantly alter key facial structures. In contrast, \textit{Age} variations induce the most substantial impact, as aging introduces prominent facial changes, including increased wrinkles and skin laxity.


\noindent\textbf{Comparison of FR Models.}
While models generally maintain high recognition rates under variations, AdaFace~\cite{kim2022adaface} achieves the highest average accuracy among open-source models and ranks best across all 10 appearance variations subcategories. This performance can be attributed to the adaptive margin loss function of AdaFace~\cite{kim2022adaface}, which targets difficult samples, as well as its high clean accuracy.

\vspace{-1ex}

  \begin{tcolorbox}[colframe=black, colback=gray!8, coltitle=black, sharp corners=all, boxrule=0.5mm, boxsep=0.1mm]
   \textbf{Insight 3:} 
   \textbf{Performance degradation under Appearance Variations is mild.} This is likely due to inherent redundancy in training data, which enhances robustness to facial variations. Models maintain high accuracy on accessory changes, contrasting with their vulnerability to age variations, indicating insufficient modeling of longitudinal facial changes.
\end{tcolorbox}

\noindent\textbf{Model Architecture Comparison.}
Among the architectures, ResNet50~\cite{he2016deep} achieves the best results across all subcategories. This may be due to the deeper network's stronger feature learning capability, allowing it to effectively capture global features in appearance changes.

\vspace{-1ex}

\noindent\textbf{Loss Function Comparison.}
CosFace-IR and ArcFace-IR are tied as the top performers, achieving an accuracy of 97.17\%. 
Their high clean accuracy suggests that these models effectively learn facial representations, enabling them to maintain capabilities when facing appearance changes.

\vspace{-1ex}
  \begin{tcolorbox}[colframe=black, colback=gray!8, coltitle=black, sharp corners=all, boxrule=0.5mm, boxsep=0.1mm]
   \textbf{Insight 4:} 
   \textbf{While Corruption robustness shows weak correlation with clean accuracy, Variation robustness exhibits strong dependence.}  
      \begin{itemize}
    \item The decoupling of corruption resilience from clean performance likely stems from architectural limitations in handling structured noise.
    \item The high correlation for variations may stem from a domain adaptation effect, where models excel by enhancing feature disentanglement via margin-based losses like LMCL.
   \end{itemize}
\end{tcolorbox}

\vspace{-3ex}

\begin{table}[]
\centering
 \scalebox{0.5}{
  \setlength{\tabcolsep}{2pt}  
\begin{tabular}{c|c|ccc|ccc|ccc}
\hline
\multicolumn{2}{c}{Models}                               & \multicolumn{3}{c}{Rejection Rate}                                          & \multicolumn{3}{c}{Accepted Samples Accuracy}                                          & \multicolumn{3}{c}{Actual Accuracy}                                                    \\
\multicolumn{2}{c}{Corruptions}                          & \multicolumn{1}{c}{Aliyun} & \multicolumn{1}{c}{iFLYTEK} & \multicolumn{1}{c}{Tencent} & \multicolumn{1}{c}{Aliyun} & \multicolumn{1}{c}{iFLYTEK} & \multicolumn{1}{c}{Tencent} & \multicolumn{1}{c}{Aliyun} & \multicolumn{1}{c}{iFLYTEK} & \multicolumn{1}{c}{Tencent} \\
\hline \hline
\rowcolor{lightgray} \multicolumn{2}{c}{None (clean)}                         & 0.00                        & 0.00                         & 0.00                         & 99.65                      & 97.99                       & \textbf{99.75}              & 99.65                      & 97.99                       & \textbf{99.75}              \\
\hline
\multirow{5}{*}{L \& W} & Brightness        & 2.02                       & \textbf{0.23}               & 0.67                        & \textbf{99.73}             & 96.48                       & 99.60                        & 97.71                      & 96.25                       & \textbf{98.93}              \\
                                     & Contrast          & 35.01                      & 17.91                       & \textbf{4.58}               & \textbf{99.79}             & 96.27                       & 99.33                       & 64.86                      & 79.03                       & \textbf{94.78}              \\
                                     & Saturate          & 0.43                       & \textbf{0.12}               & 0.13                        & \textbf{99.82}             & 97.56                       & 99.72                       & 99.39                      & 97.45                       & \textbf{99.59}              \\
                                     & Fog               & 44.27                      & \textbf{7.36}               & 17.11                       & \textbf{99.22}             & 92.15                       & 97.60                        & 55.29                      & \textbf{85.37}              & 80.90                       \\
                                     & Snow              & 24.12                      & \textbf{12.68}              & 14.38                       & \textbf{99.43}             & 93.32                       & 98.75                       & 75.45                      & 81.49                       & \textbf{84.55}              \\
                                     \hline
\multirow{3}{*}{Sensor}              & Defocus Blur      & 9.55                       & \textbf{1.20}               & 5.60                        & \textbf{98.58}             & 88.56                       & 96.79                       & 89.17                      & 87.49                       & \textbf{91.37}              \\
                                     & Color Shift       & 7.59                       & \textbf{0.38}               & 0.85                        & \textbf{99.78}             & 97.31                       & 99.71                       & 92.20                      & 96.94                       & \textbf{98.86}              \\
                                     & Pixelate          & 0.55                       & \textbf{0.02}               & 0.08                        & \textbf{99.75}             & 96.45                       & 99.63                       & 99.20                      & 96.43                       & \textbf{99.55}              \\
                                     \hline
\multirow{3}{*}{Movement}            & Motion Blur       & 4.26                       & \textbf{0.60}               & 1.20                        & \textbf{99.39}             & 93.62                       & 98.90                        & 95.15                      & 93.06                       & \textbf{97.71}              \\
                                     & Zoom Blur         & 1.12                       & \textbf{0.03}               & 2.76                        & \textbf{99.70}              & 96.54                       & 99.36                       & \textbf{98.58}             & 96.51                       & 96.62                       \\
                                     & Facial Distortion & 12.71                      & 13.45                       & \textbf{7.84}               & \textbf{98.03}             & 87.96                       & 96.81                       & 85.57                      & 76.13                       & \textbf{89.22}              \\
                                     \hline
\multirow{6}{*}{D \& P}  & Gaussian Noise    & 89.96                      & 71.75                       & \textbf{48.87}              & \textbf{99.17}             & 91.65                       & 95.94                       & 9.95                       & 25.89                       & \textbf{49.05}              \\
                                     & Impulse Noise ️   & 84.01                      & 58.37                       & \textbf{33.47}              & \textbf{99.37}             & 93.21                       & 97.01                       & 15.89                      & 38.80                       & \textbf{64.54}              \\
                                     & Shot Noise        & 93.74                      & 84.76                       & \textbf{63.05}              & \textbf{97.06}             & 88.04                       & 93.62                       & 6.07                       & 13.41                       & \textbf{34.59}              \\
                                     & Speckle Noise     & 81.10                      & 72.85                       & \textbf{49.06}              & \textbf{99.12}             & 91.31                       & 96.29                       & 18.73                      & 24.79                       & \textbf{49.05}              \\
                                     & Salt Pepper Noise & 100.00                     & 98.95                       & \textbf{86.03}              & 0.00                          & \textbf{90.48}              & 66.59                       & 0.00                       & 0.95                        & \textbf{9.30}               \\
                                     & Jpeg Compression  & 0.54                       & \textbf{0.03}               & 0.13                        & \textbf{99.71}             & 96.30                       & \textbf{99.56}              & 99.18                      & 96.27                       & \textbf{99.43}              \\
                                     \hline
\multirow{3}{*}{Occlusion}           & Random Occlusion  & 39.91                      & \textbf{29.54}              & 40.36                       & 97.86                      & 86.87                       & 98.77                       & 58.81                      & \textbf{61.21}              & 58.91                       \\
                                     & Frost             & 62.80                      & 39.99                       & \textbf{19.35}              & \textbf{99.06}             & 92.31                       & 97.95                       & 36.85                      & 55.40                       & \textbf{79.00}              \\
                                     & Spatter           & 12.49                      & 3.36                        & \textbf{3.18}               & \textbf{99.73}             & 94.65                       & 99.33                       & 87.27                      & 91.47                       & \textbf{96.17}              \\
                                     \hline
\rowcolor{lightgray} \multicolumn{2}{c}{Average}                              & 35.31                      & 25.68                       & \textbf{19.94}              & 94.47                & 93.29                 & \textbf{96.71}         & 64.27                      & 69.72                       & \textbf{78.61}             \\
\hline
\end{tabular}
}
\vspace{-1ex}
\caption{Accuracy of commercial APIs on LFW-C. Due to a large percentage of images being rejected, we report ``Rejection Rate," ``Accepted Samples Accuracy," and ``Actual Accuracy."}
\label{tab:api_c}
   \vspace{-4ex}
\end{table}

\subsection{Commercial API Evaluation Results}
\label{sec:5.3}
\vspace{-1ex}
We evaluate three commercial FR services (Aliyun, iFLYTEK, and Tencent) on LFW-C/V.
We follow their original threshold ranges and determine the optimal threshold for each dataset. The results of LFW-C are shown in Tab.~\ref{tab:api_c}. We leave the result of LFW-V in Appendix Tab.~\ref{tab:api_v}.

\noindent\textbf{Common Corruptions Evaluation.}
For LFW-C, 
the corruption data significantly impair the performance of commercial FR systems, often returning ``No face detected.” Based on this, we compute the \textit{Rejection Rate (RR)}, representing the proportion of rejected samples.
We find that nearly all categories of corruption have cases where the algorithm fails, 
especially for \textit{Salt-and-Pepper Noise}, where Aliyun’s API completely fails and gains 100.00\% RR. We also compute \textit{Accepted Samples Accuracy (ASA)} and \textit{Actual Accuracy (AA)}, representing the success rate among accepted samples and the overall samples, respectively. While Aliyun’s \textit{ASA} is relatively high, this is due to its high \textit{RR}, resulting in a low proportion of accepted samples. Ultimately, Tencent’s API achieves the best \textit{RR}, highest \textit{ASA}, and \textit{AA}, but there is still a noticeable decline in accuracy.

\vspace{-1ex}

  \begin{tcolorbox}[colframe=black, colback=gray!8, coltitle=black, sharp corners=all, boxrule=0.5mm, boxsep=0.1mm]
   \textbf{Insight 5:} 
 \textbf{Commercial FR APIs exhibit catastrophic failures under corruptions.} Excessive \textit{Rejection Rates} paradoxically inflate \textit{Accepted Sample Accuracy} but cause complete system breakdowns in identity verification,
  highlighting the vulnerability of commercial FR systems.

\end{tcolorbox}

\vspace{-2ex}

\begin{figure}[h]
  \centering
   \includegraphics[width=0.99\linewidth]{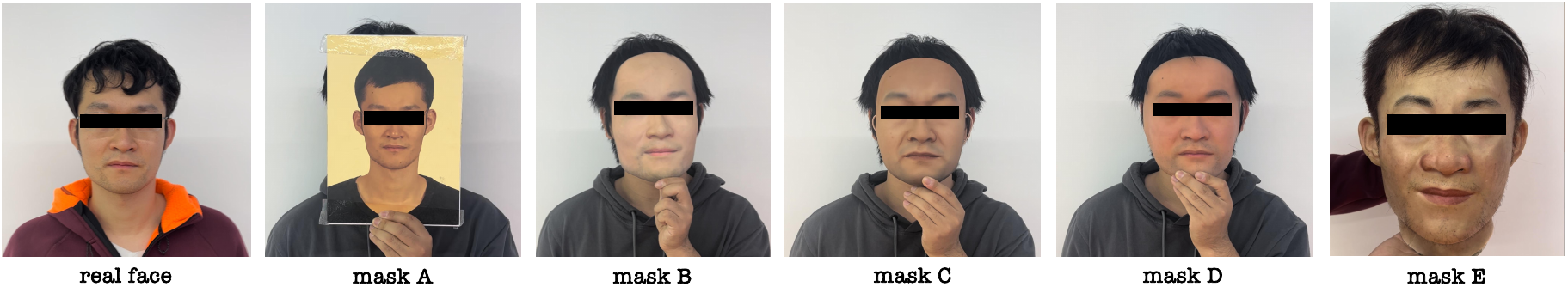}
   \caption{Display of face masks. We create 5 types of masks to test the impact of OOD cases on FR.}
   \label{fig:mask_demo}
   \vspace{-4ex}
\end{figure}

\subsection{Extended Experiments on Face Masks}
\label{sec:5.4}


Face masks typically refer to physical disguises used to obscure the identity of the wearer, preventing the system from accurately recognizing the individual~\cite{zhang2018deep}. 
As an extension, we conduct physical experiments on masks, creating five types made from various materials, as shown in Fig.~\ref{fig:mask_demo} (Detailed result and analysis in Appendix~\ref{sup:E}).
\vspace{-1ex}
  \begin{tcolorbox}[colframe=black, colback=gray!8, coltitle=black, sharp corners=all, boxrule=0.5mm, boxsep=0.1mm]
    \textbf{Insight 6:} 
    \textbf{Face masks exhibit material-dependent vulnerability patterns under OOD scenarios, with degradation trends differing from real faces.} This suggests a potential approach for spoof detection.

\end{tcolorbox}

\vspace{-2ex}

\section{Experiments Result on Potential Solutions}
In Sec.~\ref{sec:5}, FR models suffer performance degradation on the OODs. In this section, we explore potential solutions from two perspectives: applying potential defenses in Sec.~\ref{sec:5.6}; leveraging VLMs to address OOD challenges in Sec.~\ref{sec:5.5}.

\vspace{-1ex}

\subsection{Experiments on Defense Methods}
\label{sec:5.6}

\vspace{-3ex}

\begin{figure}[!th]
  \centering
   \includegraphics[width=0.99\linewidth]{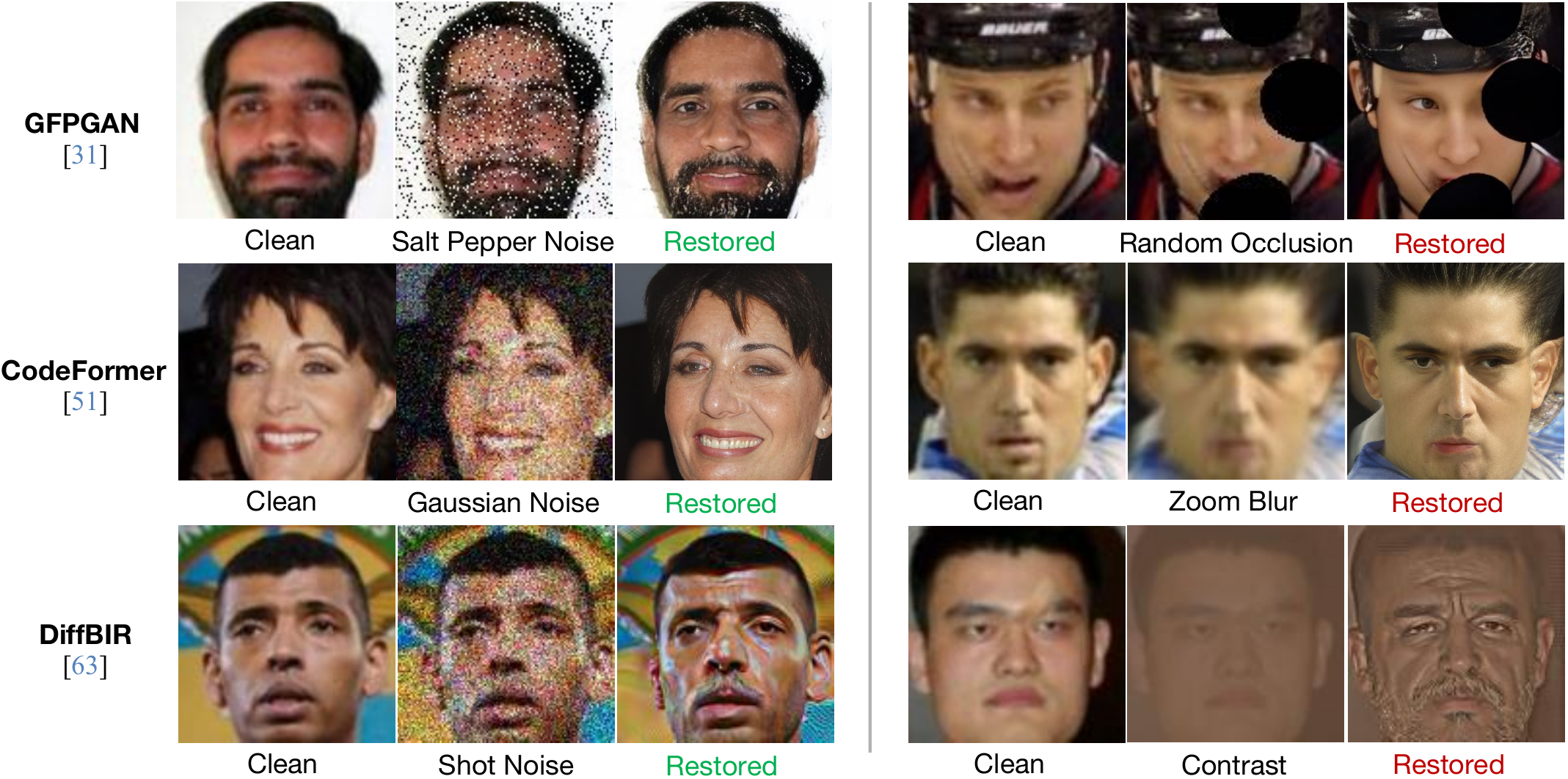}
\vspace{-2ex}
   \caption{Display of restoration methods as potential defenses. }
   \label{fig:restore}
      \vspace{-2ex}
\end{figure}

Following~\cite{yang2020robfr}, we test 10 robust models using \textit{input transformation} such as R\&P~\cite{xie2017mitigating}, Bit-Red~\cite{xu2017feature}, and \textit{adversarial training} such as PGD-AT~\cite{mkadry2017towards}, TRADES~\cite{zhang2019theoretically}. As Tab.~\ref{tab:robust_FR} in Appendix~\ref{sup:F}, the improvements are limited.
Further, following~\cite{dhake2024enhancing}, we explored three restoration methods~\cite{wang2021towards, zhou2022towards, lin2024diffbir} respectively based on GANs, Transformers, and Diffusion Models as potential defenses, and report the result in Appendix~\ref{sup:F}, Tab.~\ref{tab:restore}. Experiments show that restoration improves performance for certain OOD categories (e.g., \textit{Data \& Processing}), especially for models with weaker robustness, highlighting its potential. However, restoration can also interfere with feature extraction, leading to performance degradation in some OOD categories. Detailed analyses are provided in Appendix~\ref{sup:F}.

\vspace{-1ex}

\begin{table}[h]
  \centering
 \scalebox{0.83}{
  \setlength{\tabcolsep}{2pt}  

\begin{tabular}{c|cc|cc}
\hline
         & \multicolumn{2}{c|}{Open-source} & \multicolumn{2}{c}{Commercial} \\
Models   & LLaVA-NeXT      & InternVL2.5      & Qwen-VL-Plus      & GPT4o-mini         \\
\hline
Accuracy & 52.04           & 50.00         & 87.76     & \textbf{98.98}     \\
\hline
\end{tabular}
}
\vspace{-2ex}
\caption{VLMs FR performance on corruption data.}
\label{tab:VLMs}
\vspace{-4ex}
\end{table}

\subsection{Experiments on VLMs}
\label{sec:5.5}

\vspace{-1ex}

To further explore solutions for OOD scenarios, we investigate the use of Vision-Language Models for FR, a previously unexplored approach. We find that GPT-4o-mini~\cite{achiam2023gpt} demonstrates robust FR capabilities and can identify specific types of corruption in the images (see Fig.~\ref{fig:gpt_demo} in Appendix.~\ref{sup:G}). We further test more open-source (LLaVA-NeXT-8B, InternVL2.5-8B) and commercial (Qwen-VL-Plus) VLMs on the corruption data, with results shown in Tab.~\ref{tab:VLMs},
and additional analysis in Appendix.~\ref{sup:G}.

\vspace{-1ex}

  \begin{tcolorbox}[colframe=black, colback=gray!8, coltitle=black, sharp corners=all, boxrule=0.5mm, boxsep=0.1mm]
   \textbf{Insight 7:} 
   \textbf{Existing defenses fail to fully mitigate OOD scenarios. VLMs demonstrate robust FR potential under OODs, suggesting a solution for robust FR.} However, only commercial models achieve strong performance, the architectures and training processes remain opaque, posing challenges for deployment. 
\textbf{\textcolor{red}{Enhancing FR robustness remains an open challenge for future research.}}

\end{tcolorbox}

\vspace{-2ex}

\section{Discussion and Conclusion}
\label{sec:6}
\vspace{-1ex}

In this paper, we introduce \textbf{OODFace}, a comprehensive benchmark for OOD robustness in FR, systematically designed with 20 common corruptions across 5 categories and 10 appearance variations across 4 categories. By augmenting public datasets, we establish three robustness benchmarks: LFW-C/V, CFP-FP-C/V, and YTF-C/V. We conduct extensive evaluations on 19 FR models and 3 commercial APIs, along with additional experiments towards face masks, VLMs, and defense strategies.
Experimental results demonstrate that FR models suffer severe performance degradation under OOD scenarios, while existing strategies fail to fully mitigate these challenges. Based on the results and our insights, we outline potential research directions:

\begin{itemize}[leftmargin=3ex]
\item[1)] \textbf{Adaptive Defense Orchestration:} Future defenses could focus on adaptive defense orchestration, enabling OOD-aware defense selection.
\item[2)] \textbf{Decoupled Feature Restoration:} Developing decoupled feature restoration modules could help prevent feature distortions introduced by defense strategies.
\item[3)] \textbf{Robust FR Models:} Training with our comprehensive OOD dataset could be a promising approach to improve generalization and robustness of FR models.
\item[4)] \textbf{VLMs for FR:} It is essential to enhance the interpretability of the robust FR performance before exploring the design of specialized FR-VLMs or integrating VLMs into FR pipelines to enhance robustness. Above all, privacy concerns must be carefully addressed.

\end{itemize}

We hope that our comprehensive benchmarks, detailed analysis and insights will aid in understanding the robustness of FR models against OOD scenarios, and provide guidance for future improvements in FR model robustness.

{
    \small
    \bibliographystyle{ieeenat_fullname}
    \bibliography{main}
}

\clearpage
\setcounter{page}{1}
\maketitlesupplementary

\appendix
\numberwithin{equation}{section}
\numberwithin{figure}{section}
\numberwithin{table}{section}

\section{More Details of OOD Scenarios}
\label{sup:A}

\subsection{Implementation Details of Corruptions}
\label{sup:A1}

First, we describe the implementation details and hyperparameters of 20 common corruptions used in the LFW-C, CFP-C, and YTF-C benchmarks. Note that each corruption is evaluated at five severity levels, with specific hyperparameter configurations corresponding to each level.

\textbf{\emph{Gaussian Noise.}}  
Gaussian noise simulates sensor noise by adding random values with a normal distribution to the image. The noise intensity is controlled by the standard deviation, with five levels of severity: \{0.08, 0.12, 0.18, 0.26, 0.38\}. Noise is added to each pixel, creating effects of varying intensities. We implement this using  \textit{imagecorruptions}~\cite{imagecorruptions} library, simulating different levels of Gaussian noise with predefined severities \{1, 2, 3, 4, 5\}.

\textbf{\emph{Shot Noise.}}  
Shot noise simulates photon counting noise that occurs during image capture, particularly noticeable under low-light conditions. The intensity depends on the noise amplitude and the illumination level of the image. Severity levels are set as \{60, 25, 12, 5, 3\}, with higher levels introducing noticeable random brightness variations. We implement this using  \textit{imagecorruptions}~\cite{imagecorruptions} library, with severities \{1, 2, 3, 4, 5\}.

\textbf{\emph{Impulse Noise.}}  
Impulse noise replaces random pixel values with extremes (e.g., 0 or 255) to simulate transmission errors in images. The intensity of the noise is determined by its density, with levels \{0.03, 0.06, 0.09, 0.17, 0.27\}. Higher noise density results in more black-and-white speckles. We implement this using  \textit{imagecorruptions}~\cite{imagecorruptions} library, with severities \{1, 2, 3, 4, 5\}.

\textbf{\emph{Speckle Noise.}}  
Speckle noise simulates multiplicative noise caused by scattering, adding random values to each pixel. Noise intensity is controlled by levels \{0.15, 0.2, 0.35, 0.45, 0.6\}. As the intensity increases, the image becomes blurrier and the speckles more pronounced. We implement this using  \textit{imagecorruptions}~\cite{imagecorruptions} library, with severities \{1, 2, 3, 4, 5\}.

\textbf{\emph{Defocus Blur.}}  
Defocus blur simulates the effect of misfocused cameras, with the degree of blur controlled by the focal radius. Severity levels \{1, 2, 3, 4, 5\} correspond to different blur radii: level 1 uses a radius of 3, level 2 uses 4, level 3 uses 6, level 4 uses 8, and level 5 uses 10. Alias blur parameters range from 0.1 to 0.5 for each level. We implement this using \textit{imagecorruptions}~\cite{imagecorruptions} library to simulate defocus blur with predefined severity levels.

\textbf{\emph{Motion Blur.}}  
Motion blur simulates the relative movement of the camera or object during image capture. The intensity is controlled by the blur radius and standard deviation, which represent the motion's angle and length. Parameters for the five levels are: \{radius: 10, std: 3\}, \{radius: 15, std: 5\}, \{radius: 15, std: 8\}, \{radius: 15, std: 12\}, \{radius: 20, std: 15\}. Higher levels result in increased blur and more prominent directional effects. We implement this using  \textit{imagecorruptions}~\cite{imagecorruptions} library, using severities \{1, 2, 3, 4, 5\}.

\textbf{\emph{Zoom Blur.}}  
Zoom blur simulates the effect of changing the camera focal length during capture, with intensity controlled by the zoom factor. The zoom factors for five severity levels are: \{1.01, 1.11\}, \{1.01, 1.16\}, \{1.01, 1.21\}, \{1.01, 1.26\}, \{1.01, 1.31\}. Higher severity levels produce stronger zoom effects. We implement this using  \textit{imagecorruptions}~\cite{imagecorruptions} library, simulating predefined severity levels.

\textbf{\emph{Fog.}}  
Fog simulates the scattering effect caused by fog in the atmosphere, reducing image brightness and contrast. The intensity levels are set as \{0.1, 0.2, 0.3, 0.4, 0.5\}. As the intensity increases, the image becomes more blurred and grayish. Specifically, we implement this using \textit{imagecorruptions}~\cite{imagecorruptions} library, and the fog effect is implemented by adjusting image brightness and applying intensity factors, with parameters set for each level as \{1.5, 2\}, \{2.0, 2.0\}, \{2.5, 1.7\}, \{2.5, 1.5\}, and \{3.0, 1.4\}. These parameters control the strength and diffusion of the fog effect.

\textbf{\emph{Frost.}}  
Frost simulates the effect of frost forming on glass surfaces, with intensity controlled by frost density. The intensity levels are \{0.1, 0.2, 0.3, 0.4, 0.5\}, and as intensity increases, the image becomes increasingly obscured by frost, with details becoming blurred. We implement this using \textit{imagecorruptions}~\cite{imagecorruptions} library, specific intensity factors are set as \{1.0, 0.4\}, \{0.8, 0.6\}, \{0.7, 0.7\}, \{0.65, 0.7\}, and \{0.6, 0.75\}, controlling the strength and coverage of the frost effect.

\textbf{\emph{Snow.}}  
Snow simulates the appearance of snow by adding snowflake particles to the image, with intensity controlled by the density and size of the flakes. The intensity levels are \{10, 20, 30, 40, 50\}, we implement this using \textit{imagecorruptions}~\cite{imagecorruptions} library, with specific parameters as \{0.1, 0.3, 3, 0.5, 10, 4, 0.8\}, \{0.2, 0.3, 2, 0.5, 12, 4, 0.7\}, \{0.55, 0.3, 4, 0.9, 12, 8, 0.7\}, \{0.55, 0.3, 4.5, 0.85, 12, 8, 0.65\}, and \{0.55, 0.3, 2.5, 0.85, 12, 12, 0.55\}. These parameters control snowflake size, density, blur, and coverage. As intensity increases, snow density rises, gradually obscuring image details.

\textbf{\emph{Spatter.}}  
Spatter simulates the effect of splashes, such as water or paint, on a surface. The intensity is controlled by the size and distribution of splash particles. The intensity levels are \{1, 5, 10, 15, 20\}, we implement this using \textit{imagecorruptions}~\cite{imagecorruptions} library, with specific parameters as \{0.65, 0.3, 4, 0.69, 0.6, 0\}, \{0.65, 0.3, 3, 0.68, 0.6, 0\}, \{0.65, 0.3, 2, 0.68, 0.5, 0\}, \{0.65, 0.3, 1, 0.65, 1.5, 1\}, and \{0.67, 0.4, 1, 0.65, 1.5, 1\}. These parameters control the number, size, blur, and coverage of splashes. With higher intensity levels, splash marks become more prominent, with particles increasing and gradually covering image details.

\textbf{\emph{Contrast.}}  
Contrast adjustment modifies the range of brightness and the difference between light and dark areas, affecting the visual appearance of the image. The contrast levels are \{0.4, 0.3, 0.2, 0.1, 0.05\}, where higher values increase contrast and lower values decrease it. When contrast is increased, the light and dark areas become more distinct. Conversely, reduced contrast makes details and differences less visible. We implement this using \textit{imagecorruptions}~\cite{imagecorruptions} library.

\textbf{\emph{Brightness.}}  
Brightness adjustment changes the overall luminance of the image. The brightness levels are \{0.1, 0.2, 0.3, 0.4, 0.5\}, where lower values darken the image, and higher values brighten it. Increased brightness makes details more visible, but over-brightening may result in detail loss.

\textbf{\emph{Saturate.}}  
Saturation adjustment affects the intensity of colors in the image. The saturation levels are \{0.3, 0.1, 2.0, 5.0, 20.0\}, with higher values indicating stronger color saturation. When saturation increases, colors become more vivid, while lower saturation results in softer colors.

\textbf{\emph{JPEG Compression.}}  
JPEG compression simulates the information loss that occurs during image compression. The compression levels are \{25, 18, 15, 10, 7\}, where smaller values correspond to higher compression ratios, resulting in greater loss of image details and more noticeable compression artifacts. As the compression level increases, the image quality degrades, and more compression artifacts appear.

\textbf{\emph{Pixelate.}}  
The pixelate effect blurs image details by reducing the resolution, with the intensity controlled by the size of the pixel blocks. The pixelation levels are \{0.6, 0.5, 0.4, 0.3, 0.25\}, where smaller values correspond to larger pixel blocks and more loss of image detail. We implement this using \textit{imagecorruptions}~\cite{imagecorruptions} library.

\textbf{\emph{Facial Distortion.}}  
Facial distortion distorts the image by simulating elastic deformations on the object's surface, with the intensity controlled by the magnitude of the deformation. The deformation levels are \{0.05, 0.065, 0.085, 0.1, 0.12\}, where larger values result in stronger deformations of the image edges and shapes.

\textbf{\emph{Random Occlusion.}}  
Random occlusion simulates object occlusion by randomly generating elliptical occlusion regions within the image. The area of occlusion is controlled by the level, which is \{5\%, 10\%, 15\%, 20\%, 25\%\}. Higher levels correspond to larger occlusion regions, resulting in more covered details. Specifically, occlusion regions are generated by randomly selecting multiple locations in the image and drawing ellipses at those positions. The size and position of each occlusion are random, with the occlusion area being proportional to the level. By adjusting the number and size of the occlusion regions, image details are covered to varying degrees.

\textbf{\emph{Salt and Pepper Noise.}}  
Salt and pepper noise simulates dirty spots in an image by randomly setting some pixel values to 0 or 255. The noise density levels are \{0.01\%, 0.05\%, 0.1\%, 0.2\%, 0.5\%\}, with higher levels introducing more noise points in the image. Specifically, the intensity of the noise is controlled by the noise ratio at each level, where 0.01\% corresponds to fewer noise points, and 0.5\% corresponds to more. Based on the noise density, the code calculates the number of pixels to which noise should be added and sets these pixels' values to 0 or 255 at random positions to generate varying degrees of salt and pepper noise.

\textbf{\emph{Color Shift.}}  
Color shift simulates different lighting conditions or color changes by altering the hue values of the image. The hue shift levels are \{0, 7, 14, 21, 28\}, where each level represents the maximum hue shift in degrees (0-180). Higher levels result in more noticeable hue shifts and a more varied color palette. The magnitude of the hue shift is controlled by the level, followed by randomly generating a hue shift value within this range, applied to the image in the HSV color space, thereby altering the color style of the image.

\begin{table*}[!th]
\centering
   \scalebox{0.60}{
\setlength{\tabcolsep}{2pt}  
\begin{tabular}{c|c|ccc|ccccccc}
\hline
\multicolumn{2}{c}{Models}                               & \multicolumn{3}{c}{Input Transformation}              & \multicolumn{7}{c}{Adversarial Training}                                                                                        \\
\multicolumn{2}{c}{Corruptions}                          & Softmax-BR & Softmax-RP & Softmax-JPEG & TradesSoftmax & TradesCosFace & TradesArcFace & PGDSoftmax & PGDCosFace & PGDAm & PGDArcFace \\ \hline\hline
\rowcolor{lightgray}\multicolumn{2}{c}{None (clean)}                         & 99.53           & 99.48           & 99.53             & 90.65              & 90.43              & 94.92              & 90.85           & 85.53           & 84.57      & 87.07           \\ \hline
\multirow{5}{*}{Lighting \& Weather} & Brightness        & 99.02           & 98.85           & 98.94             & 84.64              & 84.33              & 91.23              & 86.50           & 79.62           & 77.79      & 80.25           \\
                                     & Contrast          & 90.92           & 88.79           & 89.13             & 66.46              & 64.21              & 62.02              & 61.64           & 57.84           & 58.80      & 59.46           \\
                                     & Saturate          & 99.29           & 99.20           & 99.29             & 89.16              & 87.39              & 93.99              & 89.64           & 84.21           & 82.62      & 86.27           \\
                                     & Fog               & 89.89           & 89.12           & 89.72             & 61.22              & 59.05              & 59.14              & 57.67           & 55.95           & 57.42      & 56.44           \\
                                     & Snow              & 95.14           & 94.31           & 95.13             & 84.16              & 80.83              & 88.68              & 85.65           & 79.45           & 77.36      & 80.71           \\
                                     \hline
\multirow{3}{*}{Sensor}              & Defocus Blur      & 87.11           & 86.78           & 87.38             & 83.90              & 70.14              & 80.44              & 83.41           & 77.41           & 78.30      & 79.61           \\
                                     & Color Shift       & 99.46           & 99.39           & 99.44             & 88.69              & 87.17              & 93.71              & 88.33           & 81.54           & 78.68      & 83.83           \\
                                     & Pixelate          & 98.91           & 98.81           & 98.91             & 89.72              & 88.41              & 93.96              & 90.04           & 84.28           & 83.86      & 86.10           \\
                                     \hline
\multirow{3}{*}{Movement}            & Motion Blur       & 94.12           & 93.58           & 94.38             & 85.69              & 79.42              & 86.25              & 85.60           & 79.49           & 80.06      & 81.81           \\
                                     & Zoom Blur         & 99.06           & 98.93           & 99.05             & 89.60              & 88.50              & 93.72              & 90.27           & 84.30           & 83.86      & 86.32           \\
                                     & Facial Distortion & 92.37           & 92.19           & 92.46             & 87.37              & 81.40              & 89.30              & 86.50           & 82.12           & 81.70      & 83.60           \\
                                     \hline
\multirow{6}{*}{Data \& Processing}  & Gaussian Noise    & 81.74           & 81.34           & 80.61             & 79.05              & 79.03              & 81.08              & 71.37           & 71.01           & 70.63      & 73.78           \\
                                     & Impulse Noise ️   & 83.63           & 83.11           & 82.39             & 75.65              & 78.89              & 79.05              & 72.05           & 70.06           & 69.79      & 73.26           \\
                                     & Shot Noise        & 78.61           & 78.16           & 77.54             & 76.34              & 76.00              & 76.79              & 68.23           & 68.48           & 67.86      & 71.09           \\
                                     & Speckle Noise     & 86.18           & 85.71           & 84.78             & 78.28              & 78.69              & 80.63              & 71.46           & 70.99           & 69.96      & 73.33           \\
                                     & Salt Pepper Noise & 64.83           & 64.27           & 66.48             & 68.31              & 67.48              & 66.51              & 60.96           & 62.60           & 63.29      & 66.88           \\
                                     & Jpeg Compression  & 98.93           & 98.88           & 98.94             & 90.19              & 88.93              & 94.43              & 90.54           & 84.80           & 84.17      & 86.65           \\
                                     \hline
\multirow{3}{*}{Occlusion}           & Random Occlusion  & 94.08           & 93.17           & 94.02             & 72.39              & 68.36              & 73.09              & 72.00           & 68.23           & 64.94      & 68.84           \\
                                     & Frost             & 93.62           & 93.09           & 93.40             & 76.59              & 75.20              & 85.34              & 78.24           & 71.54           & 70.25      & 72.24           \\
                                     & Spatter           & 96.87           & 96.48           & 96.63             & 83.84              & 79.02              & 86.58              & 81.85           & 78.68           & 77.31      & 79.45           \\
                                     \hline
\rowcolor{lightgray} \multicolumn{2}{c}{Average}                              & 91.19           & 90.71           & 90.93             & 80.56              & 78.12              & 82.80              & 78.60           & 74.63           & 73.93      & 76.50   \\
\hline
\end{tabular}
}
\caption{Accuracy of 19 robust FR models on LFW-C, categorized into input transformation, adversarial training.}
\label{tab:robust_FR}
\vspace{-2ex}
\end{table*}

\begin{figure*}[!th]
  \centering
   \includegraphics[width=0.99\linewidth]{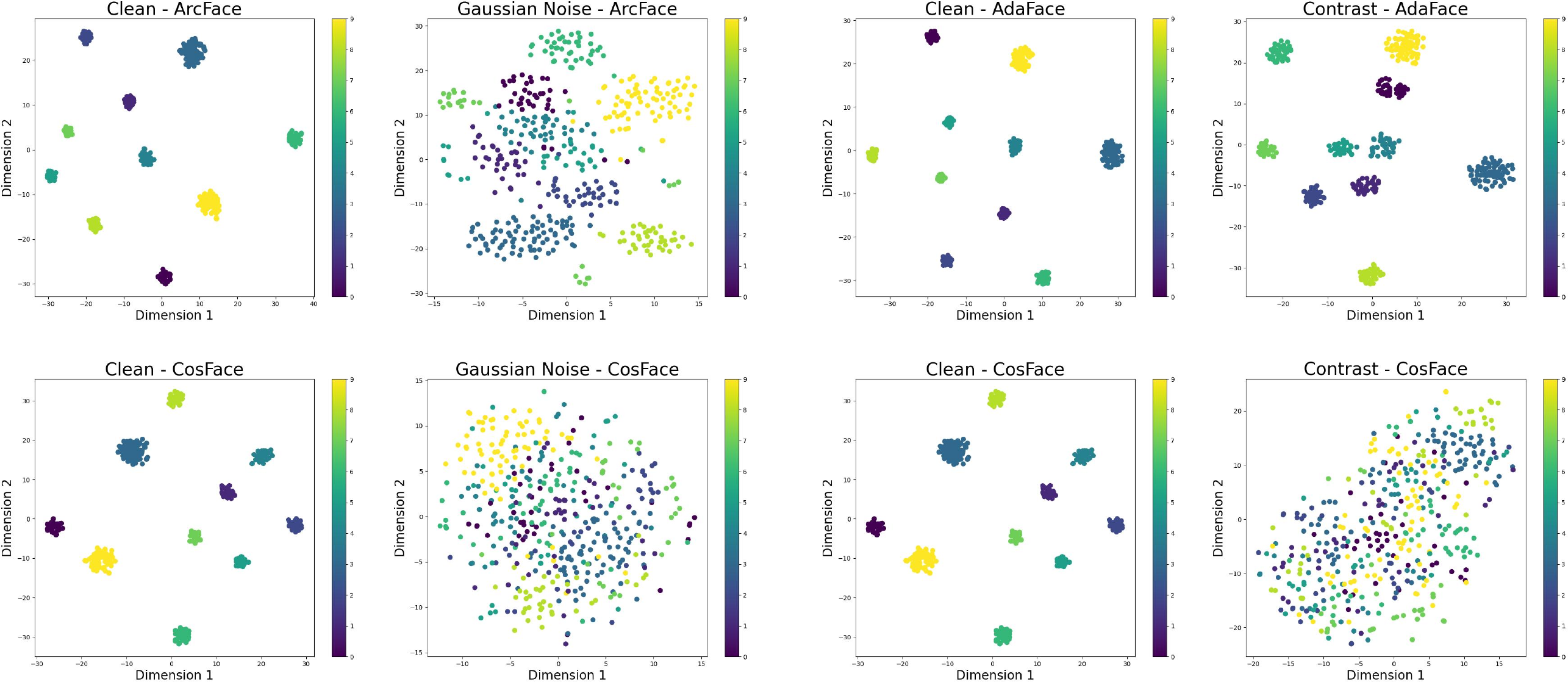}
   \caption{t-SNE visualization of feature distributions before and after applying different OOD simulations to the models.}
   \label{fig:tsne}
      \vspace{-2ex}
\end{figure*}

\begin{table*}[ht]
 \scalebox{0.45}{
   \setlength{\tabcolsep}{2pt}  
\begin{tabular}{c|cc|cccccccc|cccccc|ccccc|cc}
\hline
\multirow{2}{*}{Restoration} & \multicolumn{2}{c|}{Models}                                & \multicolumn{8}{c}{Open-Source Model Eval}                                                      & \multicolumn{6}{c}{Model Architecture Eval}                                                                        & \multicolumn{5}{c}{Optimization Loss Eval}            & \multirow{2}{*}{Average} \\
                              & \multicolumn{2}{c|}{Corruptions}              & FaceNet        & SphereFace & CosFace & ArcFace        & ElasticFace& AdaFace        & TransFace      & TopoFR & MobileFace     & Mobilenet & Mobilenet-v2 & ShuffleNet& ShuffleNet-v2& ResNet50       & Softmax-IR & SphereFace-IR  & Am-IR          & CosFace-IR     & ArcFace-IR     \\ \hline\hline

\multirow{21}{*}{GFPGAN}     & \multirow{5}{*}{L \& W} & Brightness        & \cellcolor{myred} -0.68   & \cellcolor{myred} -0.53      & \cellcolor{myred} -0.72   & \cellcolor{myred} -0.47   & \cellcolor{myred} -0.40            & \cellcolor{myred} -0.67   & \cellcolor{myred} -0.20     & \cellcolor{myred} -0.08       & \cellcolor{myred} -0.30      & \cellcolor{myred} -0.35     & \cellcolor{mygreen} 0.02        & \cellcolor{myred} -0.42                  & \cellcolor{myred} -0.50                                   & \cellcolor{myred} -0.43    & \cellcolor{myred} -0.42      & \cellcolor{myred} -0.38                  & \cellcolor{myred} -0.38  & \cellcolor{myred} -0.35      & \cellcolor{myred} -0.37      & \cellcolor{myred} -0.40                    \\
\hline             
\end{tabular}
}
\caption{Accuracy differences after implementing three defense methods on LFW-C. Green squares indicate improvement, while red squares indicate a decrease.}
\label{tab:restore}
\end{table*}

\subsection{Implementation Details of Variations}  
\label{sup:A2}
Next, we present the implementation details and hyperparameters for the 10 appearance variations across the three benchmarks: LFW-V, CFP-V, and YTF-V. Additionally, each corruption has five severity levels.

\textit{\textbf{Age-.}  }
Age reduction is an important factor in facial changes, as facial features undergo noticeable alterations with age, such as changes in skin texture, sagging, wrinkles, and overall facial structure. We simulate facial rejuvenation using a generative model, which reduces signs of aging, making the face appear younger. The age reduction levels consist of five stages, ranging from mild to significant rejuvenation. The higher the level, the more pronounced the reduction in aging signs. We implement this using the PTI~\cite{roich2022pivotal} algorithm.

\textit{\textbf{Age+.}  }
In contrast to age reduction, age increment simulates the process of facial aging, with features such as wrinkles, sagging, and skin aging becoming more prominent. We simulate facial aging using a generative model, increasing the aging features to make the face appear older. The age increment levels are also divided into five stages, with higher levels corresponding to more pronounced aging features. This is implemented using the PTI~\cite{roich2022pivotal} algorithm.

\textit{\textbf{Mouth-close.}  }
Mouth closure is a significant facial expression change, commonly occurring in calm or serious states. We simulate the effect of mouth closure using a generative model, which tightens the lips and hides any expression. The mouth closure levels range from minimal to complete closure. The higher the level, the more pronounced the mouth closure, from slight changes to fully closed lips. We implement this using the PTI~\cite{roich2022pivotal} algorithm.

\textit{\textbf{Mouth-open.}  }
Mouth opening is typically associated with changes in facial expression, such as smiling, surprise, or speaking. We simulate the effect of mouth opening using a generative model, separating the lips to display various expressions. The mouth opening levels are also divided into five stages, with higher levels corresponding to larger openings, from slight to fully open. We implement this using the PTI~\cite{roich2022pivotal}.

\textit{\textbf{Eye-close.}  }
Eye closure is often related to fatigue, drowsiness, or certain emotional states. We simulate the effect of eye closure using a generative model, completely closing the eyes and covering the eyeballs. The eye closure levels range from slight to complete closure. The higher the level, the more pronounced the eye closure, from barely closed to fully shut. We implement this using the Ganspace~\cite{harkonen2020ganspace}, which uses principal component analysis (PCA) in the latent or feature space to identify important directions and demonstrates that large amounts of interpretable control can be defined by progressively perturbing along these main directions.

\textit{\textbf{Eye-open.}  }
Eye opening is generally associated with alertness, wakefulness, or certain expressions. We simulate the effect of eye opening using a generative model, fully opening the eyes to display various emotions or states. The eye opening levels are similarly divided into five stages, with higher levels corresponding to larger openings, from slightly open to fully open. We implement this using the Ganspace~\cite{harkonen2020ganspace}.

\begin{table*}[!th]
\centering
 \scalebox{0.65}{
\begin{tabular}{c|ccccc|cccc|c|cc|ccc}
\hline

& \multicolumn{5}{c|}{Common Corruptions}   & \multicolumn{4}{c|}{Appearance Variations}   &    & \multicolumn{2}{c|}{FR Models} & \multicolumn{3}{c}{Extensions}      \\
 
& Sensor & Movement & Occlusion & L \& W & D \& P & Age & Rotation & Facial Exp. & Accessories & Categories Num. & Open-source    & API  & Defense  & VLMs & Phys. Exp.        \\
\hline \hline
\cite{ahsan2021evaluating}                    & $\times$ & $\times$ & $\times$ & \faCheck & $\times$ & $\times$ & $\times$ & $\times$ & $\times$ & 5                    & 3           & $\times$ & $\times$ & $\times$ & $\times$ \\
\cite{neto2022beyond}                    & $\times$ & $\times$ & \faCheck & $\times$ & $\times$ & $\times$ & $\times$ & $\times$ & $\times$ & 9                    & 8           & $\times$ & $\times$ & $\times$ & $\times$ \\
\cite{agarwal2024face}                    & $\times$ & $\times$ & $\times$ & $\times$ & $\times$ & $\times$ & $\times$ & \faCheck & $\times$ & 13                   & 3           & $\times$ & $\times$ & $\times$ & $\times$ \\
\cite{dhake2024enhancing}                   & \faCheck & $\times$ & $\times$ & \faCheck & \faCheck & $\times$ & $\times$ & $\times$ & $\times$ & 6                    & 8           & $\times$ & \faCheck & $\times$ & $\times$ \\
Ours                 & \faCheck & \faCheck & \faCheck & \faCheck & \faCheck & \faCheck & \faCheck & \faCheck & \faCheck & \textbf{30}                   & \textbf{19}          & \textbf{3}                     & \faCheck & \faCheck & \faCheck \\
\hline
\end{tabular}
}
\caption{Comparison with related work. Our benchmark provides the most comprehensive evaluation.}
\label{tab:compare}
\end{table*}

\textit{\textbf{Rotation-left.}  }
Left rotation of the face simulates different head poses, especially from side views. We simulate the effect of rotating the head to the left using an editing algorithm, enabling the face to change from different angles, thus enhancing the model's adaptability to various viewpoints. The rotation levels are divided into five stages, with higher levels corresponding to larger angles, from slight leftward tilts to a full left-side view. We implement this using the PTI~\cite{roich2022pivotal}.

\textit{\textbf{Rotation-right.}  }
Right rotation, in contrast to left rotation, simulates the effect of turning the face to the right. By adjusting the head rotation with an editing algorithm, we simulate the transition from a frontal view to a right-side perspective. The rotation levels are similarly divided into five stages, with higher levels corresponding to larger angles, from slight rightward tilts to a full right-side view. We implement this using the PTI~\cite{roich2022pivotal}.

\textit{\textbf{Bangs \& Glasses.}  }
Bangs and glasses are common facial occlusions that significantly alter the visual appearance of the face. By adding various accessories such as bangs or glasses (e.g., clear lenses, reflective lenses), we simulate real-life facial occlusion scenarios. The levels consist of five stages: wearing only clear-lens glasses, wearing only bangs, wearing reflective-lens glasses, wearing clear-lens glasses + bangs, and wearing reflective-lens glasses + bangs. Each level adds a different degree of facial occlusion, making facial features less visible. We implement this using the HiSD~\cite{li2021image}.

\textit{\textbf{Makeup.}  }
Makeup can significantly alter the appearance of the face by changing features such as eyebrows, eyeliner, blush, and lipstick. We simulate different styles of makeup using a generative model Beautygan~\cite{li2018beautygan} to mimic real-life makeup scenarios. The makeup levels are divided into five stages, including: Japanese style, Korean style, and others.

\textbf{Variation Levels:}
For age adjustments, the levels indicate the degree of age reduction or increase. For facial expressions, the levels represent the extent of mouth and eye movements. Head pose levels denote the degree of facial rotation. For accessory additions, the levels are defined as follows: glasses (clear lenses), bang occlusion, glasses (reflective lenses), glasses (clear lenses) + bang, and glasses (reflective lenses) + bang. For the makeup variation, we adopt the styles defined in~\cite{li2018beautygan}: Japanese style, Korean style, Retro style, Flashy style, and Smoky-eyes style. Each level represents an increase in perceived appearance interference.

\subsection{Comparison with Related Work}  
\label{sup:A3}
As mentioned in Section 2.2, related studies \cite{agarwal2024face, dhake2024enhancing, ahsan2021evaluating, neto2022beyond} have also explored data corruption in face recognition. Compared to them, our benchmark is more comprehensive in terms of OOD types, evaluation datasets, and the FR models studied, as shown in Tab.~\ref{tab:compare}.

For instance, \cite{ahsan2021evaluating} examines 5 Lighting \& Weather OOD categories and evaluates 3 open-source models; \cite{neto2022beyond} primarily focuses on 9 OOD challenges under the Occlusion category, testing 8 open-source models; \cite{agarwal2024face} centers on variations in Facial Expression, while \cite{dhake2024enhancing} covers a broader range of common corruptions, including Sensor, Lighting \& Weather, and Data \& Processing, alongside discussions on defense strategies.
However, our benchmark systematically evaluates the most extensive set of OOD scenarios, categorized into Common Corruptions and Appearance Variations, covering nine main categories and 30 subcategories. We assess 19 open-source models, 3 commercial APIs, and further investigate defense mechanisms, physical-world face mask experiments, and the potential utilization of VLMs.

Notably, previous works have not comprehensively considered corruption scenarios specific to FR. In contrast, we are the first to investigate such OOD cases within a unified robustness benchmark, including OOD challenges like Facial Distortion, Random Occlusion, Age, and Accessories—factors explicitly designed for FR yet previously unexplored.

\subsection{Naturalness of OOD Synthesis.}
\label{sup:A4}

 While it is impossible to exhaust all real-world OOD types, we systematically designed 30 OOD categories into 5 levels, to serve as a practical testbed for controllable robustness evaluation. 
 Certain OOD types originate from digital artifacts, such as noise corruptions and JPEG compression, while others follow the synthesis strategy in~\cite{hendrycks2019benchmarking}, where we carefully adjust parameters to ensure realism (details in Appendix~\ref{sup:A1}). Specifically, for Appearance Variations, we employ state-of-the-art generative simulation methods proven to closely approximate real facial variations~\cite{roich2022pivotal,li2021image,li2018beautygan}. Despite the inevitable gap between synthetic and real data, our experiments demonstrate that model performance under synthetic conditions aligns well with real-world results. Quantitative analysis is provided in Tab.~\ref{tab:age-naturalness}.

\textbf{Quantitative Analysis.}
Firstly, we focus on one of the most complex variations—age progression—as an example, by leveraging the AgeDB~\cite{moschoglou2017agedb} dataset. We analyze FR model performance under both synthetically generated and real-world age variations. AgeDB provides identity-labeled age variations, allowing us to evaluate the four FR models introduced in Sec.~\ref{sec:4.4} under both conditions, results are shown in Tab.~\ref{tab:age-naturalness}. Notably, model performance remains consistent across synthetic and real-world aging scenarios, validating the effectiveness of our proposed OOD benchmark for assessing FR robustness.

\textbf{Data Quality Verification.}
Data quality verification is also a critical part of our benchmark. During OOD scenario simulation, we ensure data integrity by carefully adjusting the severity levels of each corruption and variation to maintain human-recognizable faces. Detailed verification procedures are provided in Appendix~\ref{sup:A1} and \ref{sup:A2}.

\begin{table}[]
 \scalebox{0.8}{
\begin{tabular}{c|c|cccc}
\hline
\multicolumn{2}{l|}{}               & ElasticFace & AdaFace        & TransFace      & TopoFR \\
\hline
\multirow{2}{*}{Synthetic} & Age-  & 96.32       & \textbf{96.48} & 96.12 & 96.08  \\
                           & Age+  & 96.39       & \textbf{96.6}  & 96.39          & 96.19  \\
                           \hline
\multirow{2}{*}{Real}      & Young & 95.00       & \textbf{95.50} & 94.50          & 94.00  \\
                           & Old   & 96.50       & \textbf{97.00} & 96.50          & 96.00  \\ \hline
\end{tabular}
}
\caption{Comparison of simulation methods and real data on age variations. The results demonstrate that models exhibit similar trends across both data types.}
\label{tab:age-naturalness}
\end{table}

\begin{figure*}[!th]
  \centering
   \includegraphics[width=0.99\linewidth]{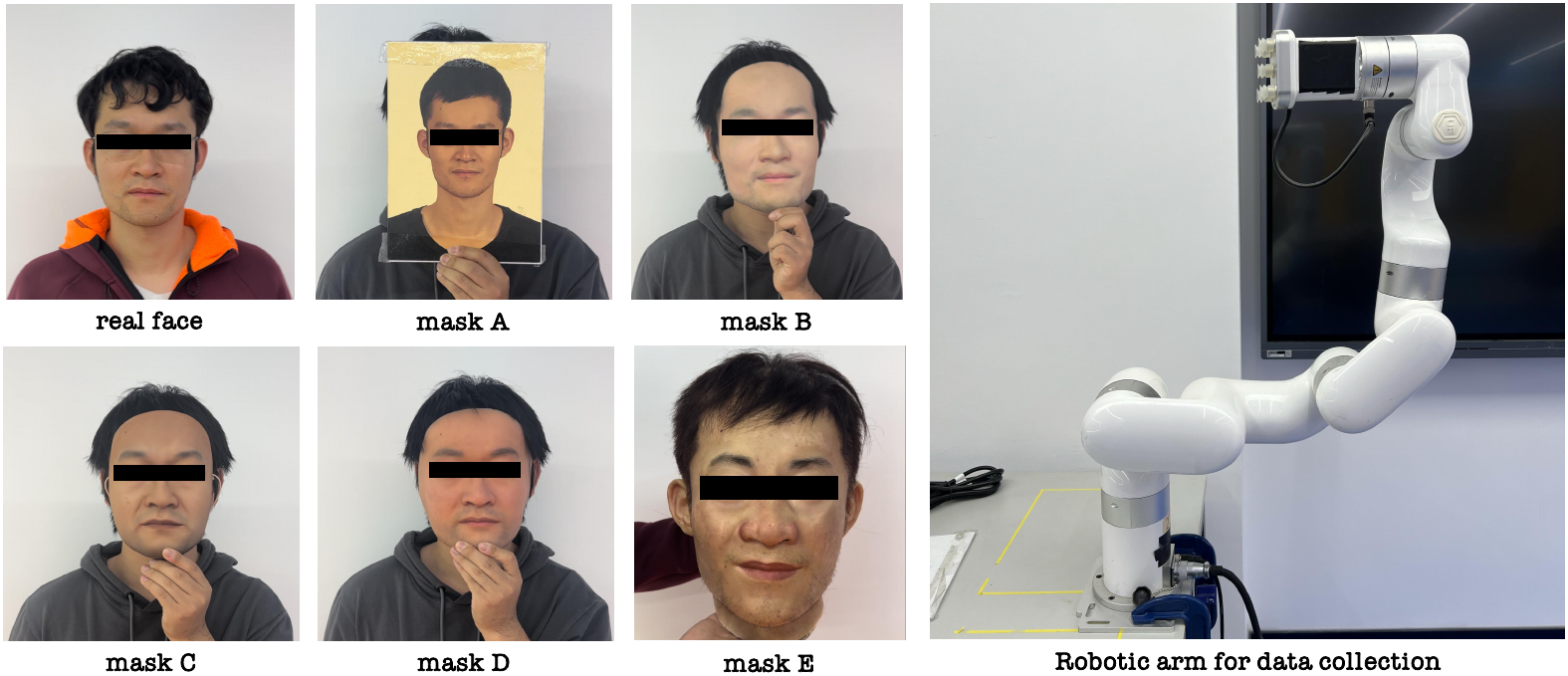}
   \caption{Display of face masks and robotic arm for data collection}
   \label{fig:robotic_arm}
\end{figure*}

\section{Additional Results on LFW-C/V}
\label{sup:B}

Due to space limitations in the main text, we supplement additional data from LFW-C/V in this section.

\subsection{LFW-C}
\begin{figure*}[h]
  \centering
   \includegraphics[width=0.99\linewidth]{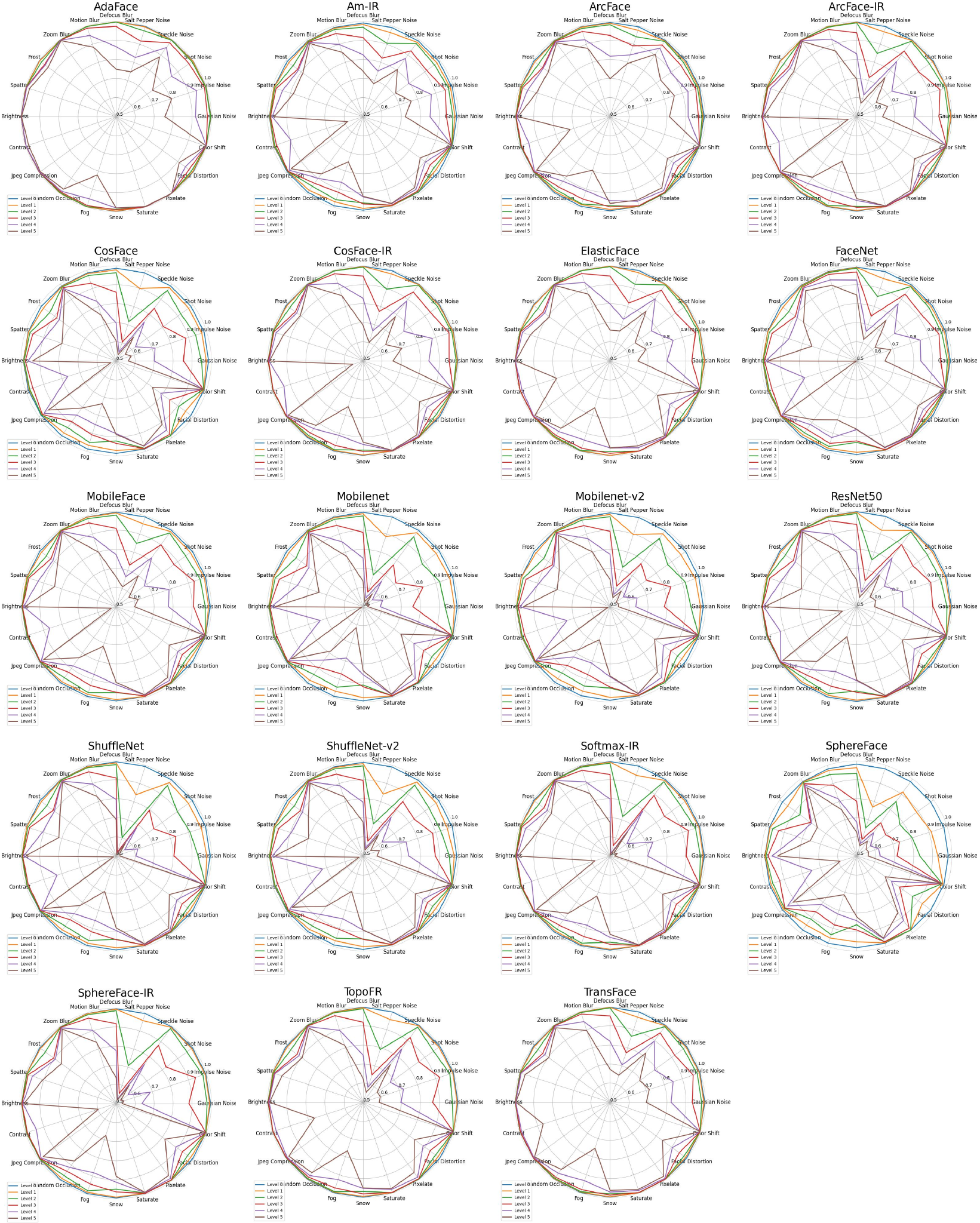}
   \caption{Graded radar charts for each model on LFW-C.}
   \label{fig:radar_LFW_c}
\end{figure*}

\begin{figure*}[h]
  \centering
   \includegraphics[width=0.99\linewidth]{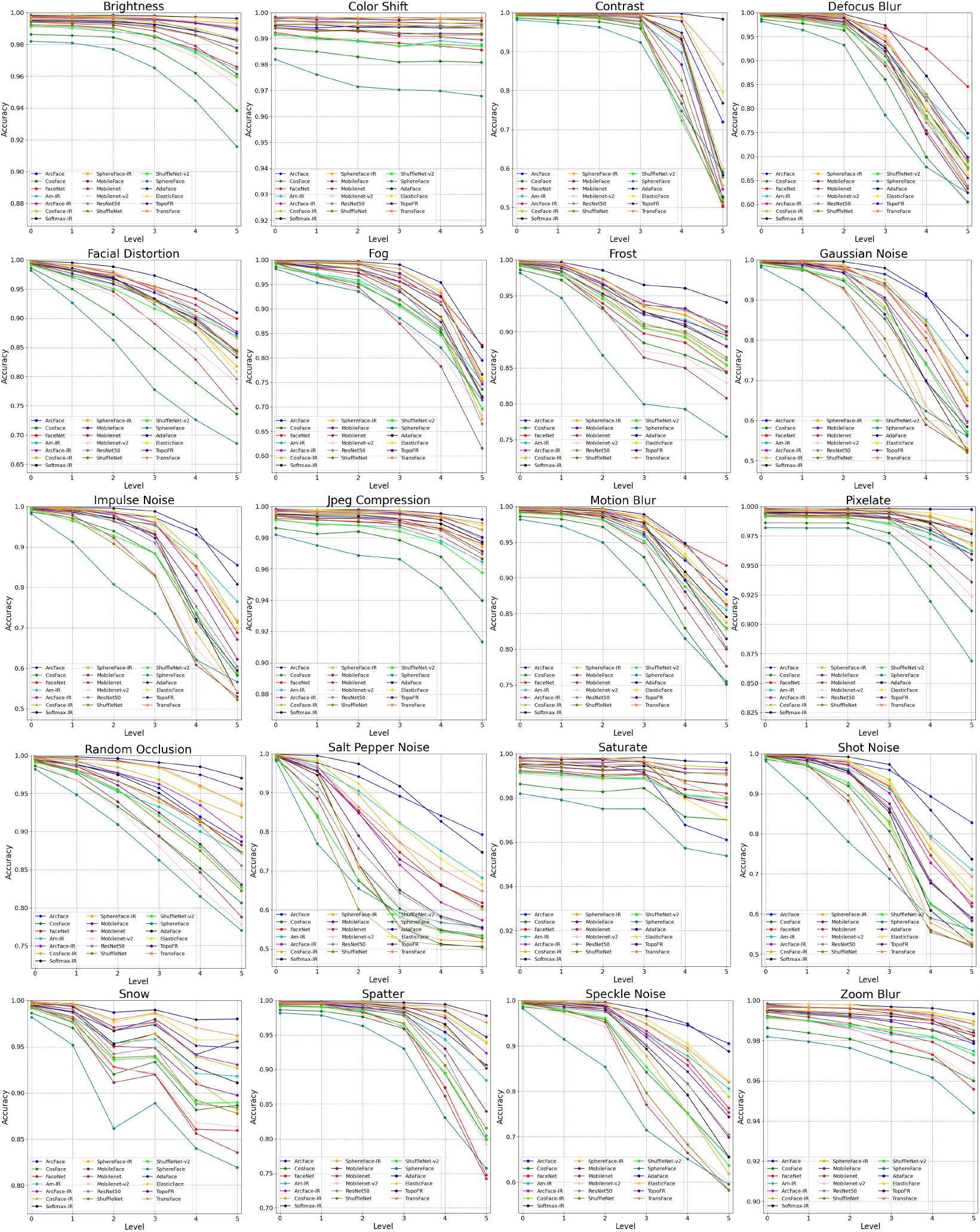}
   \caption{Graded line charts for each corruption on LFW-C.}
   \label{fig:line_LFW_c}
\end{figure*}

For LFW-C, we design 5 levels for each corruption, ranging from level 1 (mild) to level 5 (extreme). In this section, we provide the specific data tables corresponding to levels 1-5, as shown in Tab.~\ref{tab:accuracy_LFW-C_level1}, Tab.~\ref{tab:accuracy_LFW-C_level2}, Tab.~\ref{tab:accuracy_LFW-C_level3}, Tab.~\ref{tab:accuracy_LFW-C_level4}, Tab.~\ref{tab:accuracy_LFW-C_level5}.
For each model, we present a radar chart that shows the performance of the model across 20 different types of corruption, providing a clearer view of how the model's performance varies under different types of corruption in each category, as shown in Fig.~\ref{fig:radar_LFW_c}.
For each corruption category, we also provide line charts showing the model's performance across the 5 levels of each corruption, illustrating the performance fluctuations as the level of corruption increases, as shown in Fig.~\ref{fig:line_LFW_c}.

From the graded data results, we observe that as the severity level increases, the best-performing model against corruption changes. Taking ``Open-source Model Eval" as an example, although AdaFace maintains the best average accuracy across all 5 levels, we find that at level 5, ArcFace significantly surpasses AdaFace in accuracy on Data \& Processing. This further demonstrates that different models exhibit varying robustness to different types of corruption.

\subsection{LFW-V}
\label{sup:LFW-V}
\begin{figure*}[h]
  \centering
   \includegraphics[width=0.99\linewidth]{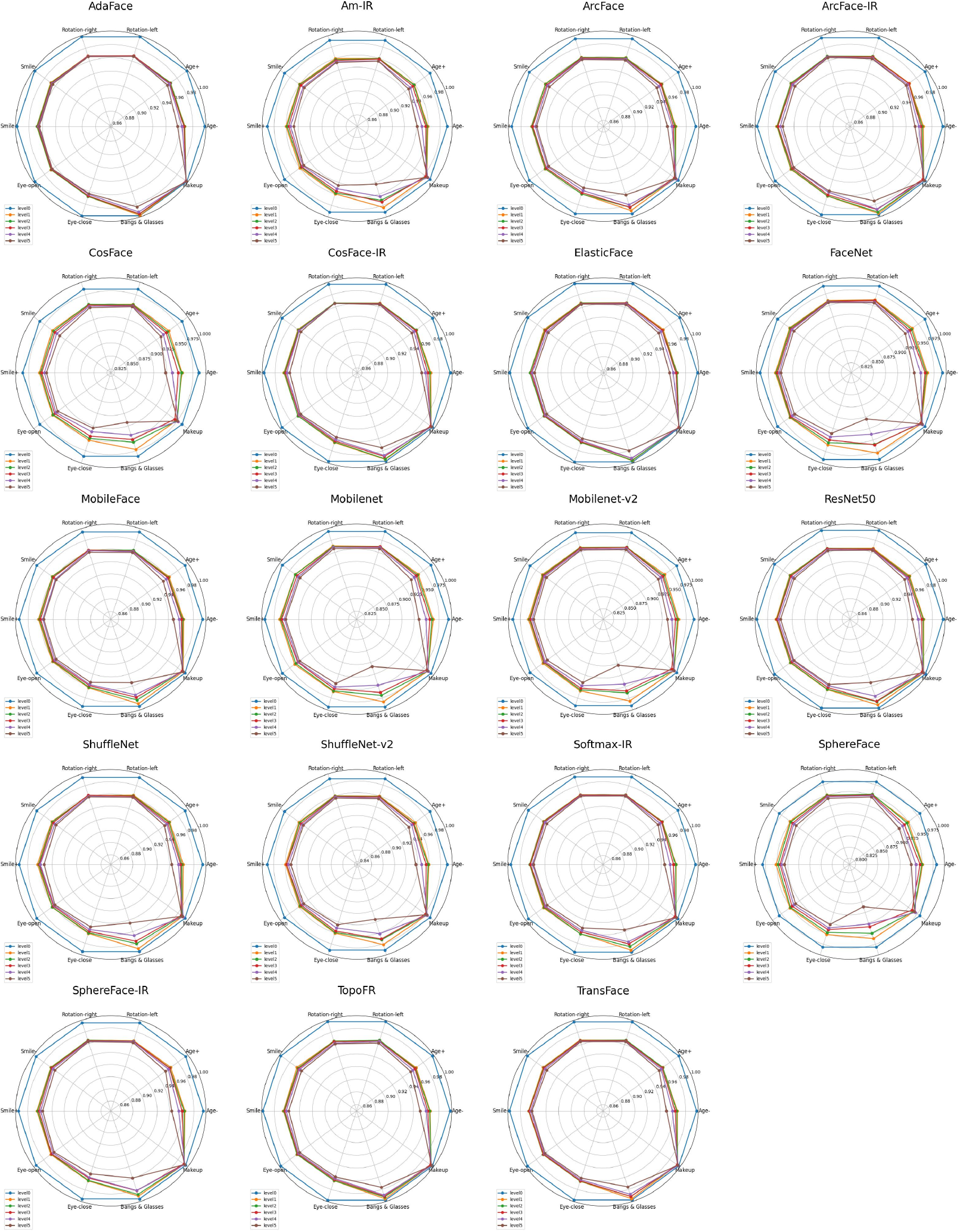}
   \caption{Graded radar charts for each model on LFW-V.}
   \label{fig:radar_LFW_v}
\end{figure*}

\begin{figure*}[h]
  \centering
   \includegraphics[width=0.99\linewidth]{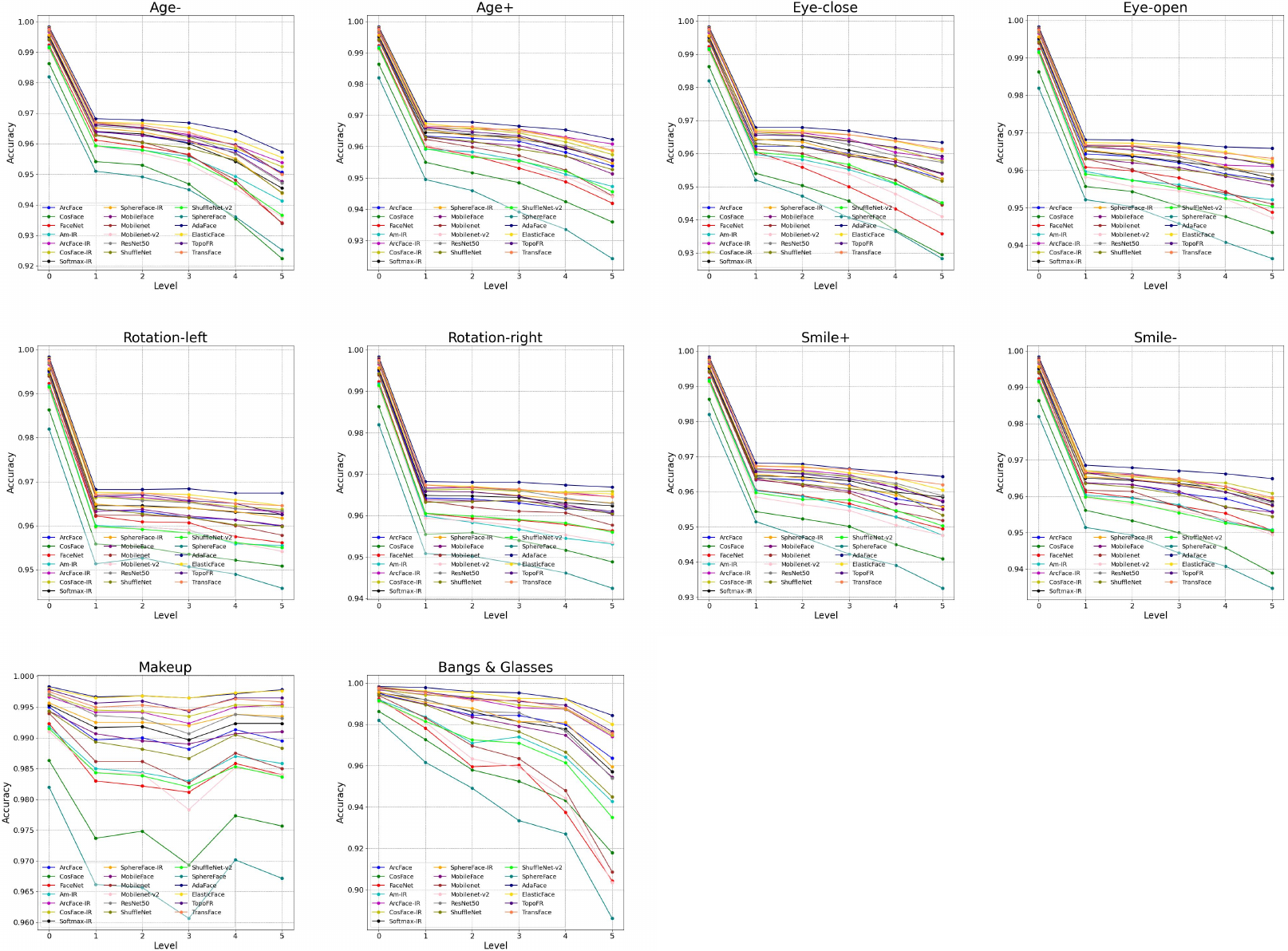}
   \caption{Graded line charts for each corruption on LFW-V.}
   \label{fig:line_LFW_v}
\end{figure*}

Similarly, for LFW-V, we design 5 levels for each variation. We supplement the specific data tables corresponding to levels 1-5 in this section, as shown in Tab.~\ref{tab:accuracy_LFW-v_level1}, Tab.~\ref{tab:accuracy_LFW-v_level2}, Tab.~\ref{tab:accuracy_LFW-v_level3}, Tab.~\ref{tab:accuracy_LFW-v_level4}, Tab.~\ref{tab:accuracy_LFW-v_level5}. We present the radar charts in Fig.~\ref{fig:radar_LFW_v}, as well as the line charts showing the 5 levels for each category, in Fig.~\ref{fig:line_LFW_v}.

\textbf{Variations Evaluation for APIs.}
 Additionally, as shown in Tab.~\ref{tab:api_v}, for LFW-V, no detection rejection occurs as seen in LFW-C, so we report the direct accuracy results. Appearance variations also cause performance degradation, but the impact is smaller compared to corruptions. Among the APIs, Tencent achieves the highest average accuracy, while Aliyun performs better in the Age and Facial Expression. The robustness of these services remains highly correlated with their clean performance, aligning with the conclusions in Sec.~\ref{sec:5.2}.

\begin{table}[]
\centering
 \scalebox{0.85}{
  \setlength{\tabcolsep}{2pt}  
\begin{tabular}{c|c|ccc}
\hline
\multicolumn{2}{c}{Models}                          & \multicolumn{3}{c}{Accuracy}                                                           \\
\multicolumn{2}{c}{Variations}                     & \multicolumn{1}{c}{Aliyun} & \multicolumn{1}{c}{iFLYTEK} & \multicolumn{1}{c}{Tencent} \\
\hline

\rowcolor{lightgray} \multicolumn{2}{c}{None (clean)}                    & 99.65                      & 97.99                       & \textbf{99.75}              \\
\hline
\multirow{2}{*}{Age}               & Age-           & \textbf{96.60}             & 93.47                       & 96.45                       \\
                                   & Age+           & \textbf{96.60}             & 93.50                       & 96.59                       \\
                                   \hline
\multirow{4}{*}{Facial Expression} & Mouth-close    & \textbf{96.67}             & 93.80                       & 96.47                       \\
                                   & Mouth-open     & \textbf{96.55}             & 93.68                       & 96.55                       \\
                                   & Eye-close      & \textbf{96.71}             & 93.23                       & 96.64                       \\
                                   & Eye-open       & 96.67                      & 94.10                       & \textbf{96.72}              \\
                                   \hline
\multirow{2}{*}{Rotation}          & Rotation-left  & 96.67                      & 94.23                       & \textbf{96.71}              \\
                                   & Rotation-right & \textbf{96.71}             & 94.10                       & 96.67                       \\
                                   \hline
\multirow{2}{*}{Accessories}       & Bangs\&Glasses & 98.06                      & 93.91                       & \textbf{98.85}              \\
                                   & Makeup         & 99.13                      & 96.60                       & \textbf{99.28}              \\
                                   \hline
\rowcolor{lightgray} \multicolumn{2}{c}{Average}                         & 97.27                      & 94.42                       &  \textbf{97.33}           \\
\hline  
\end{tabular}
}
\vspace{-1ex}
\caption{Accuracy of 3 commercial APIs on LFW-V.}
\label{tab:api_v}
   \vspace{-3ex}
\end{table}

\section{Results on CFP-C/V}
\label{sup:C}

\begin{figure*}[h]
  \centering
   \includegraphics[width=0.99\linewidth]{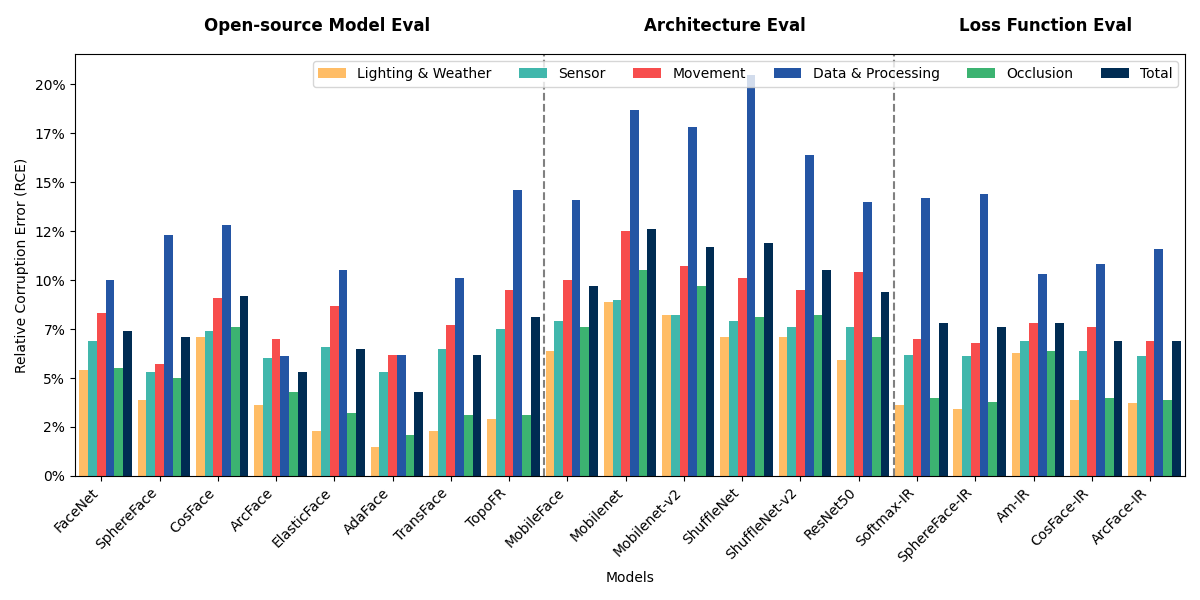}
   \caption{RCE results on CFP-C.}
   \label{fig:RCE_CFP_c}
\end{figure*}

\begin{figure*}[h]
  \centering
   \includegraphics[width=0.99\linewidth]{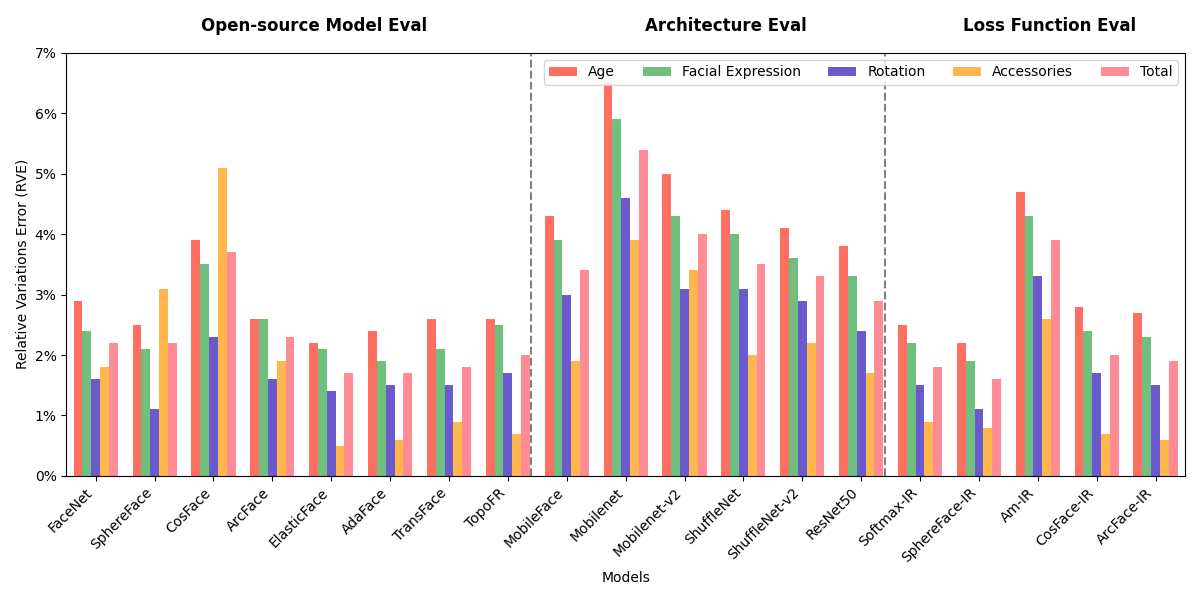}
   \caption{RVE results on CFP-V.}
   \label{fig:RVE_CFP_v}
\end{figure*}

For CFP, we follow the experimental setup used for LFW-C/V and test 19 FR models on 20 types of common corruptions and 10 types of appearance variations, categorized into \textit{Open-source Model Eval}, \textit{Architecture Eval}, and \textit{Loss Function Eval}. The test results are shown in Tab.~\ref{tab:accuracy_CFP_C} and Tab.~\ref{tab:accuracy_CFP_v}. We also report the \textit{Relative Corruption Error (RCE)}, as shown in Fig.~\ref{fig:RCE_CFP_c}, for each model across different corruption categories. Additionally, in Appearance Variations, we further illustrate the \textit{Relative Variation Error (RVE)}, as shown in Fig.~\ref{fig:RVE_CFP_v}, for each model across variation categories.

\begin{figure*}[h]
  \centering
   \includegraphics[width=0.99\linewidth]{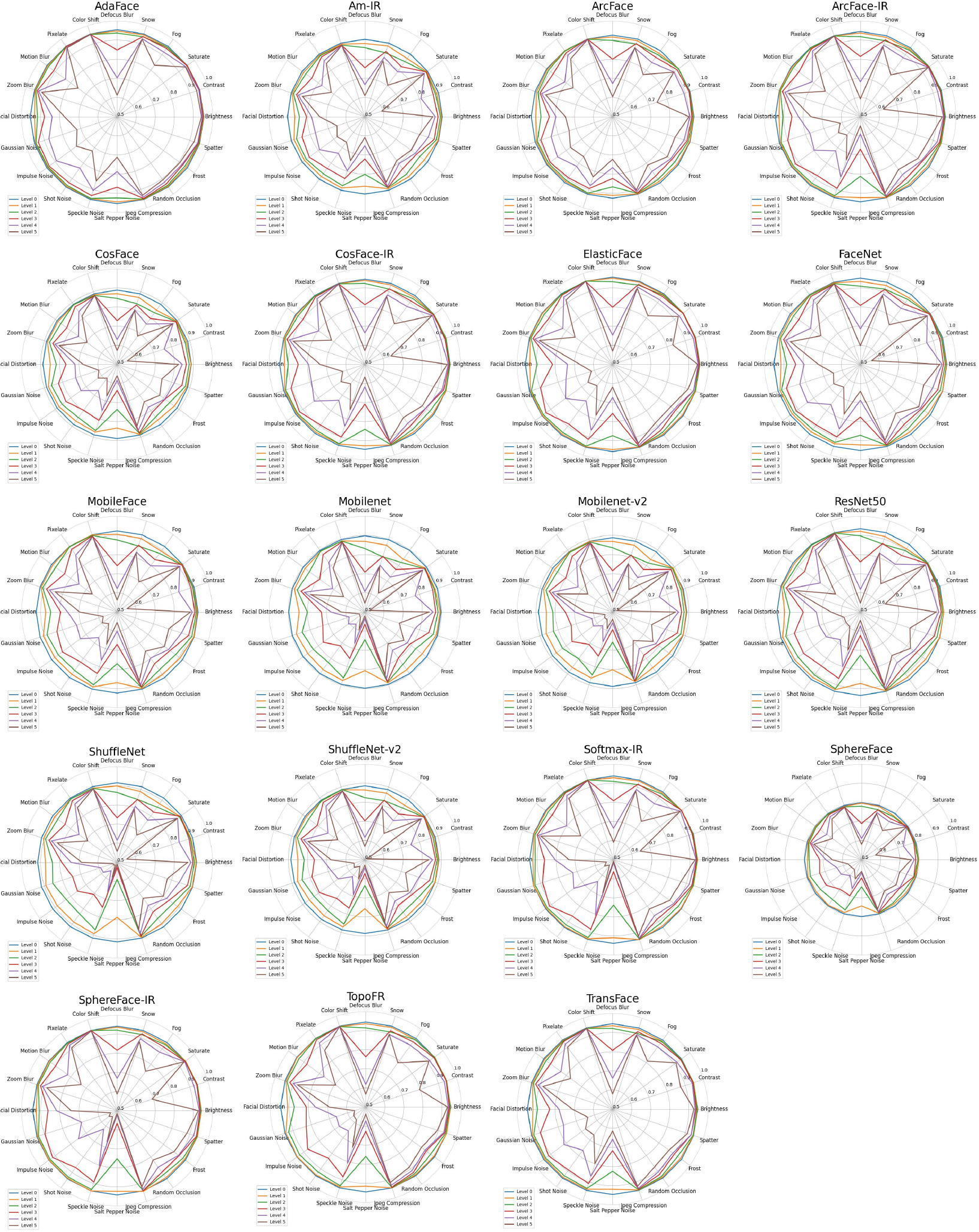}
   \caption{Graded radar charts for each model on CFP-C.}
   \label{fig:radar_CFP_c}
\end{figure*}

\begin{figure*}[h]
  \centering
   \includegraphics[width=0.99\linewidth]{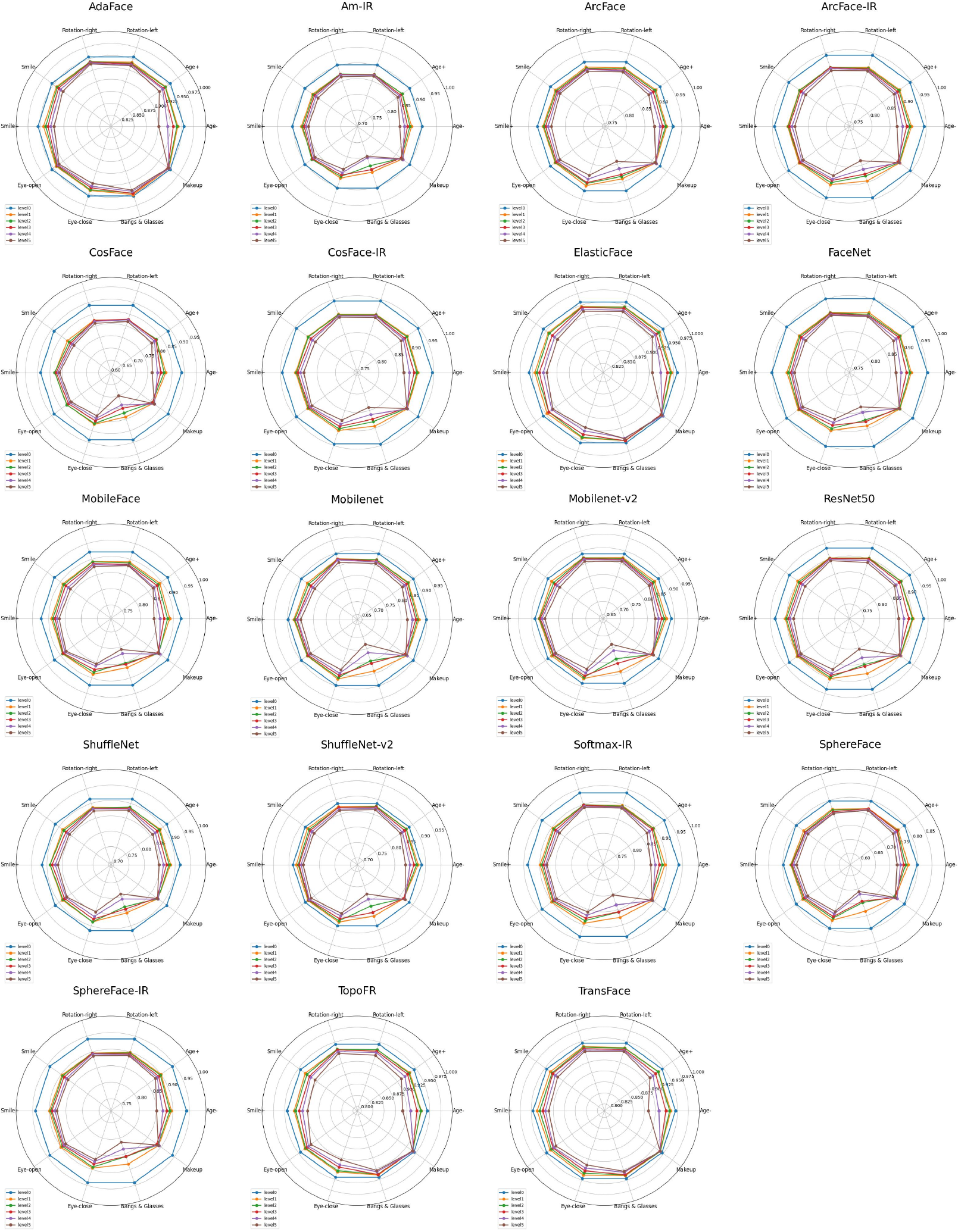}
   \caption{Graded radar charts for each model on CFP-V.}
   \label{fig:radar_CFP_v}
\end{figure*}

\begin{figure*}[h]
  \centering
   \includegraphics[width=0.99\linewidth]{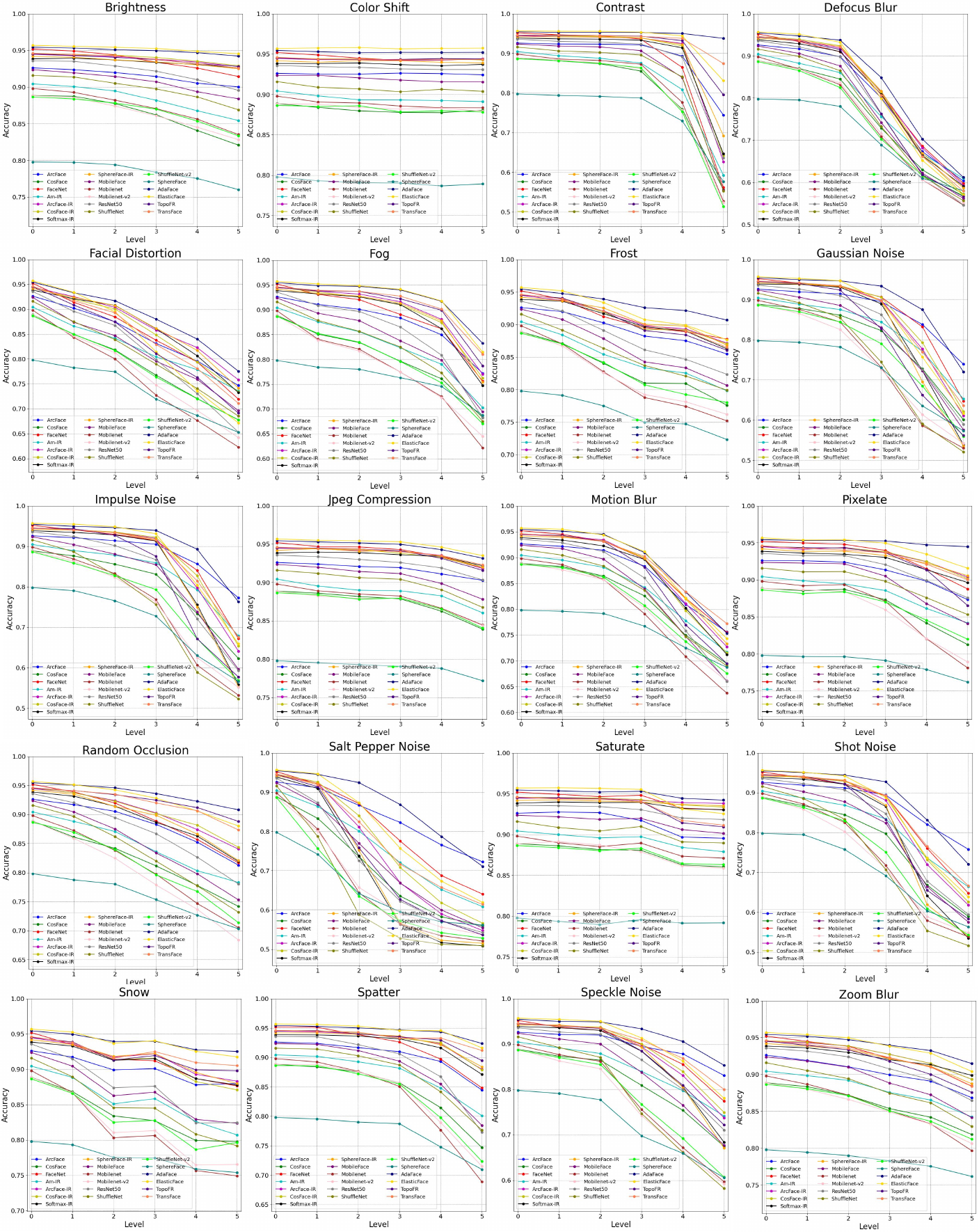}
   \caption{Graded line charts for each corruption on CFP-C.}
   \label{fig:line_CFP_c}
\end{figure*}

\begin{figure*}[h]
  \centering
   \includegraphics[width=0.99\linewidth]{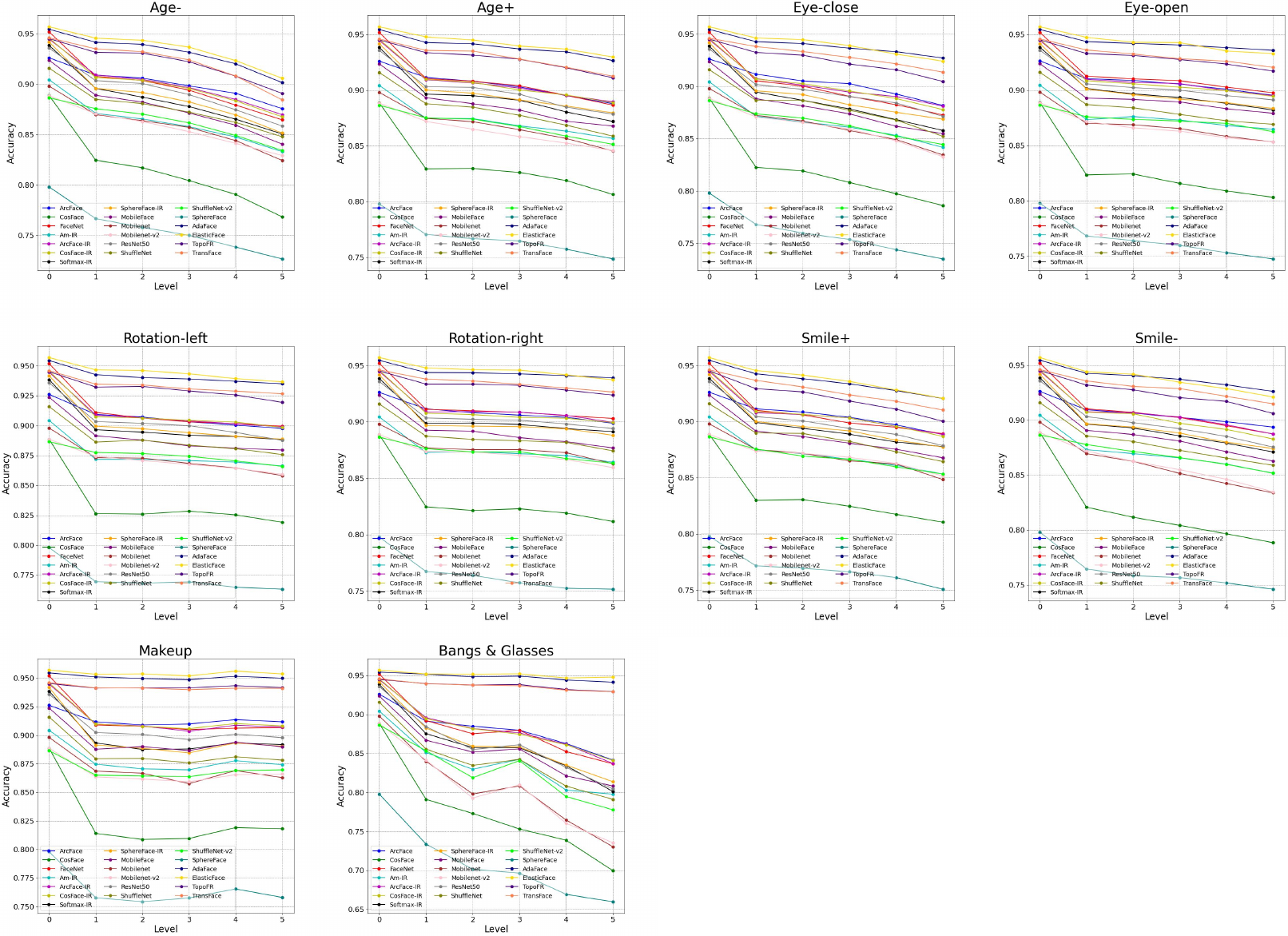}
   \caption{Graded line charts for each corruption on CFP-V.}
   \label{fig:line_CFP_v}
\end{figure*}

We also provide the specific data tables corresponding to levels 1-5, as shown in Tab.~\ref{tab:accuracy_CFP_C_level1}, Tab.~\ref{tab:accuracy_CFP_C_level2}, Tab.~\ref{tab:accuracy_CFP_C_level3}, Tab.~\ref{tab:accuracy_CFP_C_level4}, Tab.~\ref{tab:accuracy_CFP_C_level5} and Tab.~\ref{tab:accuracy_CFP_v_level1}, Tab.~\ref{tab:accuracy_CFP_v_level2}, Tab.~\ref{tab:accuracy_CFP_v_level3}, Tab.~\ref{tab:accuracy_CFP_v_level4}, Tab.~\ref{tab:accuracy_CFP_v_level5}. We present the radar charts and line charts showing the 5 levels for each category, in Fig.~\ref{fig:radar_CFP_c}, Fig.~\ref{fig:radar_CFP_v} and Fig.~\ref{fig:line_CFP_c}, Fig.~\ref{fig:line_CFP_v}.

From the data, we observe that due to differences in data formats, the results on CFP-C/V show patterns that differ from those on LFW-C/V. Taking common corruptions as an example, overall, AdaFace still maintains the best robustness. However, at CFP-C level 1, we find that ElasticFace achieves the best results across all categories. This is because ElasticFace has a higher clean accuracy, and level 1 corruptions are relatively mild. From levels 2-5, AdaFace’s robustness gradually becomes more apparent. On CFP-V, ElasticFace, due to its higher clean accuracy, maintains the highest accuracy across levels 1-5, which aligns with our conclusion in the main text that clean accuracy has a significant impact on variations.

\section{Results on YTF-C/V}
\label{sup:D}

\begin{figure*}[h]
  \centering
   \includegraphics[width=0.99\linewidth]{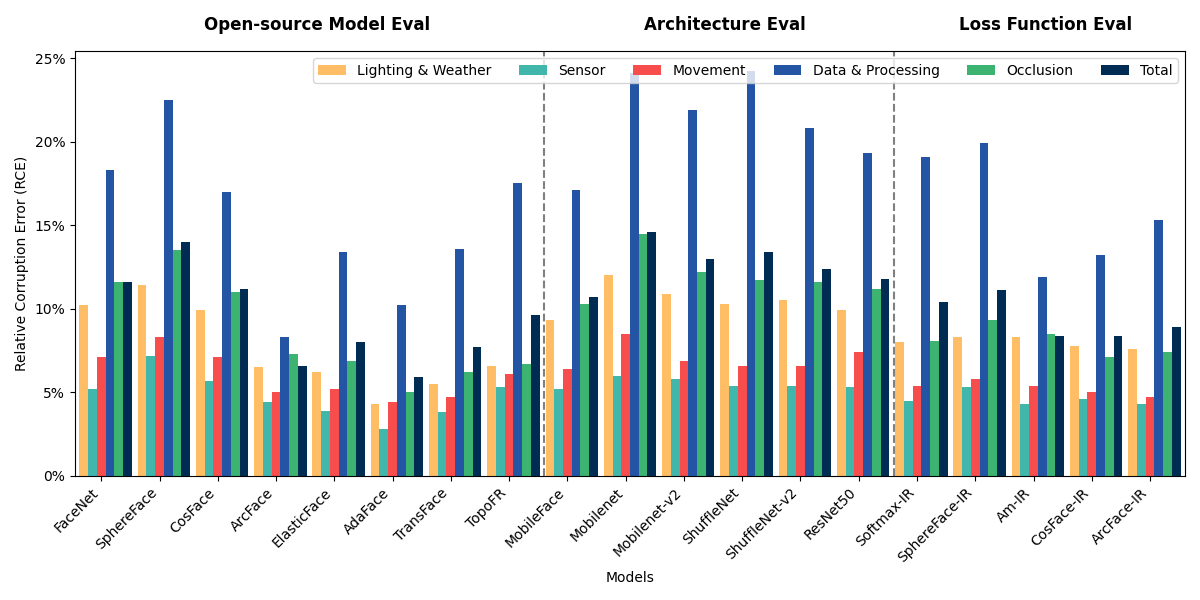}
   \caption{RCE results on YTF-C.}
   \label{fig:RCE_YTF_c}
\end{figure*}

\begin{figure*}[h]
  \centering
   \includegraphics[width=0.99\linewidth]{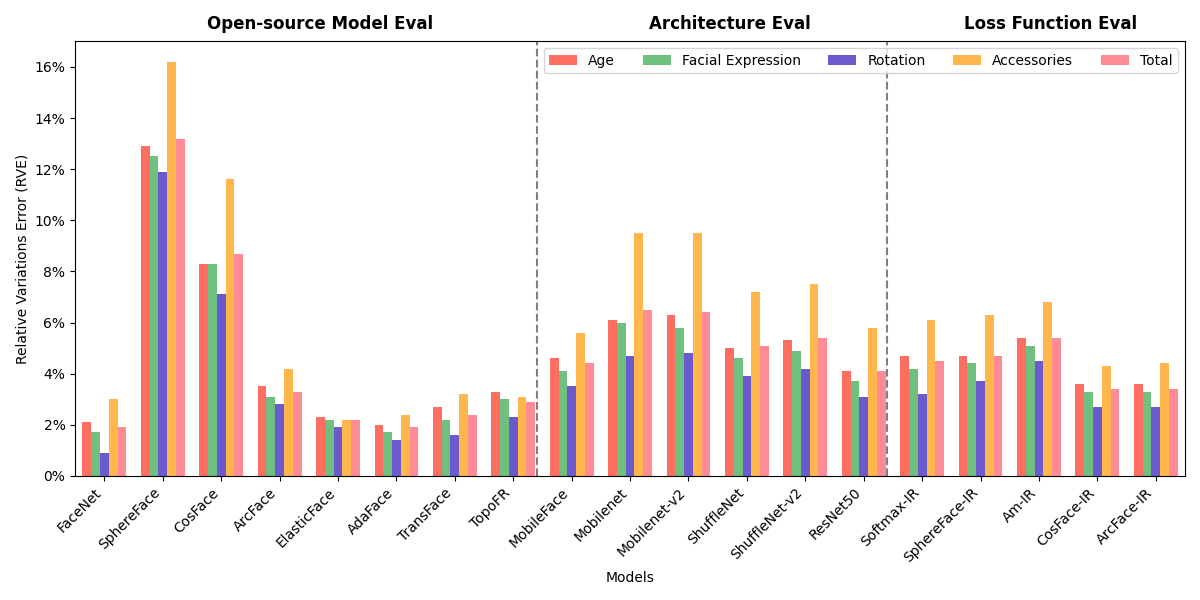}
   \caption{RVE results on YTF-V.}
   \label{fig:RVE_YTF_v}
\end{figure*}

\begin{figure*}[h]
  \centering
   \includegraphics[width=0.99\linewidth]{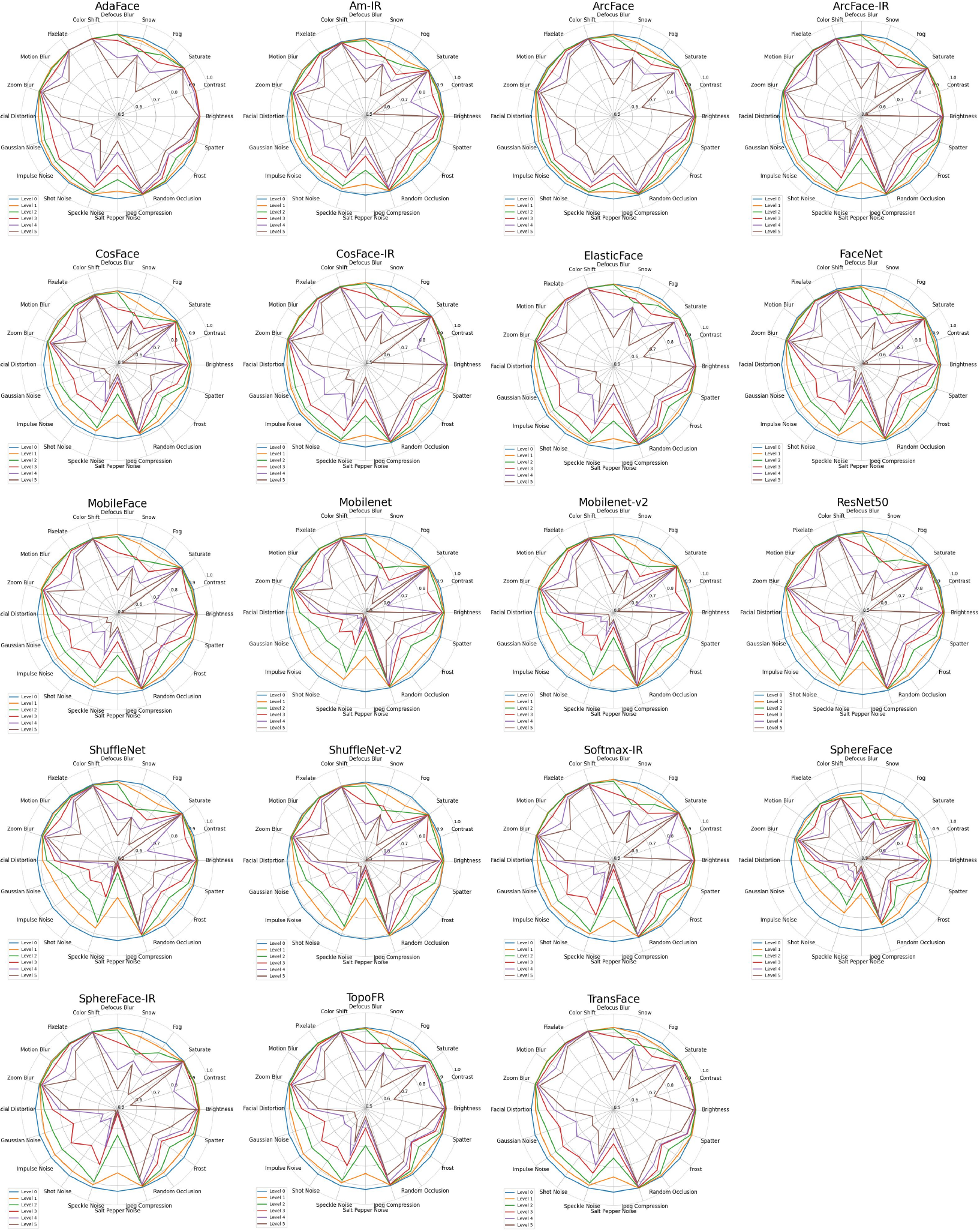}
   \caption{Graded radar charts for each model on YTF-C.}
   \label{fig:radar_YTF_c}
\end{figure*}

\begin{figure*}[h]
  \centering
   \includegraphics[width=0.99\linewidth]{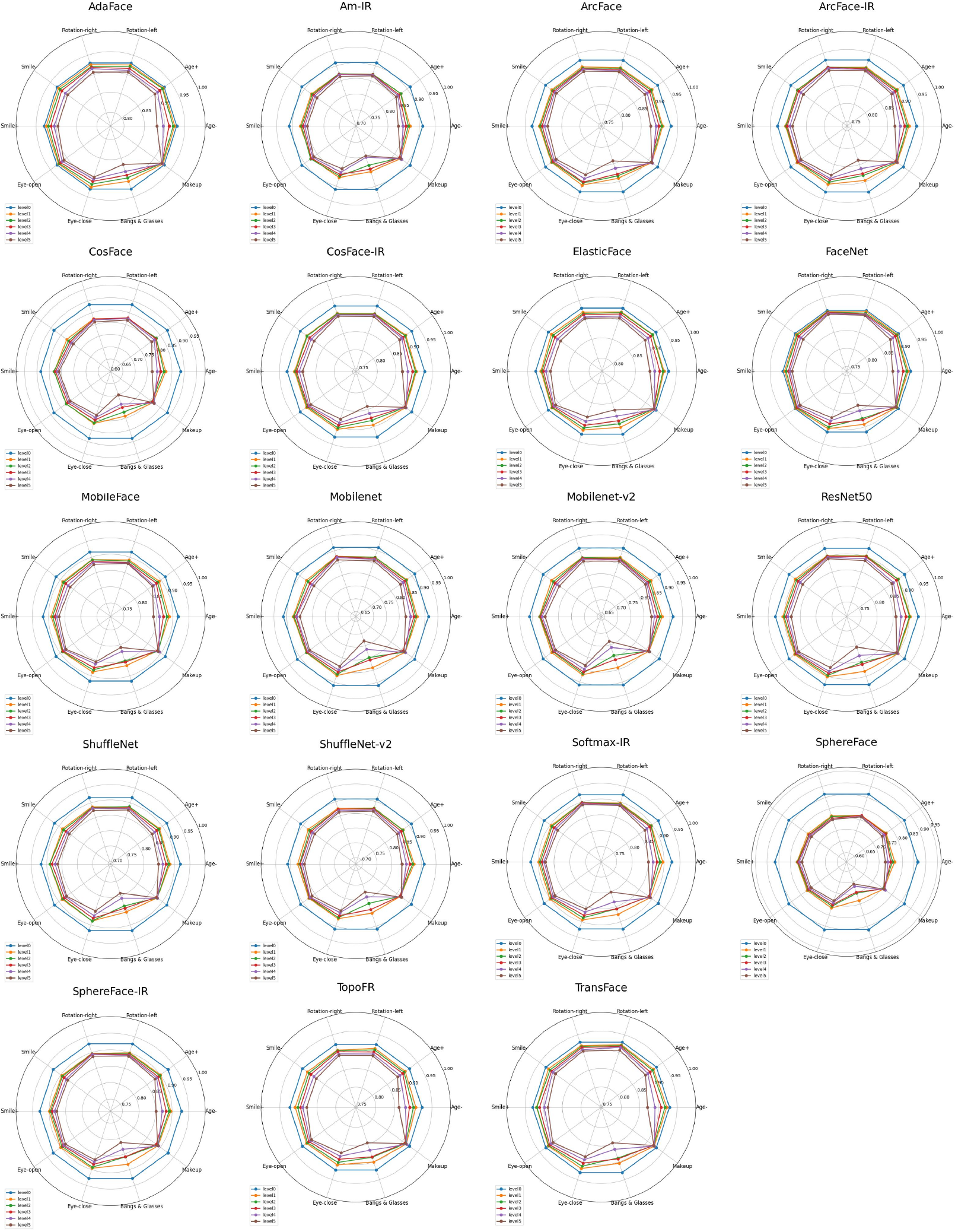}
   \caption{Graded radar charts for each model on YTF-V.}
   \label{fig:radar_YTF_v}
\end{figure*}

\begin{figure*}[h]
  \centering
   \includegraphics[width=0.99\linewidth]{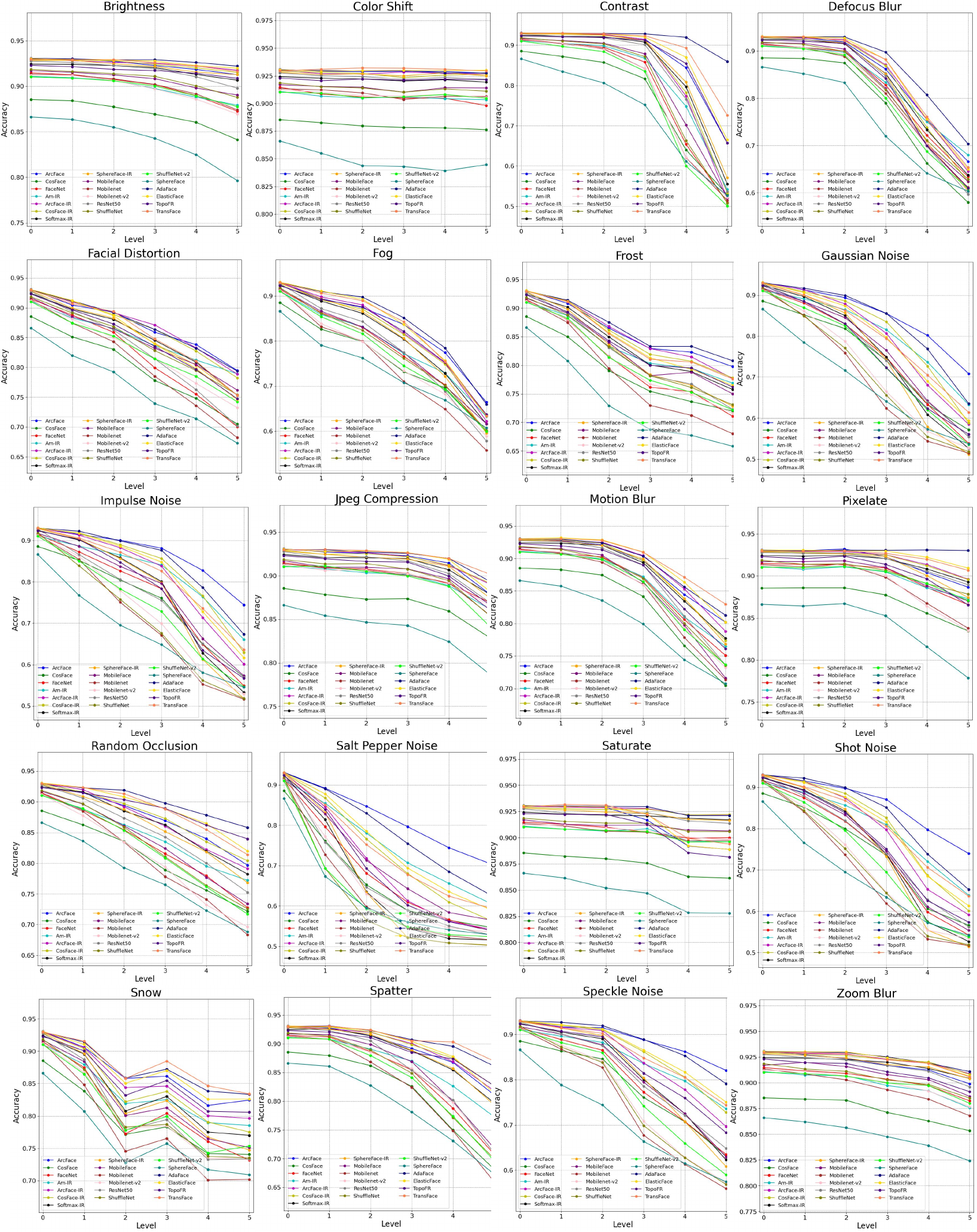}
   \caption{Graded line charts for each corruption on YTF-C.}
   \label{fig:line_YTF_c}
\end{figure*}

\begin{figure*}[h]
  \centering
   \includegraphics[width=0.99\linewidth]{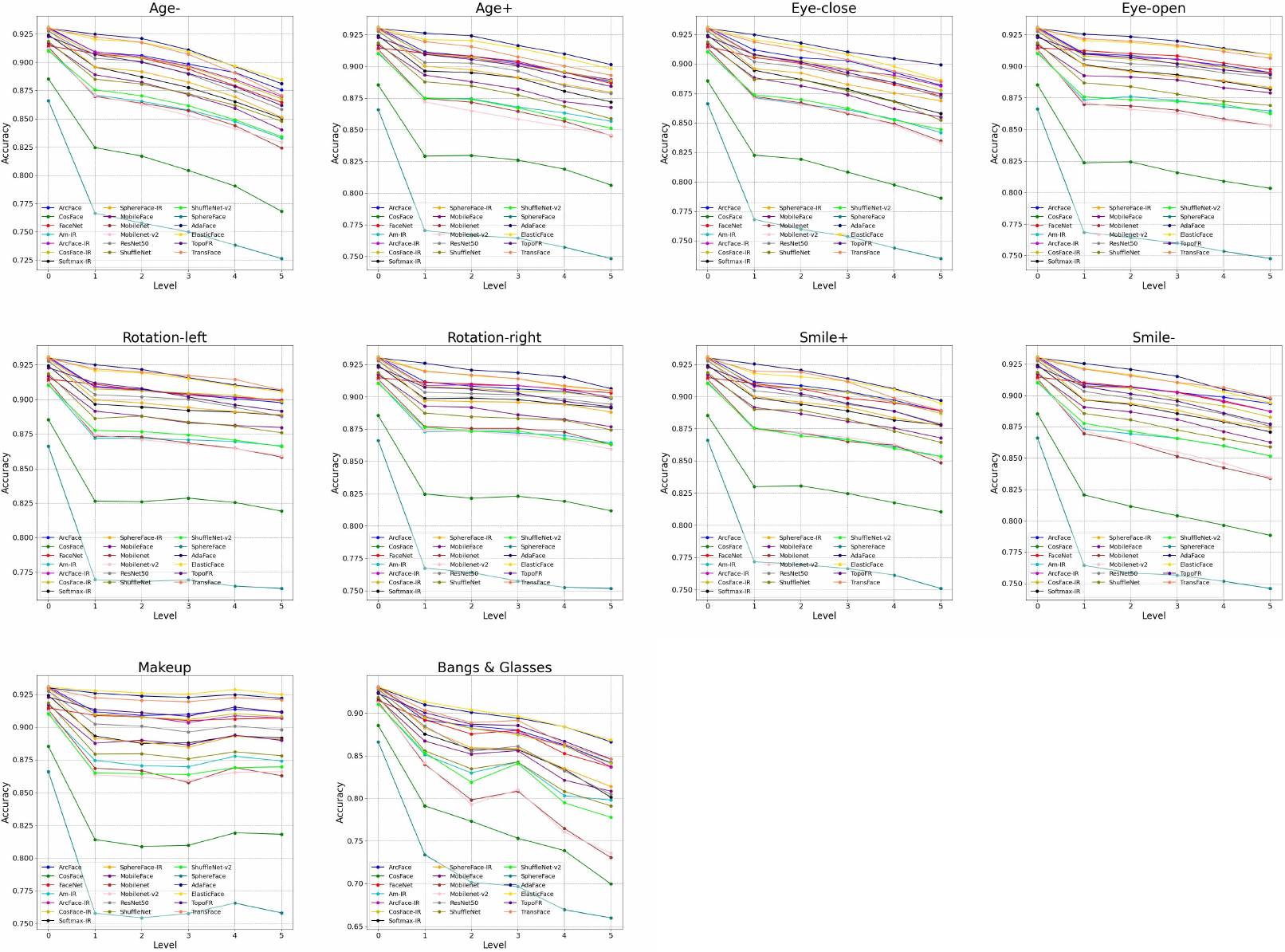}
   \caption{Graded line charts for each corruption on YTF-V.}
   \label{fig:line_YTF_v}
\end{figure*}

For YTF, we also provide the average accuracy data on YTF-C and YTF-V, shown in Tab.~\ref{tab:accuracy_YTF-C} and Tab.~\ref{tab:accuracy_YTF_v}, respectively. We similarly report the \textit{Relative Corruption Error (RCE)}, as shown in Fig.~\ref{fig:RCE_YTF_c}, and the \textit{Relative Variation Error (RVE)}, as shown in Fig.~\ref{fig:RVE_YTF_v}.
Additionally, we provide the specific data tables corresponding to levels 1-5, as shown in Tab.~\ref{tab:accuracy_YTF_C_level1}, Tab.~\ref{tab:accuracy_YTF_C_level2}, Tab.~\ref{tab:accuracy_YTF_C_level3}, Tab.~\ref{tab:accuracy_YTF_C_level4}, Tab.~\ref{tab:accuracy_YTF_C_level5} and Tab.~\ref{tab:accuracy_YTF_v_level1}, Tab.~\ref{tab:accuracy_YTF_v_level2}, Tab.~\ref{tab:accuracy_YTF_v_level3}, Tab.~\ref{tab:accuracy_YTF_v_level4}, Tab.~\ref{tab:accuracy_YTF_v_level5}. Furthermore, we present the radar charts and line charts showing the 5 levels for each category, in Fig.~\ref{fig:radar_YTF_c}, Fig.~\ref{fig:radar_YTF_v} and Fig.~\ref{fig:line_YTF_c}, Fig.~\ref{fig:line_YTF_v}.

For the YTF data, our findings are as follows: First, the FR models' clean accuracy on YTF is lower than on LFW and CFP. This is because YTF contains more complex data from various scenes, which is reflected in the final results. As shown in Tab.~\ref{tab:accuracy_YTF-C}, unlike LFW-C and CFP-C, the differences between models on YTF-C are more pronounced. For example, ArcFace achieves the highest clean accuracy and robustness in the Data \& Processing category, while TransFace demonstrates advantages in specific categories like Snow, Color Shift, and Motion Blur.

On YTF-V, due to the more complex scenes of YTF, the performance drop caused by variations is more severe, especially for ArcFace and ElasticFace. Although they achieve higher clean accuracy, their performance suffers more under variations. On the other hand, AdaFace, despite not having the highest clean accuracy, maintains the best robust accuracy under appearance variations. This finding slightly differs from the conclusions drawn on LFW and CFP.

\begin{figure*}[h]
  \centering
   \includegraphics[width=0.99\linewidth]{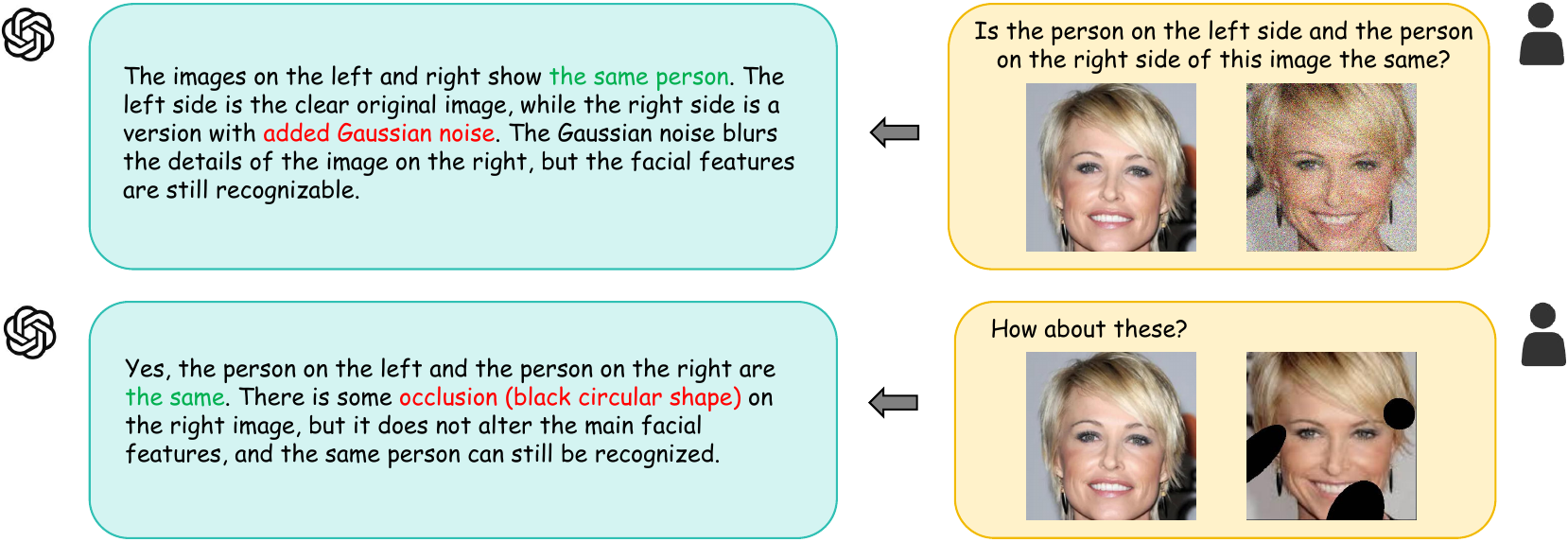}
   \caption{Testing GPT-4o mini for FR. GPT-4o mini demonstrates robust FR capabilities. 
   }
   \label{fig:gpt_demo}
      \vspace{-2ex}
\end{figure*}

\begin{figure*}[!th]
  \centering
   \includegraphics[width=0.99\linewidth]{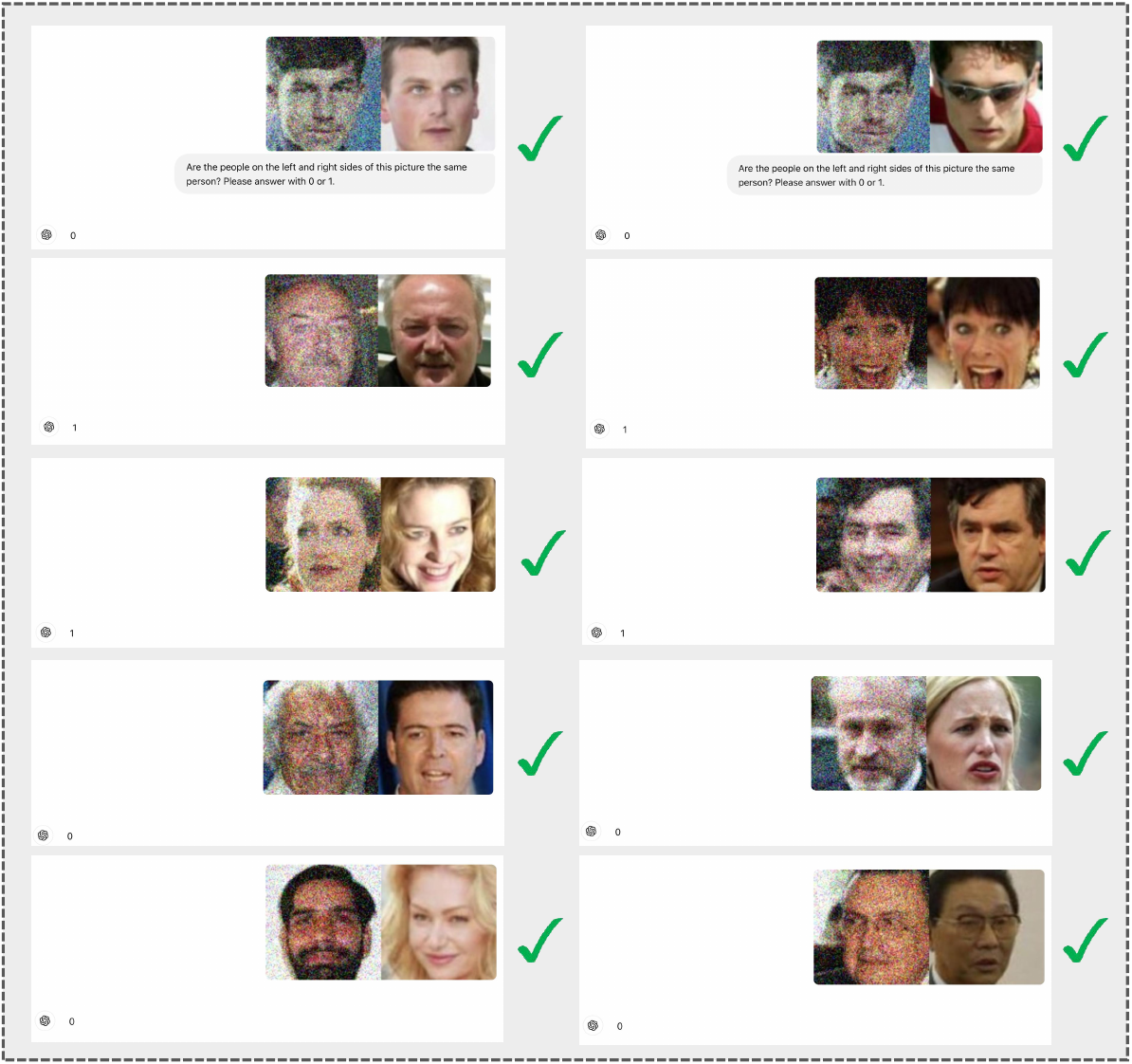}
   \caption{Additional results on testing GPT-4o-mini for FR.}
   \label{fig:gpt_demo_sup}
\end{figure*}


\begin{figure*}[!th]
  \centering
   \includegraphics[width=0.99\linewidth]{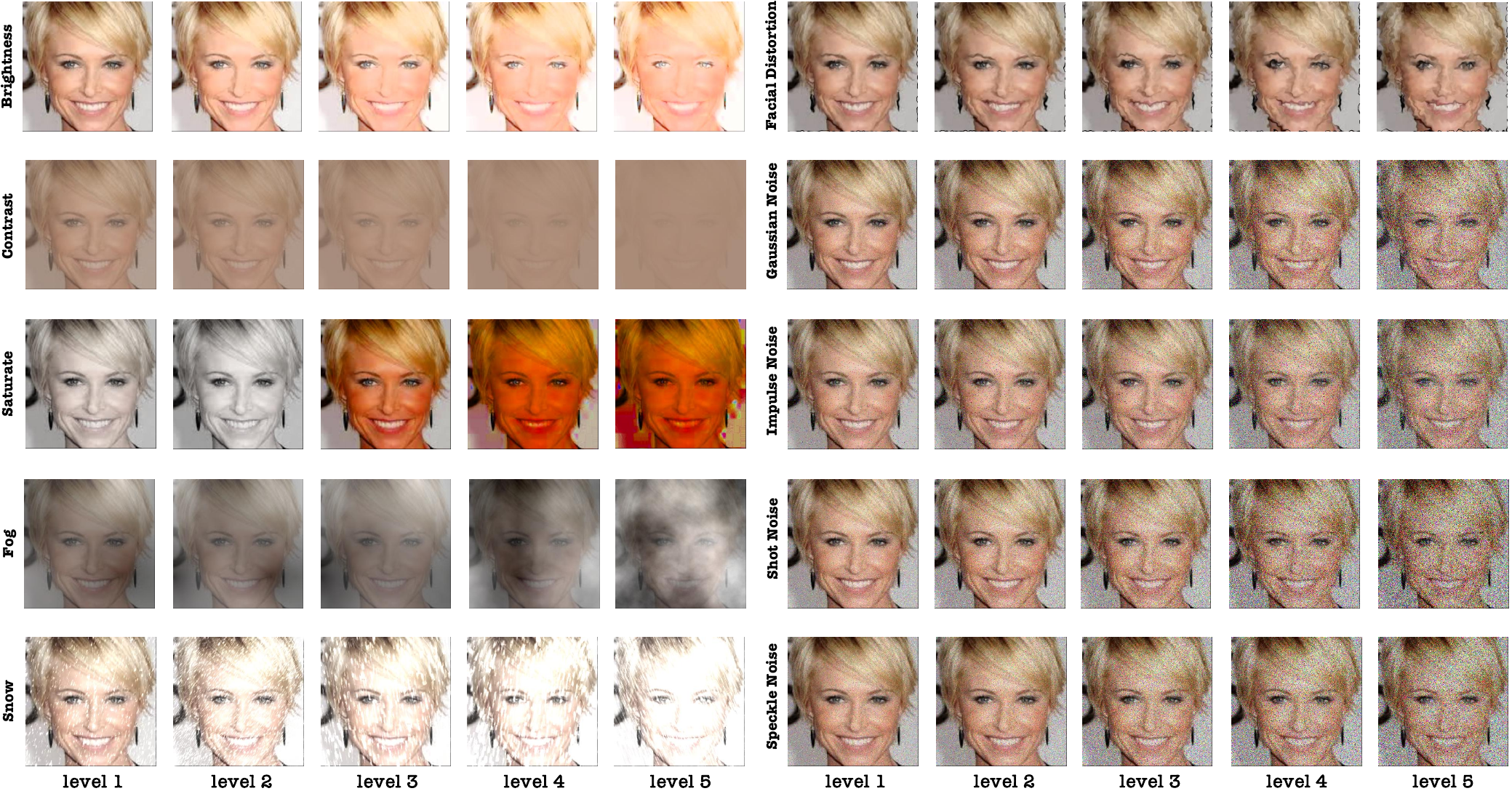}
   \caption{Full visualization of common corruptions severity levels (part 1).}
   \label{fig:corruptions_demo_sup_partA}
\end{figure*}

\begin{figure*}[!th]
  \centering
   \includegraphics[width=0.99\linewidth]{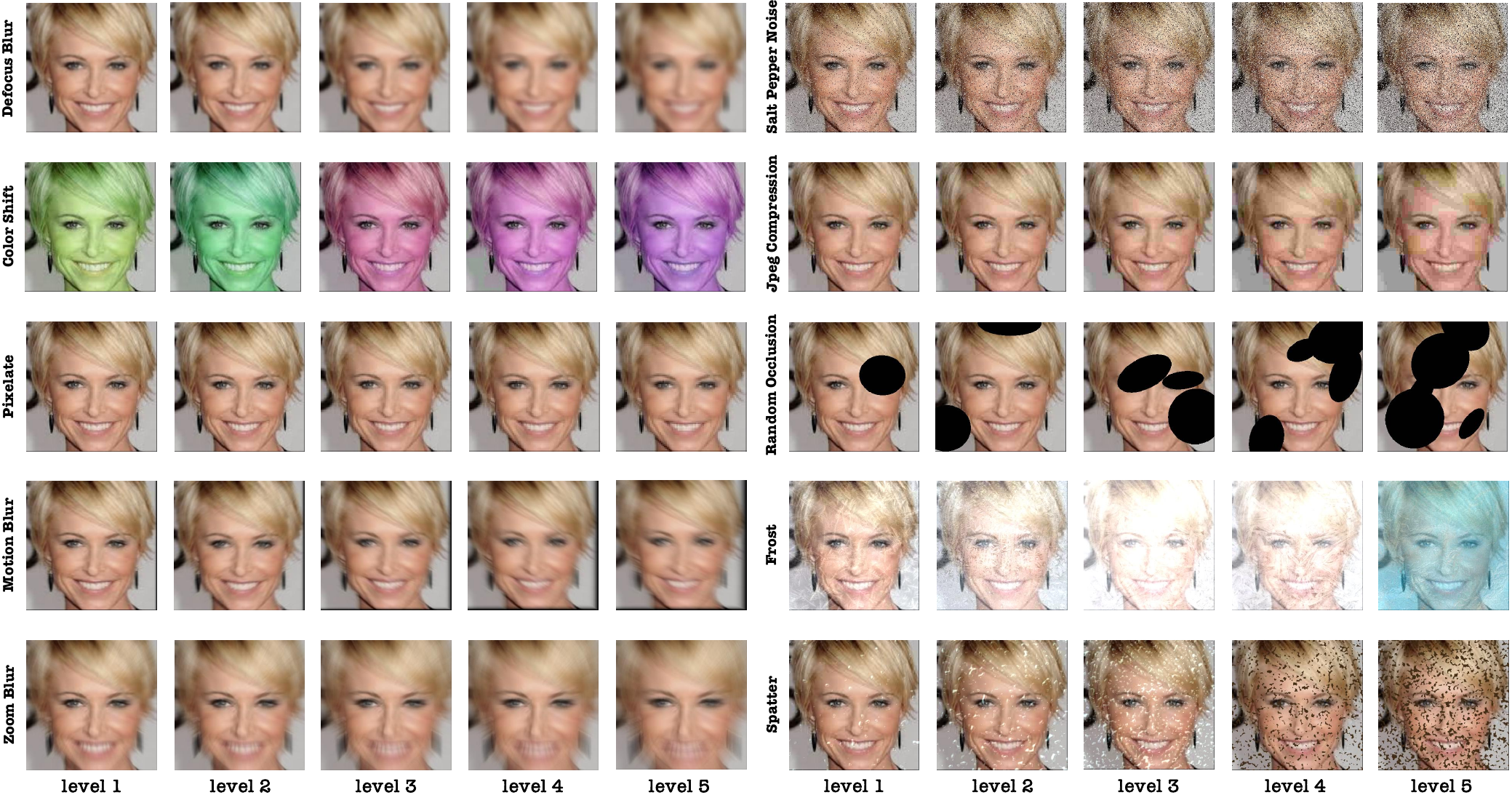}
   \caption{Full visualization of common corruptions severity levels (part 2).}
   \label{fig:corruptions_demo_sup_partB}
\end{figure*}

\begin{figure*}[!th]
  \centering
   \includegraphics[width=0.99\linewidth]{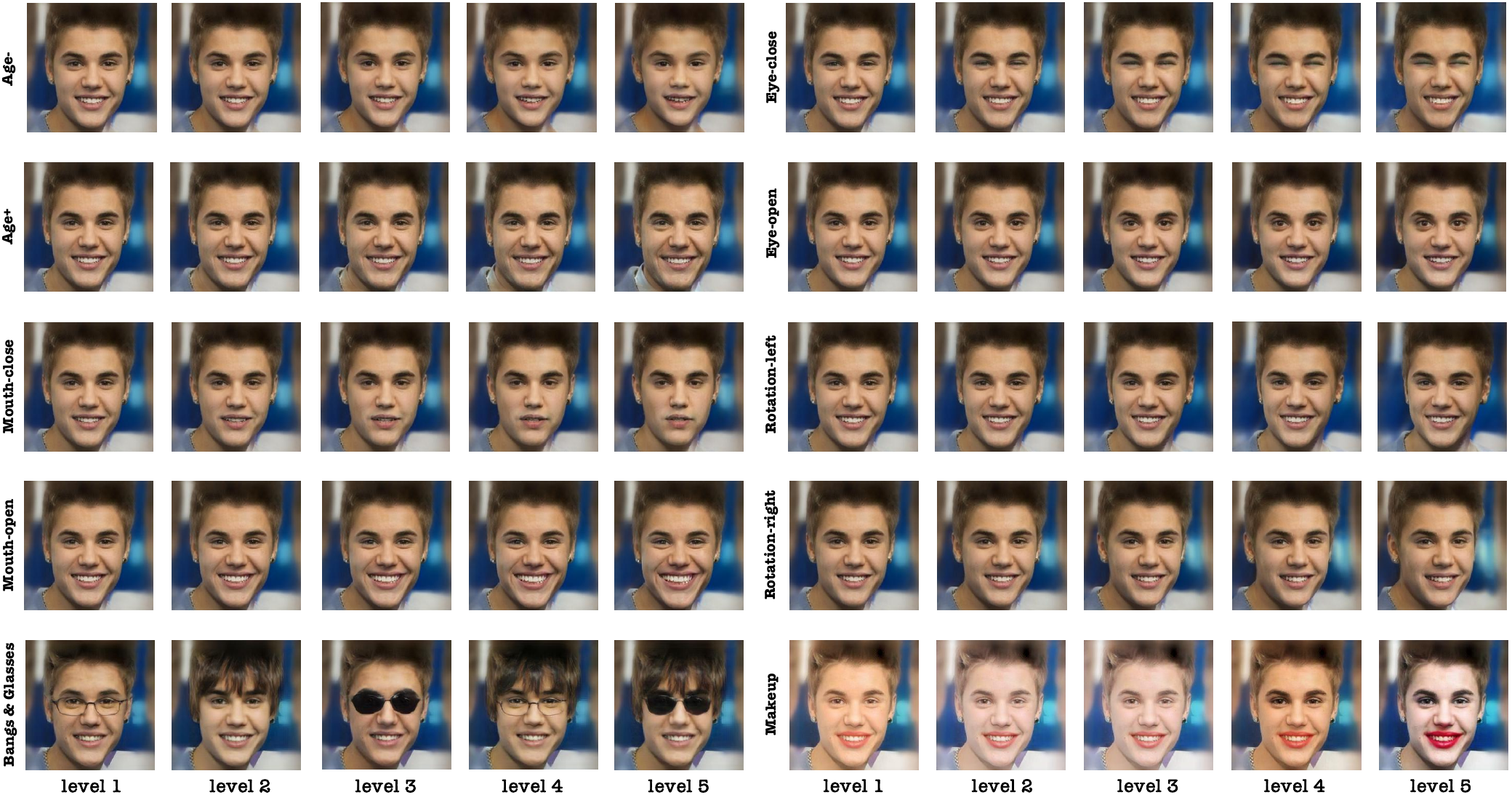}
   \caption{Full visualization of appearance variations severity levels.}
   \label{fig:variations_demo_sup}
\end{figure*}


\section{Exploration of Potential Defense}
\label{sup:F}

We further explore potential defense measures to enhance robustness. Based on the methods described in~\cite{yang2020robfr}, we employ existing defense strategies available, which can be categorized into two main approaches: \textit{input transformation} (e.g., R\&P~\cite{xie2017mitigating}, Bit-Red~\cite{xu2017feature}, JPEG~\cite{dziugaite2016study}) and \textit{adversarial training} (e.g., PGD-AT ~\cite{mkadry2017towards}, TRADES~\cite{zhang2019theoretically}). We test 10 robust models based on these approaches, as shown in Tab~.\ref{tab:robust_FR}. However, as we observe, for input transformation methods, there is no significant impact on the model's performance. Specifically, Bit-Red~\cite{xu2017feature} increases the average accuracy of Softmax-IR by 0.02, while Bit-Red~\cite{xu2017feature} and JPEG~\cite{dziugaite2016study} result in slight performance degradation. For PGD-AT~\cite{mkadry2017towards} and TRADES~\cite{zhang2019theoretically}, which are designed for adversarial training, no gains are observed under natural corruptions during testing. On the contrary, due to a notable drop in clean data performance after adversarial training, the overall model performance declines. 

Furthermore, we explored several advanced restoration methods~\cite{wang2021towards, zhou2022towards, lin2024diffbir}, based on GANs, Transformers, and Diffusion models, as potential defense strategies against OOD scenarios. Our experiments yielded interesting results, as summarized in Tab.~\ref{tab:restore}, where we report the changes in OOD robust accuracy for different FR models after applying these defense methods.

\textbf{From the Perspective of OOD Category:}
We observe that for certain OOD categories, the FR model’s recognition performance improved after applying restoration methods, particularly for noise-related corruptions under the Data \& Processing category. This is likely because restoration methods are primarily designed for denoising and deblurring, leading to notable improvements in handling noise-related corruptions, as shown in the left panel of Fig.~\ref{fig:restore}. However, for other OOD categories such as Lighting \& Weather and Occlusion, recognition performance actually degraded. This could be due to restoration methods distorting key facial landmarks or interfering with feature extraction, as illustrated in the right panel of Fig.~\ref{fig:restore}.

\textbf{From the Perspective of FR Models:}
An interesting phenomenon we observed is that models achieving the best performance in OOD testing, such as AdaFace, exhibited little to no improvement across all OOD categories after applying restoration-based defenses. This may be due to AdaFace’s inherent robustness, which already enables it to mitigate noise and other perturbations. However, the restoration methods inadvertently distorted some of the original facial landmarks, disrupting AdaFace’s feature extraction process. As a result, the performance degradation caused by these distortions outweighed the potential benefits of restoration, ultimately leading to a decline in FR accuracy.
Conversely, for models that are more vulnerable to OOD challenges, such as SphereFace, restoration-based defenses led to improvements in most categories, with a substantial increase in overall accuracy. The Data \& Processing category, which poses the greatest threat to FR performance, presented significant challenges for less robust models. In these cases, the removal of noise through restoration methods provided a notable performance gain, effectively enhancing the model’s robustness.

Therefore, employing restoration methods as a defense strategy could be a promising solution for addressing OOD scenarios, particularly for noise-related corruptions that pose significant challenges. However, since existing restoration techniques are primarily designed for denoising and deblurring, their effectiveness remains limited when facing a broader range of real-world OOD challenges, as explored in OODFace. In such cases, the interference caused by restoration methods in the feature extraction process introduces new challenges, leading to performance degradation in originally robust models.

Notably, none of the individual defense methods in our experiments were able to effectively mitigate more than 30\% of the OOD categories. This suggests the need for more generalized and precise defense strategies. These results indicate that existing defense strategies are insufficient for effectively handling comprehensive out-of-distribution scenarios. Enhancing the overall robustness of facial recognition models remains an open challenge for future research.

\section{VLMs as Potential Solutions}
\label{sup:G}

In Sec.~\ref{sec:5.4} of the main paper, we explored the potential of Vision-Language Models (VLMs) for addressing FR OOD scenarios, a direction that remains largely unexplored. Our evaluation of GPT-4o-mini revealed strong FR capabilities, including the ability to recognize specific corruption types (see Fig~\ref{fig:gpt_demo_sup} for examples). Further, we tested both closed-source commercial models (GPT-4o-mini, Qwen-VL-Plus) and open-source models (LLaVA-NeXT-LLaMA3-8B, InternVL2.5-8B) on face images corrupted with 20 common corruptions designed in our benchmark. The prompt used was:
\begin{tcolorbox}[colframe=black, colback=gray!8, coltitle=black, sharp corners=all, boxrule=0.5mm, boxsep=0.1mm]
``Determine if the faces in the image belong to the same person. Reply with 1 if they are the same, and 0 otherwise. Your answer must be either 0 or 1."
\end{tcolorbox}

This setup requires the models to directly output a binary decision on face identity matching. Notably, open-source models fail to make reliable predictions, outputting nearly all 0s, while closed-source models achieve promising results. Qwen-VL-Plus reaches 87.76\% accuracy, and GPT-4o-mini achieves an impressive 98.98\%. 

Although our experimental results suggest that large models’ generalization ability could be a promising solution for handling OOD scenarios, their practical application still faces certain limitations and challenges. Notably, only proprietary models achieve strong performance, raising concerns regarding accessibility and deployment constraints. Additionally, the use of facial data for model training introduces significant security and privacy considerations that must be seriously addressed.  
Furthermore, understanding why VLMs achieve superior FR performance and how to effectively integrate VLMs into existing FR pipelines to enhance robustness remain important directions for future research.

\section{Additional Results on Face Masks}
\label{sup:E}

In the field of face recognition, masks typically refer to physical disguises used to deceive or conceal the identity of the wearer, preventing accurate recognition by the system. As an extension, we conduct physical experiments with masks.

In our experiments, we create five different types of masks using various collection methods and fabrication processes. We categorize them from \textit{A} to \textit{E} based on their realism, with \textit{Mask A} being directly collected through photography and created using 2D printing technology with paper material. \textit{Masks B} and \textit{Masks C} are created by scanning facial data using 3D scanning technology, with sandstone and paper pulp materials, respectively. \textit{Masks D} and \textit{Masks E} are also created using 3D scanning and silicone material, with \textit{Masks D} and \textit{E} having higher facial fit and elasticity.  

Then, we have another person wear the masks and conduct data collection using a robotic arm, as shown in Fig.~\ref{fig:robotic_arm}. The collected video data is captured from certain angles. We then extract video frames, associate them with the corresponding mask IDs as positive examples, and pair them with the person wearing the mask as negative examples to generate test data.

Subsequently, we test these masks on the Common Corruptions and Appearance Variations we design, in order to examine the characteristics of different masks and their impact in OOD scenarios. The results are shown in Tab.~\ref{tab:accuracy_c_Mask-A}, Tab.~\ref{tab:accuracy_c_Mask-B}, Tab.~\ref{tab:accuracy_c_Mask-C}, Tab.~\ref{tab:accuracy_c_Mask-D}, Tab.~\ref{tab:accuracy_c_Mask-E} and Tab.~\ref{tab:accuracy_v_Mask-A}, Tab.~\ref{tab:accuracy_v_Mask-B}, Tab.~\ref{tab:accuracy_v_Mask-C}, Tab.~\ref{tab:accuracy_v_Mask-D}, Tab.~\ref{tab:accuracy_v_Mask-E}. 
The data reveals that
face masks demonstrate material-dependent vulnerability patterns in OOD scenarios, exhibiting degradation trends distinct from real faces. This observation suggests a potential avenue for enhancing spoof detection.

\section{More Visualization Results}
\label{sup:H}

To thoroughly assess the robustness of FR systems, in Sec.\ref{sec:3.1}, we follow~\cite{hendrycks2019benchmarking} to define five severity levels for each type of corruption, with the common corruptions categorized into five levels ranging from level 1 (mild) to level 5 (extreme). In Sec.\ref{sec:3.2}, we define five severity levels for appearance variations, with each level representing an increase in perceived appearance interference. In this section, we present the visualization results for all categories at severity levels 1 to 5, including 20 types of common corruptions as shown in  Fig.~\ref{fig:corruptions_demo_sup_partA}, Fig.~\ref{fig:corruptions_demo_sup_partB}, and 10 types of appearance variations as shown in Fig.~\ref{fig:variations_demo_sup}.

\section{Statement on Data Sources and User Privacy}
We would like to clarify that all images used in our experiments originate from publicly available open-source datasets that comply with the relevant data usage policies. No private or personally identifiable images were uploaded to any third-party commercial system or Vision-Language Models. Additionally, for the face images collected in our physical experiments, we obtained explicit consent from the participants for their data to be used in our research. Our study strictly adheres to ethical guidelines and privacy regulations to ensure the responsible handling of all data.

\clearpage

\begin{table*}[]
\centering
   \scalebox{0.47}{
\setlength{\tabcolsep}{2pt}  
\begin{tabular}{c|c|cccccccc|cccccc|ccccc}
\hline
\multicolumn{2}{c}{Models}                               & \multicolumn{8}{c}{Open-source Model Eval}                                                                                & \multicolumn{6}{c}{Architecture Eval}                                                                         & \multicolumn{5}{c}{Loss Function Eval}                                     \\
\multicolumn{2}{c}{Corruptions}                          & FaceNet        & SphereFace & CosFace & ArcFace        & ElasticFace& AdaFace        & TransFace      & TopoFR & MobileFace     & Mobilenet & Mobilenet-v2 & ShuffleNet& ShuffleNet-v2& ResNet50       & Softmax-IR & SphereFace-IR  & Am-IR          & CosFace-IR     & ArcFace-IR     \\ \hline\hline
\rowcolor{lightgray}\multicolumn{2}{c}{None (clean)}                         & 99.23   & 98.20      & 98.63   & 99.50   & 99.80            & \textbf{99.83} & 99.75          & 99.78       & 99.43          & 99.40     & 99.10       & 99.43                  & 99.15                          & \textbf{99.72} & 99.53      & 99.57         & 99.18 & \textbf{99.70} & 99.67          \\
\multirow{5}{*}{Lighting \& Weather} & Brightness        & 99.18   & 98.10      & 98.57   & 99.43   & 99.82            & \textbf{99.83} & 99.75          & 99.77       & 99.40          & 99.32     & 99.05       & 99.40                  & 99.12                          & \textbf{99.70} & 99.52      & 99.58         & 99.02 & \textbf{99.70} & 99.65          \\
                                     & Contrast          & 98.93   & 97.38      & 98.07   & 99.38   & 99.82            & \textbf{99.83} & 99.70          & 99.75       & 99.28          & 99.17     & 98.90       & 99.32                  & 98.90                          & \textbf{99.58} & 99.53      & 99.50         & 99.00 & \textbf{99.70} & 99.62          \\
\hline
\end{tabular}
}
\caption{Accuracy of 19 FR models on LFW-C level 1.}
\label{tab:accuracy_LFW-C_level1}
\vspace{-2ex}
\end{table*}

\begin{table*}[]
\centering
   \scalebox{0.47}{
\setlength{\tabcolsep}{2pt}  
\begin{tabular}{c|c|cccccccc|cccccc|ccccc}
\hline
\multicolumn{2}{c}{Models}                               & \multicolumn{8}{c}{Open-source Model Eval}                                                                                & \multicolumn{6}{c}{Architecture Eval}                                                                         & \multicolumn{5}{c}{Loss Function Eval}                                     \\
\multicolumn{2}{c}{Corruptions}                          & FaceNet        & SphereFace & CosFace & ArcFace        & ElasticFace& AdaFace        & TransFace      & TopoFR & MobileFace     & Mobilenet & Mobilenet-v2 & ShuffleNet& ShuffleNet-v2& ResNet50       & Softmax-IR & SphereFace-IR  & Am-IR          & CosFace-IR     & ArcFace-IR     \\ \hline\hline
\rowcolor{lightgray}\multicolumn{2}{c}{None (clean)}                         & 99.23   & 98.20      & 98.63   & 99.50   & 99.80            & \textbf{99.83} & 99.75          & 99.78       & 99.43          & 99.40     & 99.10       & 99.43                  & 99.15                          & \textbf{99.72} & 99.53      & 99.57         & 99.18          & \textbf{99.70} & 99.67          \\
\multirow{5}{*}{Lighting \& Weather} & Brightness        & 99.03   & 97.70      & 98.45   & 99.42   & 99.78            & \textbf{99.83} & 99.72          & 99.73       & 99.35          & 99.22     & 98.92       & 99.32                  & 99.00                          & \textbf{99.63} & 99.48      & 99.55         & 98.82          & \textbf{99.67} & 99.63          \\
                                     & Contrast          & 98.47   & 96.32      & 97.72   & 99.32   & 99.78            & \textbf{99.83} & 99.65          & 99.72       & 99.15          & 98.90     & 98.57       & 99.17                  & 98.63                          & \textbf{99.48} & 99.45      & 99.37         & 98.67          & \textbf{99.67} & 99.60          \\
\hline
\end{tabular}
}
\caption{Accuracy of 19 FR models on LFW-C level 2.}
\label{tab:accuracy_LFW-C_level2}
\vspace{-2ex}
\end{table*}

\begin{table*}[]
\centering
   \scalebox{0.47}{
\setlength{\tabcolsep}{2pt}  
\begin{tabular}{c|c|cccccccc|cccccc|ccccc}
\hline
\multicolumn{2}{c}{Models}                               & \multicolumn{8}{c}{Open-source Model Eval}                                                                                & \multicolumn{6}{c}{Architecture Eval}                                                                         & \multicolumn{5}{c}{Loss Function Eval}                                     \\
\multicolumn{2}{c}{Corruptions}                          & FaceNet        & SphereFace & CosFace & ArcFace        & ElasticFace& AdaFace        & TransFace      & TopoFR & MobileFace     & Mobilenet & Mobilenet-v2 & ShuffleNet& ShuffleNet-v2& ResNet50       & Softmax-IR & SphereFace-IR  & Am-IR          & CosFace-IR     & ArcFace-IR     \\ \hline\hline
\rowcolor{lightgray}\multicolumn{2}{c}{None (clean)}                         & 99.23   & 98.20      & 98.63   & 99.50   & 99.80            & \textbf{99.83} & 99.75          & 99.78       & 99.43          & 99.40     & 99.10       & 99.43                  & 99.15                          & \textbf{99.72} & 99.53      & 99.57         & 99.18          & \textbf{99.70} & 99.67          \\
\multirow{5}{*}{Lighting \& Weather} & Brightness        & 98.47   & 96.53      & 97.75   & 99.28   & 99.75            & \textbf{99.80} & 99.63          & 99.60       & 99.13          & 98.87     & 98.27       & 99.05                  & 98.53                          & \textbf{99.48} & 99.25      & 99.38         & 98.55          & \textbf{99.60} & 99.55          \\
                                     & Contrast          & 96.85   & 92.40      & 95.98   & 99.05   & 99.73            & \textbf{99.82} & 99.53          & 99.62       & 98.68          & 97.87     & 97.22       & 98.57                  & 97.13                          & \textbf{99.13} & 99.08      & 99.08         & 97.68          & \textbf{99.52} & 99.47          \\
\hline
\end{tabular}
}
\caption{Accuracy of 19 FR models on LFW-C level 3.}
\label{tab:accuracy_LFW-C_level3}
\vspace{-2ex}
\end{table*}

\begin{table*}[]
\centering
   \scalebox{0.47}{
\setlength{\tabcolsep}{2pt}  
\begin{tabular}{c|c|cccccccc|cccccc|ccccc}
\hline
\multicolumn{2}{c}{Models}                               & \multicolumn{8}{c}{Open-source Model Eval}                                                                                & \multicolumn{6}{c}{Architecture Eval}                                                                         & \multicolumn{5}{c}{Loss Function Eval}                                     \\
\multicolumn{2}{c}{Corruptions}                          & FaceNet        & SphereFace & CosFace & ArcFace        & ElasticFace& AdaFace        & TransFace      & TopoFR & MobileFace     & Mobilenet & Mobilenet-v2 & ShuffleNet& ShuffleNet-v2& ResNet50       & Softmax-IR & SphereFace-IR  & Am-IR          & CosFace-IR     & ArcFace-IR     \\ \hline\hline
\rowcolor{lightgray}\multicolumn{2}{c}{None (clean)}                         & 99.23          & 98.20      & 98.63   & 99.50          & 99.80            & \textbf{99.83} & 99.75     & 99.78       & 99.43          & 99.40     & 99.10       & 99.43                  & 99.15                          & \textbf{99.72} & 99.53      & 99.57          & 99.18          & \textbf{99.70} & 99.67          \\
\multirow{5}{*}{Lighting \& Weather} & Brightness        & 97.70          & 94.47      & 96.20   & 98.85          & 99.67            & \textbf{99.73} & 99.45     & 99.37       & 98.60          & 97.90     & 97.22       & 98.57                  & 97.48                          & \textbf{98.90} & 98.88      & 99.02          & 97.58          & \textbf{99.37} & 99.33          \\
                                     & Contrast          & 78.67          & 74.83      & 76.77   & 94.87          & 98.65            & \textbf{99.73} & 98.92     & 97.87       & 86.67          & 73.62     & 73.78       & 82.62                  & 72.30                          & \textbf{92.10} & 93.27      & \textbf{94.08} & 89.82          & 93.47          & 92.85          \\
\hline
\end{tabular}
}
\caption{Accuracy of 19 FR models on LFW-C level 4.}
\label{tab:accuracy_LFW-C_level4}
\vspace{-2ex}
\end{table*}

\begin{table*}[]
\centering
   \scalebox{0.47}{
\setlength{\tabcolsep}{2pt}  
\begin{tabular}{c|c|cccccccc|cccccc|ccccc}
\hline
\multicolumn{2}{c}{Models}                               & \multicolumn{8}{c}{Open-source Model Eval}                                                                                & \multicolumn{6}{c}{Architecture Eval}                                                                         & \multicolumn{5}{c}{Loss Function Eval}                                     \\
\multicolumn{2}{c}{Corruptions}                          & FaceNet        & SphereFace & CosFace & ArcFace        & ElasticFace& AdaFace        & TransFace      & TopoFR & MobileFace     & Mobilenet & Mobilenet-v2 & ShuffleNet& ShuffleNet-v2& ResNet50       & Softmax-IR & SphereFace-IR  & Am-IR          & CosFace-IR     & ArcFace-IR     \\ \hline\hline
\rowcolor{lightgray}\multicolumn{2}{c}{None (clean)}                         & 99.23          & 98.20      & 98.63   & 99.50          & 99.80            & \textbf{99.83} & 99.75     & 99.78       & 99.43          & 99.40     & 99.10       & 99.43                  & 99.15                          & \textbf{99.72} & 99.53      & 99.57          & 99.18          & \textbf{99.70} & 99.67          \\
\multirow{5}{*}{Lighting \& Weather} & Brightness        & 96.60          & 91.58      & 93.85   & 98.28          & 99.50            & \textbf{99.63} & 99.33     & 99.02       & 97.80          & 96.15     & 95.45       & 97.47                  & 95.97                          & \textbf{98.23} & 98.27      & 98.33          & 96.45          & \textbf{99.10} & 98.92          \\
                                     & Contrast          & 50.27          & 59.13      & 52.83   & 71.97          & 79.57            & \textbf{98.37} & 86.85     & 76.82       & 52.53          & 50.82     & 51.18       & 51.43                  & 52.07                          & \textbf{57.73} & 58.28      & \textbf{59.85} & 58.53          & 55.65          & 54.67          \\
\hline
\end{tabular}
}
\caption{Accuracy of 19 FR models on LFW-C level 5.}
\label{tab:accuracy_LFW-C_level5}
\vspace{-2ex}
\end{table*}

\begin{table*}[]
\centering
 \scalebox{0.48}{
 \setlength{\tabcolsep}{2pt}  
\begin{tabular}{c|c|cccccccc|cccccc|ccccc}
\hline
\multicolumn{2}{c}{Models}                           & \multicolumn{8}{c}{Open-source Model Eval}                                                                                                                                        & \multicolumn{6}{c}{Architecture Eval}                                                                     & \multicolumn{5}{c}{Loss Function Eval}                           \\
\multicolumn{2}{c}{Variations}                          & FaceNet        & SphereFace & CosFace & ArcFace        & ElasticFace& AdaFace        & TransFace      & TopoFR & MobileFace     & Mobilenet & Mobilenet-v2 & ShuffleNet& ShuffleNet-v2& ResNet50       & Softmax-IR & SphereFace-IR  & Am-IR          & CosFace-IR     & ArcFace-IR     \\ \hline\hline
\rowcolor{lightgray} \multicolumn{2}{c}{None (clean)}                    & 99.23   & 98.20      & 98.63   & 99.50   & 99.80            & \textbf{99.83} & 99.75     & 99.78       & 99.43      & 99.40     & 99.10       & 99.43                  & 99.15                          & \textbf{99.72} & 99.53      & 99.57         & 99.18 & \textbf{99.70} & 99.67          \\
\multirow{2}{*}{Age}               & Age-           & 96.12   & 95.10      & 95.42   & 96.39   & 96.71            & \textbf{96.82} & 96.71     & 96.62       & 96.40      & 96.30     & 95.92       & 96.27                  & 95.94                          & \textbf{96.59} & 96.40      & 96.52         & 95.92 & \textbf{96.69} & 96.67          \\
                                   & Age+           & 96.00   & 94.95      & 95.50   & 96.34   & 96.74            & \textbf{96.81} & 96.65     & 96.60       & 96.30      & 96.24     & 96.05       & 96.32                  & 95.94                          & \textbf{96.55} & 96.45      & 96.54         & 95.92 & 96.59          & \textbf{96.62} \\
\hline
\end{tabular}
}
\caption{Accuracy of 19 FR models on LFW-V level 1.}
\label{tab:accuracy_LFW-v_level1}
   \vspace{-2ex}
\end{table*}

\begin{table*}[]
\centering
 \scalebox{0.48}{
 \setlength{\tabcolsep}{2pt}  
\begin{tabular}{c|c|cccccccc|cccccc|ccccc}
\hline
\multicolumn{2}{c}{Models}                           & \multicolumn{8}{c}{Open-source Model Eval}                                                                                                                                        & \multicolumn{6}{c}{Architecture Eval}                                                                     & \multicolumn{5}{c}{Loss Function Eval}                           \\
\multicolumn{2}{c}{Variations}                          & FaceNet        & SphereFace & CosFace & ArcFace        & ElasticFace& AdaFace        & TransFace      & TopoFR & MobileFace     & Mobilenet & Mobilenet-v2 & ShuffleNet& ShuffleNet-v2& ResNet50       & Softmax-IR & SphereFace-IR  & Am-IR          & CosFace-IR     & ArcFace-IR     \\ \hline\hline
\rowcolor{lightgray} \multicolumn{2}{c}{None (clean)}                    & 99.23   & 98.20      & 98.63   & 99.50   & 99.80            & \textbf{99.83} & 99.75     & 99.78       & 99.43      & 99.40     & 99.10       & 99.43                  & 99.15                          & \textbf{99.72} & 99.53      & 99.57         & 99.18 & \textbf{99.70} & 99.67          \\
\multirow{2}{*}{Age}               & Age-           & 95.90   & 94.92      & 95.30   & 96.27   & 96.67            & \textbf{96.77} & 96.60     & 96.54       & 96.27      & 96.05     & 95.70       & 96.04                  & 95.80                          & \textbf{96.49} & 96.32      & 96.39         & 95.79 & \textbf{96.52} & 96.50          \\
                                   & Age+           & 95.72   & 94.60      & 95.17   & 96.27   & 96.59            & \textbf{96.79} & 96.64     & 96.47       & 96.14      & 95.99     & 95.74       & 96.17                  & 95.67                          & \textbf{96.40} & 96.39      & 96.34         & 95.84 & \textbf{96.57} & 96.55          \\
\hline
\end{tabular}
}
\caption{Accuracy of 19 FR models on LFW-V level 2.}
\label{tab:accuracy_LFW-v_level2}
   \vspace{-2ex}
\end{table*}

\begin{table*}[]
\centering
 \scalebox{0.48}{
 \setlength{\tabcolsep}{2pt}  
\begin{tabular}{c|c|cccccccc|cccccc|ccccc}
\hline
\multicolumn{2}{c}{Models}                           & \multicolumn{8}{c}{Open-source Model Eval}                                                                                                                                        & \multicolumn{6}{c}{Architecture Eval}                                                                     & \multicolumn{5}{c}{Loss Function Eval}                           \\
\multicolumn{2}{c}{Variations}                          & FaceNet        & SphereFace & CosFace & ArcFace        & ElasticFace& AdaFace        & TransFace      & TopoFR & MobileFace     & Mobilenet & Mobilenet-v2 & ShuffleNet& ShuffleNet-v2& ResNet50       & Softmax-IR & SphereFace-IR  & Am-IR          & CosFace-IR     & ArcFace-IR     \\ \hline\hline
\rowcolor{lightgray} \multicolumn{2}{c}{None (clean)}                    & 99.23   & 98.20      & 98.63   & 99.50   & 99.80            & \textbf{99.83} & 99.75     & 99.78       & 99.43      & 99.40     & 99.10       & 99.43                  & 99.15                          & \textbf{99.72} & 99.53      & 99.57         & 99.18 & \textbf{99.70} & 99.67          \\
\multirow{2}{*}{Age}               & Age-           & 95.65   & 94.50      & 94.68   & 96.04   & 96.52            & \textbf{96.69} & 96.37     & 96.25       & 96.10      & 95.62     & 95.33       & 95.85                  & 95.47                          & \textbf{96.25} & 96.00      & 96.09         & 95.58 & 96.20          & \textbf{96.30} \\
                                   & Age+           & 95.32   & 93.91      & 94.85   & 96.17   & 96.54            & \textbf{96.65} & 96.49     & 96.35       & 96.04      & 95.72     & 95.47       & 95.92                  & 95.55                          & \textbf{96.22} & 96.30      & 96.29         & 95.57 & 96.44          & \textbf{96.55} \\
\hline
\end{tabular}
}
\caption{Accuracy of 19 FR models on LFW-V level 3.}
\label{tab:accuracy_LFW-v_level3}
   \vspace{-2ex}
\end{table*}

\begin{table*}[]
\centering
 \scalebox{0.48}{
 \setlength{\tabcolsep}{2pt}  
\begin{tabular}{c|c|cccccccc|cccccc|ccccc}
\hline
\multicolumn{2}{c}{Models}                           & \multicolumn{8}{c}{Open-source Model Eval}                                                                                                                                        & \multicolumn{6}{c}{Architecture Eval}                                                                     & \multicolumn{5}{c}{Loss Function Eval}                           \\
\multicolumn{2}{c}{Variations}                          & FaceNet        & SphereFace & CosFace & ArcFace        & ElasticFace& AdaFace        & TransFace      & TopoFR & MobileFace     & Mobilenet & Mobilenet-v2 & ShuffleNet& ShuffleNet-v2& ResNet50       & Softmax-IR & SphereFace-IR  & Am-IR          & CosFace-IR     & ArcFace-IR     \\ \hline\hline
\rowcolor{lightgray} \multicolumn{2}{c}{None (clean)}                    & 99.23   & 98.20      & 98.63   & 99.50   & 99.80            & \textbf{99.83} & 99.75     & 99.78       & 99.43          & 99.40     & 99.10       & 99.43                  & 99.15                          & \textbf{99.72} & 99.53      & 99.57         & 99.18 & \textbf{99.70} & 99.67          \\
\multirow{2}{*}{Age}               & Age-           & 94.70   & 93.61      & 93.54   & 95.79   & 96.14            & \textbf{96.40} & 95.94     & 95.97       & \textbf{95.70} & 94.82     & 94.53       & 95.47                  & 94.70                          & 95.67          & 95.42      & 95.52         & 94.93 & 95.85          & \textbf{95.95} \\
                                   & Age+           & 94.88   & 93.36      & 94.25   & 95.82   & 96.25            & \textbf{96.54} & 96.29     & 95.97       & 95.70          & 95.25     & 95.03       & 95.70                  & 95.22                          & \textbf{96.04} & 95.95      & 96.02         & 95.12 & 96.15          & \textbf{96.30} \\
\hline
\end{tabular}
}
\caption{Accuracy of 19 FR models on LFW-V level 4.}
\label{tab:accuracy_LFW-v_level4}
   \vspace{-2ex}
\end{table*}

\begin{table*}[]
\centering
 \scalebox{0.48}{
 \setlength{\tabcolsep}{2pt}  
\begin{tabular}{c|c|cccccccc|cccccc|ccccc}
\hline
\multicolumn{2}{c}{Models}                           & \multicolumn{8}{c}{Open-source Model Eval}                                                                                                                                        & \multicolumn{6}{c}{Architecture Eval}                                                                     & \multicolumn{5}{c}{Loss Function Eval}                           \\
\multicolumn{2}{c}{Variations}                          & FaceNet        & SphereFace & CosFace & ArcFace        & ElasticFace& AdaFace        & TransFace      & TopoFR & MobileFace     & Mobilenet & Mobilenet-v2 & ShuffleNet& ShuffleNet-v2& ResNet50       & Softmax-IR & SphereFace-IR  & Am-IR          & CosFace-IR     & ArcFace-IR     \\ \hline\hline
\rowcolor{lightgray} \multicolumn{2}{c}{None (clean)}                    & 99.23   & 98.20      & 98.63   & 99.50   & 99.80            & \textbf{99.83} & 99.75     & 99.78       & 99.43          & 99.40     & 99.10       & 99.43                  & 99.15                          & \textbf{99.72} & 99.53          & 99.57         & 99.18 & \textbf{99.70} & 99.67          \\
\multirow{2}{*}{Age}               & Age-           & 93.41   & 92.52      & 92.24   & 95.07   & 95.55            & \textbf{95.74} & 95.00     & 95.02       & \textbf{94.77} & 93.39     & 93.61       & 94.40                  & 93.66                          & 94.71          & 94.55          & 94.38         & 94.13 & 95.25          & \textbf{95.38} \\
                                   & Age+           & 94.20   & 92.42      & 93.59   & 95.38   & 95.85            & \textbf{96.24} & 95.89     & 95.57       & 95.13          & 94.43     & 94.41       & 95.27                  & 94.56                          & \textbf{95.58} & 95.58          & 95.47         & 94.73 & 95.75          & \textbf{96.09} \\
\hline
\end{tabular}
}
\caption{Accuracy of 19 FR models on LFW-V level 5.}
\label{tab:accuracy_LFW-v_level5}
   \vspace{-2ex}
\end{table*}



\begin{table*}[]
\centering
   \scalebox{0.47}{
\setlength{\tabcolsep}{2pt}  
\begin{tabular}{c|c|cccccccc|cccccc|ccccc}
\hline
\multicolumn{2}{c}{Models}                               & \multicolumn{8}{c}{Open-source Model Eval}                                                                                & \multicolumn{6}{c}{Architecture Eval}                                                                         & \multicolumn{5}{c}{Loss Function Eval}                                     \\
\multicolumn{2}{c}{Corruptions}                          & FaceNet        & SphereFace & CosFace & ArcFace        & ElasticFace& AdaFace        & TransFace      & TopoFR & MobileFace     & Mobilenet & Mobilenet-v2 & ShuffleNet& ShuffleNet-v2& ResNet50       & Softmax-IR & SphereFace-IR  & Am-IR          & CosFace-IR     & ArcFace-IR     \\ \hline\hline
\rowcolor{lightgray}\multicolumn{2}{c}{None (clean)}                         & 95.19   & 79.80      & 88.87   & 92.61   & \textbf{95.69}   & 95.45          & 94.60     & 94.51       & 92.37          & 89.81     & 88.86       & 91.59                  & 88.65                          & \textbf{93.58} & 93.85      & 94.18          & 90.44 & \textbf{94.57} & 94.47          \\
\multirow{5}{*}{Lighting \& Weather} & Brightness        & 93.35   & 78.23      & 85.78   & 91.27   & \textbf{95.13}   & 94.86          & 93.66     & 93.63       & 90.38          & 86.74     & 85.89       & 89.44                  & 86.38                          & \textbf{91.83} & 93.34      & 93.35          & 87.99 & \textbf{93.84} & 93.69          \\
                                     & Contrast          & 85.08   & 73.59      & 78.53   & 88.13   & 92.81            & \textbf{94.91} & 92.92     & 91.22       & 82.72          & 78.89     & 78.05       & 81.97                  & 77.74                          & \textbf{86.21} & 87.33      & \textbf{88.73} & 81.18 & 87.76          & 87.45          \\
\hline
\end{tabular}
}
\caption{Accuracy of 19 FR models on CFP-C.}
\label{tab:accuracy_CFP_C}
\vspace{-2ex}
\end{table*}

\begin{table*}[]
\centering
   \scalebox{0.47}{
\setlength{\tabcolsep}{2pt}  
\begin{tabular}{c|c|cccccccc|cccccc|ccccc}
\hline
\multicolumn{2}{c}{Models}                               & \multicolumn{8}{c}{Open-source Model Eval}                                                                                & \multicolumn{6}{c}{Architecture Eval}                                                                         & \multicolumn{5}{c}{Loss Function Eval}                                     \\
\multicolumn{2}{c}{Corruptions}                          & FaceNet        & SphereFace & CosFace & ArcFace        & ElasticFace& AdaFace        & TransFace      & TopoFR & MobileFace     & Mobilenet & Mobilenet-v2 & ShuffleNet& ShuffleNet-v2& ResNet50       & Softmax-IR & SphereFace-IR  & Am-IR          & CosFace-IR     & ArcFace-IR     \\ \hline\hline
\rowcolor{lightgray}\multicolumn{2}{c}{None (clean)}                         & 95.19   & 79.80      & 88.87   & 92.61   & \textbf{95.69}   & 95.45   & 94.60     & 94.51       & 92.37      & 89.81     & 88.86       & 91.59                  & 88.65                          & \textbf{93.58} & 93.85      & 94.18          & 90.44 & \textbf{94.57} & 94.47          \\
\multirow{5}{*}{Lighting \& Weather} & Brightness        & 94.90   & 79.76      & 88.75   & 92.37   & \textbf{95.57}   & 95.32   & 94.47     & 94.33       & 91.95      & 89.29     & 88.57       & 91.39                  & 88.38                          & \textbf{93.52} & 93.94      & 94.07          & 90.06 & \textbf{94.36} & 94.27          \\
                                     & Contrast          & 94.67   & 79.41      & 88.12   & 92.37   & \textbf{95.59}   & 95.25   & 94.63     & 94.51       & 91.86      & 88.60     & 88.41       & 90.61                  & 88.15                          & \textbf{93.03} & 93.66      & 94.15          & 89.42 & \textbf{94.43} & 94.15          \\
\hline
\end{tabular}
}
\caption{Accuracy of 19 FR models on CFP-C level 1.}
\label{tab:accuracy_CFP_C_level1}
\vspace{-2ex}
\end{table*}

\begin{table*}[]
\centering
   \scalebox{0.47}{
\setlength{\tabcolsep}{2pt}  
\begin{tabular}{c|c|cccccccc|cccccc|ccccc}
\hline
\multicolumn{2}{c}{Models}                               & \multicolumn{8}{c}{Open-source Model Eval}                                                                                & \multicolumn{6}{c}{Architecture Eval}                                                                         & \multicolumn{5}{c}{Loss Function Eval}                                     \\
\multicolumn{2}{c}{Corruptions}                          & FaceNet        & SphereFace & CosFace & ArcFace        & ElasticFace& AdaFace        & TransFace      & TopoFR & MobileFace     & Mobilenet & Mobilenet-v2 & ShuffleNet& ShuffleNet-v2& ResNet50       & Softmax-IR & SphereFace-IR  & Am-IR          & CosFace-IR     & ArcFace-IR     \\ \hline\hline
\rowcolor{lightgray}\multicolumn{2}{c}{None (clean)}                         & 95.19   & 79.80      & 88.87   & 92.61   & \textbf{95.69}   & 95.45          & 94.60     & 94.51       & 92.37          & 89.81     & 88.86       & 91.59                  & 88.65                          & \textbf{93.58} & 93.85          & 94.18          & 90.44 & \textbf{94.57} & 94.47          \\
\multirow{5}{*}{Lighting \& Weather} & Brightness        & 94.40   & 79.42      & 87.75   & 91.98   & \textbf{95.44}   & 95.09          & 94.27     & 94.05       & 91.45          & 88.21     & 87.42       & 90.51                  & 87.82                          & \textbf{92.89} & 93.71          & 93.72          & 89.47 & \textbf{94.36} & 94.18          \\
                                     & Contrast          & 94.47   & 79.18      & 87.42   & 92.34   & \textbf{95.57}   & 95.26          & 94.47     & 94.49       & 91.65          & 88.22     & 87.65       & 90.25                  & 87.49                          & \textbf{92.86} & 93.59          & 94.13          & 88.78 & \textbf{94.28} & 94.10          \\
\hline
\end{tabular}
}
\caption{Accuracy of 19 FR models on CFP-C level 2.}
\label{tab:accuracy_CFP_C_level2}
\vspace{-2ex}
\end{table*}

\begin{table*}[]
\centering
   \scalebox{0.47}{
\setlength{\tabcolsep}{2pt}  
\begin{tabular}{c|c|cccccccc|cccccc|ccccc}
\hline
\multicolumn{2}{c}{Models}                               & \multicolumn{8}{c}{Open-source Model Eval}                                                                                & \multicolumn{6}{c}{Architecture Eval}                                                                         & \multicolumn{5}{c}{Loss Function Eval}                                     \\
\multicolumn{2}{c}{Corruptions}                          & FaceNet        & SphereFace & CosFace & ArcFace        & ElasticFace& AdaFace        & TransFace      & TopoFR & MobileFace     & Mobilenet & Mobilenet-v2 & ShuffleNet& ShuffleNet-v2& ResNet50       & Softmax-IR & SphereFace-IR  & Am-IR          & CosFace-IR     & ArcFace-IR     \\ \hline\hline
\rowcolor{lightgray}\multicolumn{2}{c}{None (clean)}                         & 95.19   & 79.80      & 88.87   & 92.61   & \textbf{95.69}   & 95.45          & 94.60     & 94.51       & 92.37          & 89.81     & 88.86       & 91.59                  & 88.65                          & \textbf{93.58} & 93.85      & 94.18          & 90.44          & \textbf{94.57} & 94.47          \\
\multirow{5}{*}{Lighting \& Weather} & Brightness        & 93.41   & 78.39      & 86.22   & 91.45   & \textbf{95.26}   & 94.96          & 93.78     & 93.75       & 90.73          & 87.07     & 86.09       & 89.75                  & 87.00                          & \textbf{92.17} & 93.36      & 93.43          & 88.19          & \textbf{94.04} & 93.81          \\
                                     & Contrast          & 93.61   & 78.72      & 85.53   & 92.17   & \textbf{95.35}   & 95.25          & 94.27     & 94.33       & 90.67          & 87.27     & 86.11       & 89.59                  & 86.35                          & \textbf{92.34} & 93.33      & 93.88          & 87.56          & \textbf{93.91} & 93.85          \\
\hline
\end{tabular}
}
\caption{Accuracy of 19 FR models on CFP-C level 3.}
\label{tab:accuracy_CFP_C_level3}
\vspace{-2ex}
\end{table*}

\begin{table*}[]
\centering
   \scalebox{0.47}{
\setlength{\tabcolsep}{2pt}  
\begin{tabular}{c|c|cccccccc|cccccc|ccccc}
\hline
\multicolumn{2}{c}{Models}                               & \multicolumn{8}{c}{Open-source Model Eval}                                                                                & \multicolumn{6}{c}{Architecture Eval}                                                                         & \multicolumn{5}{c}{Loss Function Eval}                                     \\
\multicolumn{2}{c}{Corruptions}                          & FaceNet        & SphereFace & CosFace & ArcFace        & ElasticFace& AdaFace        & TransFace      & TopoFR & MobileFace     & Mobilenet & Mobilenet-v2 & ShuffleNet& ShuffleNet-v2& ResNet50       & Softmax-IR & SphereFace-IR  & Am-IR          & CosFace-IR     & ArcFace-IR     \\ \hline\hline
\rowcolor{lightgray}\multicolumn{2}{c}{None (clean)}                         & 95.19   & 79.80      & 88.87   & 92.61   & \textbf{95.69}   & 95.45          & 94.60          & 94.51       & 92.37          & 89.81     & 88.86       & 91.59                  & 88.65                          & \textbf{93.58} & 93.85      & 94.18          & 90.44          & \textbf{94.57} & 94.47          \\
\multirow{5}{*}{Lighting \& Weather} & Brightness        & 92.58   & 77.55      & 84.07   & 90.53   & \textbf{94.86}   & 94.69          & 93.22          & 93.20       & 89.37          & 85.63     & 84.54       & 88.65                  & 85.37                          & \textbf{91.02} & 93.09      & 93.05          & 86.78          & \textbf{93.53} & 93.42          \\
                                     & Contrast          & 86.47   & 72.96      & 75.93   & 89.33   & 94.41            & \textbf{95.00} & 93.79          & 93.22       & 83.93          & 77.64     & 75.69       & 84.09                  & 75.31                          & \textbf{89.14} & 91.42      & 92.27          & 80.86          & 91.88          & \textbf{92.50} \\
\hline
\end{tabular}
}
\caption{Accuracy of 19 FR models on CFP-C level 4.}
\label{tab:accuracy_CFP_C_level4}
\vspace{-2ex}
\end{table*}

\begin{table*}[]
\centering
   \scalebox{0.47}{
\setlength{\tabcolsep}{2pt}  
\begin{tabular}{c|c|cccccccc|cccccc|ccccc}
\hline
\multicolumn{2}{c}{Models}                               & \multicolumn{8}{c}{Open-source Model Eval}                                                                                & \multicolumn{6}{c}{Architecture Eval}                                                                         & \multicolumn{5}{c}{Loss Function Eval}                                     \\
\multicolumn{2}{c}{Corruptions}                          & FaceNet        & SphereFace & CosFace & ArcFace        & ElasticFace& AdaFace        & TransFace      & TopoFR & MobileFace     & Mobilenet & Mobilenet-v2 & ShuffleNet& ShuffleNet-v2& ResNet50       & Softmax-IR & SphereFace-IR  & Am-IR          & CosFace-IR     & ArcFace-IR     \\ \hline\hline
\rowcolor{lightgray}\multicolumn{2}{c}{None (clean)}                         & 95.19   & 79.80      & 88.87   & 92.61          & \textbf{95.69}   & 95.45          & 94.60          & 94.51       & 92.37          & 89.81     & 88.86       & 91.59                  & 88.65                          & \textbf{93.58} & 93.85          & 94.18          & 90.44          & \textbf{94.57} & 94.47          \\
\multirow{5}{*}{Lighting \& Weather} & Brightness        & 91.45   & 76.03      & 82.10   & 90.02          & \textbf{94.54}   & 94.23          & 92.58          & 92.81       & 88.39          & 83.53     & 82.84       & 86.91                  & 83.35                          & \textbf{89.56} & 92.58          & 92.50          & 85.44          & \textbf{92.93} & 92.79          \\
                                     & Contrast          & 56.17   & 57.68      & 55.64   & 74.46          & 83.12            & \textbf{93.79} & 87.44          & 79.55       & 55.51          & 52.70     & 52.38       & 55.29                  & 51.39                          & \textbf{63.67} & 64.65          & \textbf{69.23} & 59.28          & 64.28          & 62.66          \\
\hline
\end{tabular}
}
\caption{Accuracy of 19 FR models on CFP-C level 5.}
\label{tab:accuracy_CFP_C_level5}
\vspace{-2ex}
\end{table*}

\begin{table*}[]
\centering
 \scalebox{0.48}{
 \setlength{\tabcolsep}{2pt}  
\begin{tabular}{c|c|cccccccc|cccccc|ccccc}
\hline
\multicolumn{2}{c}{Models}                           & \multicolumn{8}{c}{Open-source Model Eval}                                                                                                                                        & \multicolumn{6}{c}{Architecture Eval}                                                                     & \multicolumn{5}{c}{Loss Function Eval}                           \\
\multicolumn{2}{c}{Variations}                          & FaceNet        & SphereFace & CosFace & ArcFace        & ElasticFace& AdaFace        & TransFace      & TopoFR & MobileFace     & Mobilenet & Mobilenet-v2 & ShuffleNet& ShuffleNet-v2& ResNet50       & Softmax-IR & SphereFace-IR  & Am-IR          & CosFace-IR     & ArcFace-IR     \\ \hline\hline
\rowcolor{lightgray} \multicolumn{2}{c}{None (clean)}                    & 95.19          & 79.80      & 88.87   & 92.61   & 95.69            & \textbf{95.45} & 94.60     & 94.51       & 92.37      & 89.81          & 88.86       & 91.59                  & 88.65                          & \textbf{93.58} & 93.85      & 94.18          & 90.44 & \textbf{94.57} & 94.47          \\
\multirow{2}{*}{Age}               & Age-           & 92.51          & 77.71      & 85.09   & 89.82   & \textbf{93.11}   & 92.69          & 91.68     & 91.67       & 87.78      & \textbf{83.31} & 83.94       & 87.07                  & 84.67                          & \textbf{89.30} & 90.68      & 91.38          & 85.76 & 91.34          & \textbf{91.41} \\
                                   & Age+           & \textbf{92.28} & 77.86      & 85.79   & 90.53   & \textbf{93.99}   & 93.66          & 92.65     & 92.49       & 89.09      & 84.59          & 84.94       & 87.98                  & 85.44                          & \textbf{90.72} & 92.24      & \textbf{92.82} & 86.58 & 92.55          & 92.48          \\
\hline
\end{tabular}
}
\caption{Accuracy of 19 FR models on CFP-V.}
\label{tab:accuracy_CFP_v}
   \vspace{-2ex}
\end{table*}

\begin{table*}[]
\centering
 \scalebox{0.48}{
 \setlength{\tabcolsep}{2pt}  
\begin{tabular}{c|c|cccccccc|cccccc|ccccc}
\hline
\multicolumn{2}{c}{Models}                           & \multicolumn{8}{c}{Open-source Model Eval}                                                                                                                                        & \multicolumn{6}{c}{Architecture Eval}                                                                     & \multicolumn{5}{c}{Loss Function Eval}                           \\
\multicolumn{2}{c}{Variations}                          & FaceNet        & SphereFace & CosFace & ArcFace        & ElasticFace& AdaFace        & TransFace      & TopoFR & MobileFace     & Mobilenet & Mobilenet-v2 & ShuffleNet& ShuffleNet-v2& ResNet50       & Softmax-IR & SphereFace-IR  & Am-IR          & CosFace-IR     & ArcFace-IR     \\ \hline\hline
\rowcolor{lightgray} \multicolumn{2}{c}{None (clean)}                    & 95.19   & 79.80      & 88.87   & 92.61   & \textbf{95.69}   & 95.45   & 94.60     & 94.51       & 92.37      & 89.81     & 88.86       & 91.59                  & 88.65                          & \textbf{93.58} & 93.85      & 94.18          & 90.44 & \textbf{94.57} & 94.47      \\
\multirow{2}{*}{Age}               & Age-           & 94.05   & 78.93      & 87.60   & 91.27   & \textbf{94.57}   & 94.15   & 93.51     & 93.16       & 89.68      & 85.60     & 86.45       & 89.11                  & 86.42                          & \textbf{91.62} & 92.47      & \textbf{93.22} & 87.54 & 93.19          & 92.97      \\
                                   & Age+           & 93.64   & 78.88      & 87.17   & 91.26   & \textbf{94.79}   & 94.28   & 93.58     & 93.38       & 89.96      & 85.95     & 86.15       & 89.13                  & 86.49                          & \textbf{91.88} & 92.84      & \textbf{93.46} & 87.75 & 93.33          & 93.16      \\
\hline
\end{tabular}
}
\caption{Accuracy of 19 FR models on CFP-V level 1.}
\label{tab:accuracy_CFP_v_level1}
   \vspace{-2ex}
\end{table*}

\begin{table*}[]
\centering
 \scalebox{0.48}{
 \setlength{\tabcolsep}{2pt}  
\begin{tabular}{c|c|cccccccc|cccccc|ccccc}
\hline
\multicolumn{2}{c}{Models}                           & \multicolumn{8}{c}{Open-source Model Eval}                                                                                                                                        & \multicolumn{6}{c}{Architecture Eval}                                                                     & \multicolumn{5}{c}{Loss Function Eval}                           \\
\multicolumn{2}{c}{Variations}                          & FaceNet        & SphereFace & CosFace & ArcFace        & ElasticFace& AdaFace        & TransFace      & TopoFR & MobileFace     & Mobilenet & Mobilenet-v2 & ShuffleNet& ShuffleNet-v2& ResNet50       & Softmax-IR & SphereFace-IR  & Am-IR          & CosFace-IR     & ArcFace-IR     \\ \hline\hline
\rowcolor{lightgray} \multicolumn{2}{c}{None (clean)}                    & 95.19   & 79.80      & 88.87   & 92.61   & \textbf{95.69}   & 95.45   & 94.60     & 94.51       & 92.37      & 89.81     & 88.86       & 91.59                  & 88.65                          & \textbf{93.58} & 93.85      & 94.18          & 90.44 & \textbf{94.57} & 94.47          \\
\multirow{2}{*}{Age}               & Age-           & 94.01   & 78.70      & 86.84   & 90.76   & \textbf{94.36}   & 93.95   & 93.23     & 93.09       & 89.24      & 85.14     & 85.70       & 88.41                  & 85.90                          & \textbf{90.70} & 92.15      & \textbf{92.76} & 86.94 & 92.71          & 92.46          \\
                                   & Age+           & 93.13   & 78.55      & 86.75   & 91.00   & \textbf{94.51}   & 94.17   & 93.52     & 93.16       & 89.81      & 85.62     & 85.76       & 88.91                  & 86.11                          & \textbf{91.36} & 92.69      & \textbf{93.26} & 87.33 & 93.10          & 92.96          \\
\hline
\end{tabular}
}
\caption{Accuracy of 19 FR models on CFP-V level 2.}
\label{tab:accuracy_CFP_v_level2}
   \vspace{-2ex}
\end{table*}

\begin{table*}[]
\centering
 \scalebox{0.48}{
 \setlength{\tabcolsep}{2pt}  
\begin{tabular}{c|c|cccccccc|cccccc|ccccc}
\hline
\multicolumn{2}{c}{Models}                           & \multicolumn{8}{c}{Open-source Model Eval}                                                                                                                                        & \multicolumn{6}{c}{Architecture Eval}                                                                     & \multicolumn{5}{c}{Loss Function Eval}                           \\
\multicolumn{2}{c}{Variations}                          & FaceNet        & SphereFace & CosFace & ArcFace        & ElasticFace& AdaFace        & TransFace      & TopoFR & MobileFace     & Mobilenet & Mobilenet-v2 & ShuffleNet& ShuffleNet-v2& ResNet50       & Softmax-IR & SphereFace-IR  & Am-IR          & CosFace-IR     & ArcFace-IR     \\ \hline\hline
\rowcolor{lightgray} \multicolumn{2}{c}{None (clean)}                    & 95.19   & 79.80      & 88.87   & 92.61   & \textbf{95.69}   & 95.45          & 94.60     & 94.51       & 92.37      & 89.81     & 88.86       & 91.59                  & 88.65                          & \textbf{93.58} & 93.85      & 94.18          & 90.44 & \textbf{94.57} & 94.47          \\
\multirow{2}{*}{Age}               & Age-           & 93.12   & 78.14      & 85.60   & 90.08   & \textbf{93.68}   & 93.17          & 92.41     & 92.22       & 88.08      & 83.71     & 84.38       & 87.56                  & 85.14                          & \textbf{89.65} & 90.93      & \textbf{91.94} & 86.02 & 91.79          & 91.82          \\
                                   & Age+           & 92.44   & 77.85      & 86.19   & 90.54   & \textbf{93.97}   & 93.71          & 92.80     & 92.79       & 89.19      & 84.92     & 85.31       & 88.21                  & 85.56                          & \textbf{90.83} & 92.50      & \textbf{92.99} & 86.87 & 92.56          & 92.76          \\
\hline
\end{tabular}
}
\caption{Accuracy of 19 FR models on CFP-V level 3.}
\label{tab:accuracy_CFP_v_level3}
   \vspace{-2ex}
\end{table*}

\begin{table*}[]
\centering
 \scalebox{0.48}{
 \setlength{\tabcolsep}{2pt}  
\begin{tabular}{c|c|cccccccc|cccccc|ccccc}
\hline
\multicolumn{2}{c}{Models}                           & \multicolumn{8}{c}{Open-source Model Eval}                                                                                                                                        & \multicolumn{6}{c}{Architecture Eval}                                                                     & \multicolumn{5}{c}{Loss Function Eval}                           \\
\multicolumn{2}{c}{Variations}                          & FaceNet        & SphereFace & CosFace & ArcFace        & ElasticFace& AdaFace        & TransFace      & TopoFR & MobileFace     & Mobilenet & Mobilenet-v2 & ShuffleNet& ShuffleNet-v2& ResNet50       & Softmax-IR & SphereFace-IR  & Am-IR          & CosFace-IR     & ArcFace-IR     \\ \hline\hline
\rowcolor{lightgray} \multicolumn{2}{c}{None (clean)}                    & 95.19   & 79.80      & 88.87   & 92.61   & \textbf{95.69}   & 95.45          & 94.60     & 94.51       & 92.37      & 89.81     & 88.86       & 91.59                  & 88.65                          & \textbf{93.58} & 93.85      & 94.18          & 90.44 & \textbf{94.57} & 94.47          \\
\multirow{2}{*}{Age}               & Age-           & 91.68   & 77.09      & 83.79   & 89.16   & \textbf{92.34}   & 92.02          & 90.81     & 90.80       & 86.91      & 82.33     & 82.84       & 86.13                  & 83.89                          & \textbf{88.42} & 89.82      & 90.51          & 84.91 & 90.31          & \textbf{90.60} \\
                                   & Age+           & 91.53   & 77.28      & 85.07   & 90.27   & \textbf{93.69}   & 93.45          & 92.07     & 92.02       & 88.55      & 84.00     & 84.26       & 87.19                  & 84.94                          & \textbf{90.21} & 91.91      & \textbf{92.57} & 85.90 & 92.28          & 92.14          \\
\hline
\end{tabular}
}
\caption{Accuracy of 19 FR models on CFP-V level 4.}
\label{tab:accuracy_CFP_v_level4}
   \vspace{-2ex}
\end{table*}

\begin{table*}[]
\centering
 \scalebox{0.48}{
 \setlength{\tabcolsep}{2pt}  
\begin{tabular}{c|c|cccccccc|cccccc|ccccc}
\hline
\multicolumn{2}{c}{Models}                           & \multicolumn{8}{c}{Open-source Model Eval}                                                                                                                                        & \multicolumn{6}{c}{Architecture Eval}                                                                     & \multicolumn{5}{c}{Loss Function Eval}                           \\
\multicolumn{2}{c}{Variations}                          & FaceNet        & SphereFace & CosFace & ArcFace        & ElasticFace& AdaFace        & TransFace      & TopoFR & MobileFace     & Mobilenet & Mobilenet-v2 & ShuffleNet& ShuffleNet-v2& ResNet50       & Softmax-IR & SphereFace-IR  & Am-IR          & CosFace-IR     & ArcFace-IR     \\ \hline\hline
\rowcolor{lightgray} \multicolumn{2}{c}{None (clean)}                    & 95.19   & 79.80      & 88.87   & 92.61   & \textbf{95.69}   & 95.45          & 94.60     & 94.51       & 92.37      & 89.81     & 88.86       & 91.59                  & 88.65                          & \textbf{93.58} & 93.85      & 94.18          & 90.44 & \textbf{94.57} & 94.47          \\
\multirow{2}{*}{Age}               & Age-           & 89.70   & 75.67      & 81.63   & 87.85   & \textbf{90.58}   & 90.15          & 88.44     & 89.09       & 84.97      & 79.74     & 80.33       & 84.15                  & 81.99                          & \textbf{86.11} & 88.05      & 88.48          & 83.38 & 88.71          & \textbf{89.19} \\
                                   & Age+           & 90.66   & 76.73      & 83.76   & 89.59   & \textbf{92.99}   & 92.67          & 91.26     & 91.12       & 87.96      & 82.48     & 83.21       & 86.45                  & 84.09                          & \textbf{89.34} & 91.25      & \textbf{91.81} & 85.04 & 91.49          & 91.38          \\
\hline
\end{tabular}
}
\caption{Accuracy of 19 FR models on CFP-V level 5.}
\label{tab:accuracy_CFP_v_level5}
   \vspace{-2ex}
\end{table*}


 \clearpage

\begin{table*}[]
\centering
   \scalebox{0.47}{
\setlength{\tabcolsep}{2pt}  
\begin{tabular}{c|c|cccccccc|cccccc|ccccc}
\hline
\multicolumn{2}{c}{Models}                               & \multicolumn{8}{c}{Open-source Model Eval}                                                                                & \multicolumn{6}{c}{Architecture Eval}                                                                         & \multicolumn{5}{c}{Loss Function Eval}                                     \\
\multicolumn{2}{c}{Corruptions}                          & FaceNet        & SphereFace & CosFace & ArcFace        & ElasticFace& AdaFace        & TransFace      & TopoFR & MobileFace     & Mobilenet & Mobilenet-v2 & ShuffleNet& ShuffleNet-v2& ResNet50       & Softmax-IR & SphereFace-IR  & Am-IR          & CosFace-IR     & ArcFace-IR     \\ \hline\hline
\rowcolor{lightgray}\multicolumn{2}{c}{None (clean)}                         & 91.47   & 86.61      & 88.54   & \textbf{93.08} & 93.06            & 93.00          & 92.96          & 92.30       & 91.70          & 91.34     & 91.22       & 91.85                  & 91.02                          & \textbf{92.76} & 92.43      & 92.83          & 91.09          & 92.91          & \textbf{93.00} \\
\multirow{5}{*}{Lighting \& Weather} & Brightness        & 89.69   & 83.65      & 86.65   & 92.23          & 92.57            & \textbf{92.73} & 92.46          & 91.61       & 90.49          & 89.76     & 89.41       & 90.58                  & 89.65                          & \textbf{91.53} & 91.74      & 92.06          & 89.61          & 92.33          & \textbf{92.39} \\
                                     & Contrast          & 76.43   & 70.63      & 74.31   & 85.40          & 86.26            & \textbf{91.33} & 88.03          & 85.29       & 78.39          & 75.46     & 75.61       & 77.14                  & 74.36                          & \textbf{80.90} & 82.08      & \textbf{82.90} & 79.24          & 81.60          & 81.29          \\
\hline
\end{tabular}
}
\caption{Accuracy of 19 FR models on YTF-C.}
\label{tab:accuracy_YTF-C}
\vspace{-2ex}
\end{table*}

\begin{table*}[]
\centering
   \scalebox{0.47}{
\setlength{\tabcolsep}{2pt}  
\begin{tabular}{c|c|cccccccc|cccccc|ccccc}
\hline
\multicolumn{2}{c}{Models}                               & \multicolumn{8}{c}{Open-source Model Eval}                                                                                & \multicolumn{6}{c}{Architecture Eval}                                                                         & \multicolumn{5}{c}{Loss Function Eval}                                     \\
\multicolumn{2}{c}{Corruptions}                          & FaceNet        & SphereFace & CosFace & ArcFace        & ElasticFace& AdaFace        & TransFace      & TopoFR & MobileFace     & Mobilenet & Mobilenet-v2 & ShuffleNet& ShuffleNet-v2& ResNet50       & Softmax-IR & SphereFace-IR  & Am-IR          & CosFace-IR     & ArcFace-IR     \\ \hline\hline
\rowcolor{lightgray}\multicolumn{2}{c}{None (clean)}                         & 91.47   & 86.61      & 88.54   & \textbf{93.08} & 93.06            & 93.00          & 92.96          & 92.30       & 91.70          & 91.34     & 91.22       & 91.85                  & 91.02                          & \textbf{92.76} & 92.43      & 92.83         & 91.09 & 92.91          & \textbf{93.00} \\
\multirow{5}{*}{Lighting \& Weather} & Brightness        & 91.28   & 86.36      & 88.42   & \textbf{93.02} & 92.93            & \textbf{93.02} & 92.91          & 92.15       & 91.53          & 91.22     & 91.19       & 91.62                  & 90.90                          & \textbf{92.62} & 92.40      & 92.85         & 90.94 & \textbf{92.89} & 92.81          \\
                                     & Contrast          & 90.52   & 83.47      & 87.21   & 92.87          & 92.87            & 93.00          & \textbf{93.02} & 92.04       & 91.17          & 90.41     & 90.56       & 91.05                  & 89.71                          & \textbf{92.28} & 92.30      & 92.76         & 90.49 & 92.74          & \textbf{92.98} \\
\hline
\end{tabular}
}
\caption{Accuracy of 19 FR models on YTF-C level 1.}
\label{tab:accuracy_YTF_C_level1}
\vspace{-2ex}
\end{table*}

\begin{table*}[]
\centering
   \scalebox{0.47}{
\setlength{\tabcolsep}{2pt}  
%
}
\caption{Accuracy of 19 FR models on YTF-C level 2.}
\label{tab:accuracy_YTF_C_level2}
\vspace{-2ex}
\end{table*}

\begin{table*}[]
\centering
   \scalebox{0.47}{
\setlength{\tabcolsep}{2pt}  
%
}
\caption{Accuracy of 19 FR models on YTF-C level 3.}
\label{tab:accuracy_YTF_C_level3}
\vspace{-2ex}
\end{table*}

\begin{table*}[]
\centering
   \scalebox{0.47}{
\setlength{\tabcolsep}{2pt}  
%
}
\caption{Accuracy of 19 FR models on YTF-C level 5.}
\label{tab:accuracy_YTF_C_level5}
\vspace{-2ex}
\end{table*}

\begin{table*}[]
\centering
 \scalebox{0.48}{
 \setlength{\tabcolsep}{2pt}  
%
}
\caption{Accuracy of 19 FR models on YTF-V.}
\label{tab:accuracy_YTF_v}
   \vspace{-2ex}
\end{table*}

\begin{table*}[]
\centering
   \scalebox{0.47}{
\setlength{\tabcolsep}{2pt}  
%
}
\caption{Accuracy of 19 FR models on YTF-V level1.}
\label{tab:accuracy_YTF_v_level1}
\vspace{-2ex}
\end{table*}

\begin{table*}[]
\centering
   \scalebox{0.47}{
\setlength{\tabcolsep}{2pt}  
%
}
\caption{Accuracy of 19 FR models on YTF-V level2.}
\label{tab:accuracy_YTF_v_level2}
\vspace{-2ex}
\end{table*}

\begin{table*}[]
\centering
   \scalebox{0.47}{
\setlength{\tabcolsep}{2pt}  
%
}
\caption{Accuracy of 19 FR models on YTF-V level3.}
\label{tab:accuracy_YTF_v_level3}
\vspace{-2ex}
\end{table*}

\begin{table*}[]
\centering
   \scalebox{0.47}{
\setlength{\tabcolsep}{2pt}  
%
}
\caption{Accuracy of 19 FR models on YTF-V level5.}
\label{tab:accuracy_YTF_v_level5}
\vspace{-2ex}
\end{table*}


\begin{table*}[]
\centering
   \scalebox{0.60}{
\setlength{\tabcolsep}{2pt}  
%
}
\caption{Accuracy of 8 open-source models on Mask-A of corruptions.}
\label{tab:accuracy_c_Mask-A}
\vspace{-2ex}
\end{table*}

\begin{table*}[]
\centering
   \scalebox{0.60}{
\setlength{\tabcolsep}{2pt}  
%
}
\caption{Accuracy of 8 open-source models on Mask-B of corruptions.}

\label{tab:accuracy_c_Mask-B}
\vspace{-2ex}
\end{table*}

\begin{table*}[]
\centering
   \scalebox{0.60}{
\setlength{\tabcolsep}{2pt}  
%
}
\caption{Accuracy of 8 open-source models on Mask-C of corruptions.}

\label{tab:accuracy_c_Mask-C}
\vspace{-2ex}
\end{table*}

\begin{table*}[]
\centering
   \scalebox{0.60}{
\setlength{\tabcolsep}{2pt}  
%
}
\caption{Accuracy of 8 open-source models on Mask-D of corruptions.}

\label{tab:accuracy_c_Mask-D}
\vspace{-2ex}
\end{table*}

\begin{table*}[]
\centering
   \scalebox{0.60}{
\setlength{\tabcolsep}{2pt}  
%
}
\caption{Accuracy of 8 open-source models on Mask-E of corruptions.}

\label{tab:accuracy_c_Mask-E}
\vspace{-2ex}
\end{table*}



\begin{table*}[]
\centering
   \scalebox{0.60}{
\setlength{\tabcolsep}{2pt}  
%
}
\caption{Accuracy of 8 open-source models on Mask-A of variations.}

\label{tab:accuracy_v_Mask-A}
\vspace{-2ex}
\end{table*}

\begin{table*}[]
\centering
   \scalebox{0.60}{
\setlength{\tabcolsep}{2pt}  
%
}
\caption{Accuracy of 8 open-source models on Mask-B of variations.}
\label{tab:accuracy_v_Mask-B}
\vspace{-2ex}
\end{table*}

\begin{table*}[]
\centering
   \scalebox{0.60}{
\setlength{\tabcolsep}{2pt}  
%
}
\caption{Accuracy of 8 open-source models on Mask-C of variations.}
\label{tab:accuracy_v_Mask-C}
\vspace{-2ex}
\end{table*}

\begin{table*}[!t]
\centering
   \scalebox{0.60}{
\setlength{\tabcolsep}{2pt}  
%
}
\caption{Accuracy of 8 open-source models on Mask-D of variations.}
\label{tab:accuracy_v_Mask-D}
\vspace{-2ex}
\end{table*}

\begin{table*}[!t]
\centering
   \scalebox{0.60}{
\setlength{\tabcolsep}{2pt}  
%
}
\caption{Accuracy of 8 open-source models on Mask-E of variations.}
\label{tab:accuracy_v_Mask-E}
\vspace{-2ex}
\end{table*}


\end{CJK}
\end{document}